%% file: main.tex
\documentclass[10pt,twocolumn,letterpaper]{article}

\newif\ifwacvfinal

\newif\ifnographics

\wacvfinaltrue
\nographicsfalse

\ifwacvfinal
\usepackage[algorithms]{wacv}
\else
\usepackage[review,algorithms]{wacv}
\fi

\usepackage{times}
\usepackage{graphicx}
\ifnographics
\usepackage[allfiguresdraft]{draftfigure}

\else
\fi

\usepackage{amsmath}
\usepackage{amssymb}

\usepackage{booktabs}
\usepackage{multirow} %
\usepackage{array}
\usepackage[export]{adjustbox}
\usepackage{todonotes}
\usepackage{subcaption}
\usepackage{caption}

\usepackage{tikz}
\usetikzlibrary{backgrounds}
\usetikzlibrary{fit}
\usetikzlibrary{positioning} %
\usetikzlibrary{calc}
\usetikzlibrary{patterns} %
\usetikzlibrary{quotes} %
\usetikzlibrary{calc}
\usetikzlibrary{intersections}

\usepackage{pgfplots}
\pgfplotsset{compat=1.18}

\usetikzlibrary{shapes.arrows}
\usetikzlibrary{decorations.markings}
\usepackage{siunitx}

\tikzset{outside/.style={
    postaction={
        decorate,
        decoration={
            markings,
            mark=at position \pgfdecoratedpathlength-0.1pt with {\arrow[cyan,line width=#1] {>}; },
            mark=between positions 0 and \pgfdecoratedpathlength-1.5pt step 0.1pt with {
                \pgfmathsetmacro\myval{multiply(divide(
                    \pgfkeysvalueof{/pgf/decoration/mark info/distance from start}, \pgfdecoratedpathlength),100)};
                \pgfsetfillcolor{cyan!\myval!blue};
                \pgfpathcircle{\pgfpointorigin}{#1};
                \pgfusepath{fill};}
}}}}
\tikzset{inside/.style={
    postaction={
        decorate,
        decoration={
            markings,
            mark=at position \pgfdecoratedpathlength-0.1pt with {\arrow[yellow,line width=#1] {>}; },
            mark=between positions 0 and \pgfdecoratedpathlength-1.5pt step 0.1pt with {
                \pgfmathsetmacro\myval{multiply(divide(
                    \pgfkeysvalueof{/pgf/decoration/mark info/distance from start}, \pgfdecoratedpathlength),100)};
                \pgfsetfillcolor{yellow!\myval!red};
                \pgfpathcircle{\pgfpointorigin}{#1};
                \pgfusepath{fill};}
}}}}
\usepackage{colortbl}
\def\cca#1{\cellcolor{black!#10}\ifnum #1>5\color{white}\fi{#1}} %
\usepackage{pgf}
\usepackage{collcell}
\usepackage[accsupp]{axessibility} 

\newcommand*{\MinNumber}{0.28}%
\newcommand*{\MaxNumber}{0.9}%

\newcommand{\ApplyGradient}[1]{%
  \pgfmathsetmacro{\PercentColor}{100.0*(#1-\MinNumber)/(\MaxNumber-\MinNumber)}%
  \edef\x{\noexpand\cellcolor{blue!\PercentColor!red}}\x\textcolor{white}{#1}%
}
\newcolumntype{R}{>{\collectcell\ApplyGradient}{r}<{\endcollectcell}}

\usepackage{diagbox} %

\usepackage[acronym]{glossaries}
\newacronym{ssc}{SSC}{Semantic Scene Completion}
\newacronym{sdf}{SDF}{Signed Distance Function}
\newacronym{udf}{UDF}{Unsigned Distance Function}
\newacronym{miou}{mIoU}{Mean Intersection over Union}
\newacronym{iou}{IoU}{Intersection over Union}
\newacronym{icp}{ICP}{Iterative Closest Point}
\newacronym{op}{OP}{Occupancy Predictor}
\newacronym{pce}{PCE}{Point Cloud Encoder}

\usepackage[breaklinks=true,bookmarks=false]{hyperref}

\def\wacvPaperID{12486} %

\ifwacvfinal\fi

\usepackage[capitalize]{cleveref}
\crefname{section}{Sec.}{Secs.}
\Crefname{section}{Section}{Sections}
\Crefname{table}{Table}{Tables}
\crefname{table}{Tab.}{Tabs.}

\def\wacvPaperID{173} %
\def\confName{WACV}
\def\confYear{2023}

\begin{document}

\title{Registered and Segmented Deformable Object Reconstruction\\ from a Single View Point Cloud}

\author{
    Pit Henrich$^{1}$,
    Bal\'{a}zs Gyenes$^{2}$,
    Paul Maria Scheikl$^{1}$,\\
    Gerhard Neumann$^{2}$,
    Franziska Mathis-Ullrich$^{1\ +}$\\
    $^1$ Dep. Artificial Intelligence in Biomedical Engineering - FAU Erlangen-Nürnberg, Erlangen, Germany\\
    $^2$ Institute for Anthropomatics and Robotics - Karlsruhe Institute of Technology, Karlsruhe, Germany\\
    {\tt\small \{pit.henrich,
    paul.m.scheikl,
    franziska.mathis-ullrich\}@fau.de}\\
    {\tt\small \{balazs.gyenes,
    gerhard.neumann\}@kit.edu}
}

\twocolumn[{%
\renewcommand\twocolumn[1][]{#1}%
\maketitle
\begin{center}
    \centering
    \captionsetup{type=figure}
    \includegraphics[width=\textwidth]{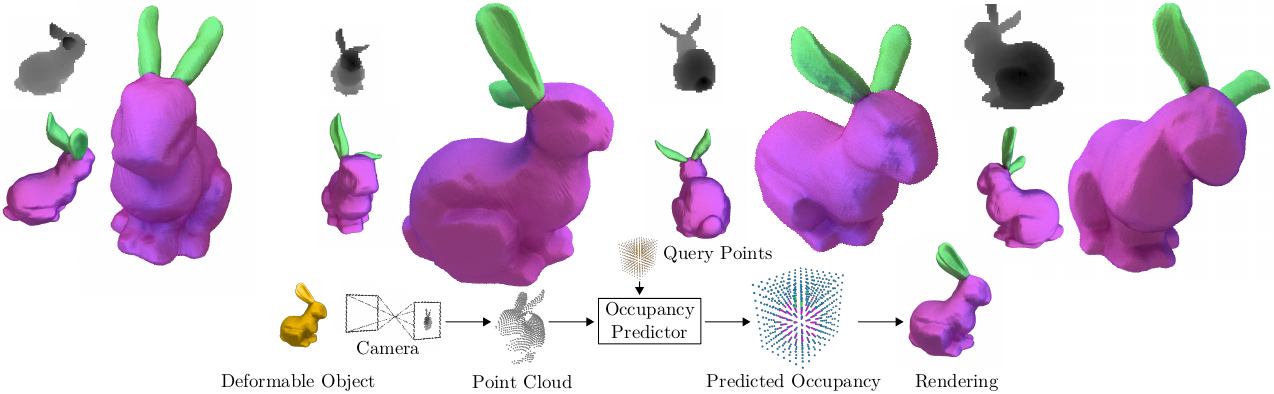}
    \captionof{figure}{A Point Cloud and Query Points from a regular grid are passed to the Occupancy Predictor to infer the volumetric model of a deformed object.
    }
    \label{fig:overview}
\end{center}%
}]
\maketitle

\ifwacvfinal
    \def\thefootnote{+}\footnotetext{Corresponding author}\def\thefootnote{\arabic{footnote} }
\fi

\begin{abstract}
    In deformable object manipulation, we often want to interact with specific segments of an object that are only defined in non-deformed models of the object.
    We thus require a system that can recognize and locate these segments in sensor data of deformed real world objects.
    This is normally done using deformable object registration, which is problem specific and complex to tune.
    Recent methods utilize neural occupancy functions to improve deformable object registration by registering to an object reconstruction.
    Going one step further, we propose a system that in addition to reconstruction learns segmentation of the reconstructed object.
    As the resulting output already contains the information about the segments, we can skip the registration process.
    Tested on a variety of deformable objects in simulation and the real world, we demonstrate that our method learns to robustly find these segments.
    We also introduce a simple sampling algorithm to generate better training data for occupancy learning.
\end{abstract}

\section{Introduction}
    \label{sec:intro}
    
    Our research is driven by the demands of robot-assisted surgical applications.
    These include the ability to find and interact with specific segments, for example to grasp or cut them.
    Soft organs are highly deformable, which makes these interactions challenging.
    To find segments of interest, current work focuses on registering a known segmented model to sensor data.
    The known models can be obtained from a medical CT, where the segments are then defined manually.

    Instead of using deformable registration methods, we propose a method to directly reconstruct deformable objects with all their segments from depth images.
    The segmented reconstruction can then be used instead of a registered model for planning the interactions.

    Emerging object reconstruction methods learn a continuous object representation to create voxel or surface meshes at any resolution.
    For example, neural occupancy functions~\cite{meschederOccupancy2019, jiaLearning2020} or \gls{sdf}s~\cite{parkDeepSDF2019, sitzmann2020metasdf}.
    Labeled points in 3D space are used to train such continuous representation.
    However, the impact of generation and distribution of these points is rarely detailed or investigated.

    \begin{figure}[tbh]
        \centering
        \includegraphics{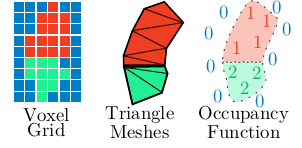}
        \caption{Three ways to represent objects: using a regular grid of voxels, using two Triangle Meshes, and using an Occupancy Function. In this example, the Occupancy Function outputs values of $0,1$, or $2$ for every point depending on its location. The shape of the object is implicitly defined by the boundaries between regions of the same value.}
        \label{fig:occupancy_vs_others}
    \end{figure}

    Binary neural occupancy is usually defined as a function in 3D space that takes on discrete values $0$ and $1$ to indicate if the point is inside or outside of a solid object, \ie $o_{\text{binary}} : \mathbb{R}^3 \rightarrow \{0, 1\}.$
    We consider an extension to this definition to include multiple class labels $o_{\text{multi}} : \mathbb{R}^3 \rightarrow \{0, 1, ..., n\}$, 
    where $n$ is the number of segments.
    \cref{fig:occupancy_vs_others} illustrates how an occupancy function encodes an object in comparison to voxel and triangle models.
    We consider $0$ to be the label for empty space.
    Objects are implicitly defined by the boundaries between the clusters of points with the same occupancy value.
    A multi-class occupancy function with ${0, \cdots, n}$ labels can be used to represent an object with $n$ segments.
    By conditioning this multi-class occupancy function on sensor data, it can be used to reconstruct objects in different deformation states.

    \paragraph{Contribution}
    We train a neural system end-to-end for 3D reconstruction and segmentation using and demonstrate its ability to reconstruct objects accurately from a single-view point cloud, despite deformations and occlusions.
    Our contributions include:
    
    \noindent
    1) Extending binary occupancy functions over continuous domains to multi-class occupancy.
    
    \noindent
    2) Developing an algorithm with few hyperparameters for generating high-quality training data for occupancy learning.

    \noindent
    3) Provide an alternative to traditional deformable object registration methods.

\section{Related Work}
    \subsection{Semantic Scene Completion}
        \gls{ssc} provides a labeled 3D environment model based on sensor data.
        This model can be used for tasks such as autonomous navigation.
        Based on a single depth image, Song et al.~\cite{songSemantic2017} predict semantic labels for a voxel grid.
        
        To improve \gls{ssc}, multi-modal (color and depth images) sensor data can be used~\cite{liRGBD2019}.
        Cai et al.~\cite{caiSemantic2021} further improve \gls{ssc} by extracting object instances from the scene and propagating the object details back up to the scene. 

        When labeling a voxel grid using convolutional architectures, there is a trade-off between spatial resolution and memory requirements.
        For example, based on a depth image, Song et al.~\cite{songSemantic2017} use a limited resolution of ($240\times 144\times240$) to approximate the structure of a room.        
        Rist et al.~\cite{rist2021semantic} use a continuous representation that is not based on voxelization for semantic scene completion.
        Thereby, surpassing previous voxel-based methods in geometric accuracy.
    
    \subsection{Shape Reconstruction by Learning Implicit Functions}
        A central challenge in utilizing neural networks for 3D reconstruction is choosing a shape representation.
        Several representations, including voxel grids~\cite{choy20163d, girdhar2016learning, tulsiani2017multi}, dense point clouds~\cite{fan2017point,lin2018learning, yu2018pu,luo2021diffusion, zhou20213d}, polygonal meshes~\cite{litany2018deformable, gkioxari2019mesh,hanocka2019meshcnn, chen2020bsp, deng2020cvxnet, gao2020learning, halimi2020whole, hanocka2020point2mesh, shen2021deep, hui2022neural}, and manifold atlases~\cite{groueix2018papier, williams2019deep, deprelle2019learning, badki2020meshlet, gadelha2021deep} have been proposed.
        Each of these have unique advantages and disadvantages.

        Voxel-based methods suffer from memory constraints.
        Point cloud methods do not define any surfaces, so they have to be approximated.
        Polygonal mesh and manifold atlas methods often output meshes with degenerate geometry or holes.
        Mesh learning is complex as it involves defining problem specific loss functions and strong regularizers.

        Recent work investigates continuous reconstruction methods that circumvent these challenges.
        Continuous methods represent surfaces as decision boundaries.
        There is no direct output of a mesh or model.
        The object is encoded implicitly in the weights of a neural network.
        For any point in space, the network predicts if it is inside or outside of an object.
        Therefore, the object can be retrieved at any resolution.
        By using equidistant query points, a voxel model can be obtained.
        Marching cubes along the decision boundary can be used to create a triangle mesh.

        Popular continuous representations are occupancy and \gls{sdf} networks.
        Occupancy networks~\cite{meschederOccupancy2019} divide the object space into clusters.
        The decision boundaries between the clusters represent the surfaces of the object.

        Alternatively, a \gls{sdf} can be used~\cite{parkDeepSDF2019}.
        A \gls{sdf} provides the distance to an object's surface from any point in space.
        The sign indicates if the point is inside or outside of an object.
        All points with a distance of $0$ define the surface of the object.
        Meta-learning allows faster adaptation to a specific object instances~\cite{sitzmann2020metasdf}.
        \gls{sdf} have also been used to represent deformable objects\cite{litany2018deformable}.

        A \gls{sdf}, like an occupancy function, divides a space into insides and outsides.
        There are scenarios where this is not possible.
        For example, walls scanned by a LiDAR sensor have a thicknesses that can not be inferred.
        Chebane et al.\cite{chibaneNeural2020} learn an \gls{udf}.
        This \gls{udf} can represent thin surfaces such as a plane.
        
        Lamb et al.~\cite{lamb2022deepjoin} propose a hybrid approach that combines occupancy, signed distance field, and normal estimation methods, leading to improved reconstruction accuracy.

        Williams et al.~\cite{williams2022neural} propose predicting an implicit surface using a trained kernel.
        They demonstrate that neural kernel fields can effectively reconstruct shapes when the kernel possesses an appropriate inductive bias.

    \subsection{Point Cloud Encoders}
        Object reconstruction usually depends on sensor information, including RGB images, depth images, or point clouds.
        Image data, RGB and depth, obtained from cameras contain camera dependent distortion, such as perspective distortion.
        Objects further away from the camera occupy less pixels in the depth image.
        Projecting a depth image back into 3D space creates a point cloud.
        In this point cloud, the scale of equally sized objects in the fore- and background are the same.

        Working with point clouds presents unique challenges, as any function applied to a point cloud should remain invariant to the point order~\cite{qiPointNet2017}.

        PointNet++~\cite{qiPointNet2017a} is a popular hierarchical architecture that is capable of recognizing structures at various size scales. %
        PCNN~\cite{atzmonPoint2018} extend convolutional operations to point clouds, outperforming PointNet++ in segmentation and classification tasks.
        Zhao et al.~\cite{zhao2021point} adapt self-attention networks for point cloud processing.
        Their Point Transformer network and its successor Point Transformer v2 by Wu et al.~\cite{wu2022point} consistently surpasses PointNet++ in segmentation and classification tasks.

        Test-time optimization with autodecoders has gained traction in this field~\cite{parkDeepSDF2019, sitzmann2020metasdf, chabra2020deep} as an alternative to point order invariant encoders.
        In test-time optimization, the encoder is replaced with an optimization of a commutative loss function acting on the points. The commutative property ensures the point order invariance.

        Point Cloud Encoders are also investigated for applications in occupancy learning
        Jia et al.~\cite{jiaLearning2020} successfully utilize an altered PCNN with an additional sampling operator to learn occupancy functions from point clouds.
        The method was applied to improve deformable liver registration by registering to a reconstruction instead of the raw sensor data~\cite{jiaImproving2021}.

\section{Method}

    \begin{figure}[tb]
        \centering
        \includegraphics[width=\columnwidth]{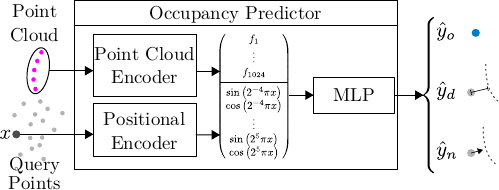}
        \caption{The Occupancy Predictor estimates the occupancy value $\hat{y}_o$, signed distance $\hat{y}_d$ and direction $\hat{y}_n$ to the nearest surface. The Point Cloud is passed to a Point Cloud Encoder to obtain the latent representation $(f_1, \cdots, f_{1024})$. From Query Points a random point $x$ is taken and encoded by a Positional Encoder. The encoded $x$ and the latent representation of the Point Cloud are concatenated and passed to the MLP which estimates $\hat{y}_o$, $\hat{y}_d$, and $\hat{y}_n$.}
        \label{fig:methods_occupancy_predictor}
    \end{figure}

    \subsection{Occupancy Predictor}
    The \gls{op} is an approximator for an occupancy function as illustrated in \cref{fig:occupancy_vs_others}.
    Receiving the position of a query point, the \gls{op} is trained on estimating the occupancy value of the point.
    To train the \gls{op},  pairs of position and occupancy are used.

    By using only pairs of position and occupancy the \gls{op} is limited to learning only a single shape.
    To learn multiple shapes or objects, the \gls{op} requires information about which object to estimate.
    A simple solution is to pass an additional value to the \gls{op} to identify the object by.
    The training data then consists of $(\text{position}, \text{occupancy}, \text{identifier})$ triplets.

    Instead of manually provided identifiers, depth images from a camera may be used for an end-to-end solution.
    A depth image is transformed into a Point Cloud so the \gls{op} does not need to compensate the perspective distortion of the camera.
    We use a \gls{pce} to produce a latent representation that is used as the identifier.
    The \gls{pce} handles varying Point Cloud sizes while also being invariant to point order.

    Sitzmann et al.~\cite{sitzmann2020metasdf} suggest that learning auxiliary tasks can be beneficial to representing objects implicitly.
    Therfore, instead of only estimating the occupancy of a point, the \gls{op} predicts the distance and direction to the nearest surface as an auxiliary task.
    The resulting loss function is:
    \begin{multline}
        Loss = CrossEntropy(Y_o, \hat{Y}_o) + \lambda \cdot L1(Y_d, \hat{Y}_d) \\- MeanCosineSimilarity(Y_{n}, \hat{Y}_{n})
    \end{multline}
    $\lambda = 100$ was chosen heuristically such that all loss components are in the same order of magnitude.
    We found that clamping the distance values as suggested by DeepSDF~\cite{parkDeepSDF2019} did not yield any significant benefits during training.

    The latent representation of the observed Point Cloud is concatenated with the encoded position of a query point to create a query vector.
    The query vector is then passed to an MLP to predict occupancy, distance, and direction to the nearest surface.
    The MLP has $7$ fully-connected layers with $512$ neurons each.
    Batch normalization is applied after every layer, and a skip connection concatenates the query vector with the activations after $4$ layers.

    Neural networks tend to learn low-frequency representations~\cite{mildenhallNeRF2020, tancikFourier2020}.
    We therefore use the positional encoding $\beta$ as proposed in NeRF~\cite{mildenhallNeRF2020} to transform the query point positions into a higher dimensional space
    \begin{multline}
        \beta(x) = \left( \sin( 2^0 \pi x ), \cos( 2^0 \pi x), ..., \right. \\
            \left. \sin( 2^{L - 1} \pi x), \cos( 2^{L - 1} \pi x) \right)
    \end{multline}
    to improve the reconstruction of high-frequency detail.
    Frequency encoding increases sensitivity to small changes in position, as the high-frequency components will still produce a significant signal change.
    For example, compare the L1 distance of original values $|0.07 - 0.09| = 0.02$ with the distance for encoded values $|sin(2^5 \cdot 0.07) - sin(2^5 \cdot 0.09)| \approx 0.53$.

    The inputs to the \gls{op} are normalized based on the sensor data.
    As the sensor data does not encompass the complete object, but all parts of the object will be queried, there will be points outside of the normalization range (see output of Joint Normalizer in \cref{fig:methods_data_generation_overview}).
    Positional encoding uses sinusoidal functions that are $2\pi\text{-peroidic}$.
    To ensure that queried points can be encoded uniquely, we also use negative exponents for $\beta$.
    We chose frequency encoding because it allows this simple modification to encode points outside of the normalization range.

    The architecture of the \gls{op} is illustrated in \cref{fig:methods_occupancy_predictor}.
    All hyperparameters are defined in Appendix B.

    \subsection{Data Generation}
    \begin{figure*}[tbh]
        \centering
        \includegraphics[width=0.90\textwidth]{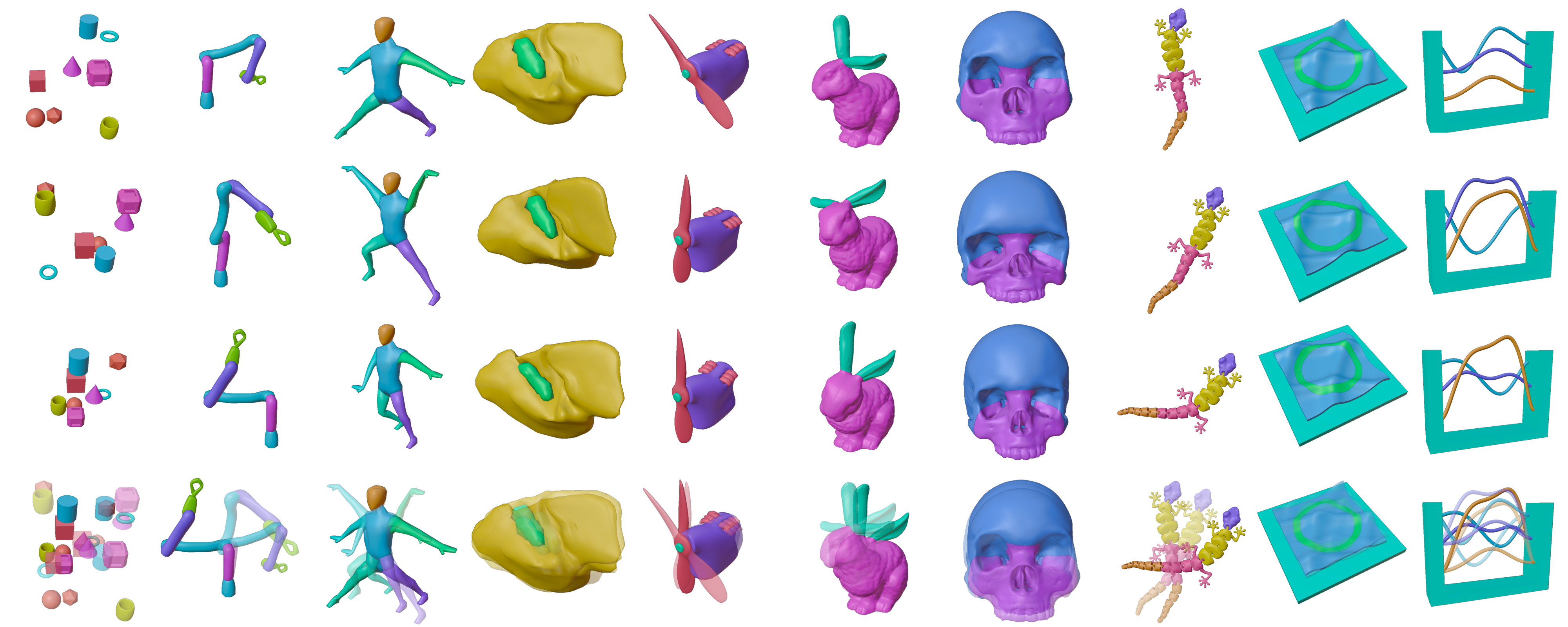}
        \caption{All scenes used to create the training and test data. Each scene contains objects with at least two moving or deforming segments, as indicated by color. In this illustration the camera is fixed. In the last row, the images of the previous rows are overlayed.}
        \label{fig:scenes}
    \end{figure*}

    \begin{figure*}[tbh]
        \centering
        \includegraphics{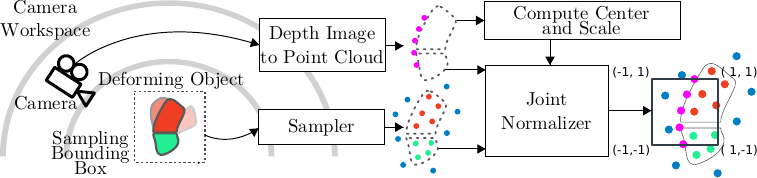}
    
        \caption{2D simplification of the dataset generation process, where different camera and object configurations are captured. For each sample, the Camera is positioned randomly within a spherical workspace around the deformable object. The depth image is transformed into a Point Cloud using the Camera's intrinsic parameters. Simultaneously, the Sampler takes points from the Sampling Bounding Box of the deformable object to create the Query Points. Each query point has a position $(x,y)$, occupancy value $o$ (indicating in which segment the point resides), signed distance $d$ and direction $(n_x,n_y)$ to the nearest surface. Point Cloud and Query Points are then jointly scaled and re-centered to fit inside the square $[-1,1]^2$ that encloses the observed Point Cloud.}
        \label{fig:methods_data_generation_overview}
    \end{figure*}

    We create a synthetic dataset of $10$ deformable objects with a varying number of segments.
    The key features that are represented in the data are 1) high-frequency detail, 2) self-occlusion, and 3) deformations. For a detailed description of the objects, see Appendix B.
    
    Each example in our dataset consists of a Point Cloud and Query Points.
    Examples are created by first randomly sampling a camera perspective and a deformed state of the object.
    Camera perspectives are sampled inside a spherical workspace and are always aimed at the object.
    The camera captures a depth image that is converted into a Point Cloud using the camera's intrinsic parameters.
    Query points are generated with a sampler that scans over the deformed object and transforms them into camera space.
    As both the Point Cloud and the Query Points are in camera space, all reconstructions are in the camera space and therefore already registered to the camera.
    Each query point has a position $(x,y,z)$, an occupancy value $o$, and a distance $d$ and a direction $(n_x, n_y, n_z)$ to the nearest surface.
    As the last step, the Point Cloud and Query Points are jointly centered and scaled.
    The translation $t$ and the scaling factor $s$ are computed using only the Point Cloud.
    Only the Point Cloud will be known at inference time with real-world sensor data.
    With $c$ being the center of the  Point Cloud:
    \begin{multline}
        t = -\frac{1}{2} \left( \max x_i + \min x_i, \max y_i + \min y_i, \cdots \right)\\
        s = \frac{1}{\max\left( |\max x_i - c_x|, |\max y_i - c_y|, \cdots \right)}
    \end{multline}
    An overview of the data generation process is given in \cref{fig:methods_data_generation_overview}.

    \paragraph{Sampling Method for Query Points}
        \label{sec:occupancy_query_gen}
        To improve surface reconstruction quality, it is desirable to concentrate query points near segment boundaries.
        Jia et al.~\cite{jiaLearning2020} achieves this by sampling points on all object surfaces and adding a small random offsets along the surface normal.
        This method requires accurate normals in the model, which are not always available.
        The magnitude of the random offset also introduces an object-specific hyperparameter that needs manual tuning.
        Additionally, this method introduces local density biases in locally convex (over-sampled) and concave (under-sampled) regions.
        
        We propose an algorithm, SortSample, for generating Query Points that benefit the learning of 3D reconstructions of multi-segment objects, regardless of segment thickness or surface area-to-volume ratio.
        SortSample samples points uniformly within each segment's bounding box, extended by $50$\% in all dimensions.
        We separate points inside the segment (added to $S_{\text{inside}}$) from points in empty space (added to $S_{\text{outside}}$), discarding points that are outside the sampled segment but within other segments.
        Once $S_{\text{inside}}$ and $S_{\text{outside}}$ contain at least $n$ points, we sort the lists based on their distance to the segment's surface and select the nearest $k$ points.
        
        Setting $n=k$ typically includes all inside points and outside points up to a distance determined by the geometry of the object.
        Choosing $n=2k$ leads to points that are closer to the surface.
        Using $n=2k$ can increase the computation time needed to generate the dataset noticeably as more samples are drawn and discarded.
        
        SortSample ensures that the generated query points are concentrated around decision boundaries.
        SortSample does not require hyperparameter adjustments based on scale or segment surface area-to-volume ratio and is straightforward to implement.
        While watertight meshes are necessary for occupancy testing, well-behaved normals are not required for SortSample.
        SortSample is neither locally nor globally bias for single segment objects.        
        If there are many segments, there will still be higher point density for smaller objects, introducing a local bias towards them.
        
        In our final dataset, we randomly sample an additional $n_{uniform}$ points within the joint boundaries of all segments to ensure the network learns to classify regions far from the decision boundary.

\section{Experiments}
    \input{plots/reconstructions_table}

    \subsection{Overview}
        We evaluate different Point Cloud Encoders, loss functions, and the impact of positional encoding to find the best performing variant of our system.
        Performance is evaluated by \gls{iou} to quantify reconstruction and \gls{miou} to quantify segmentation quality.
        For Point Cloud Encoders, we compare PointNet++, the autodecoder of DeepSDF, and Point Transformer.
        For this, we extended DeepSDF to perform segmnetation, see Appendix B.
        For the loss function, we investigate the importance of the individual components (CE, L1, and cosine) of our composite loss function.
        We show that PointNet++ as the Point Cloud Encoder, the loss function without the cosine component, and position encoding is the best-performing variant.
        Using our this best-performing variant, we evaluate 1) the reconstruction accuracy on synthetic and real-world objects, 2) the value of SortSampling, 3) the importance of frequency based positional encoding.
        Experiments that investigate the effect of noise on reconstruction accuracy and the metric used for measuring the reconstruction quality can be found in Appendix C and B, respectively.
            
    \subsection{Reconstructions and Segmentation of Synthetic Objects}
        The object reconstructions and segmentations on synthetic data are visualized in \cref{fig:reconstructions}.
        Most objects are reconstructed with an \gls{miou} of between $0.93$ and $0.73$, with the exception of the Rope object, where the relatively high \gls{iou} of the base segment is contrasted by the low \gls{iou} of the rope segments.
        \gls{iou} values are typically close to \gls{miou} values, which indicates that the reconstruction task is more challenging than the segmentation.
        Despite deformations, occlusions, and complex geometries, the system produces highly accurate 3D reconstructions. 

    \subsection{Reconstruction and Segmentation of Real-World Objects}
        The object reconstructions and segmentations on real-world data are visualized in \cref{fig:reconstructions_real_world_small}.
        We test the system's capability to reconstruct real-world objects using the Lizard object (all attributions in Appendix B).
        Lizard was chosen due to its ability to be 3D printed and deformed while also allowing manual registration of its rigid segments.
        We capture Point Clouds of $15$ deformed states with a Zivid One+ (Zivid, Norway) camera.
        Background points that are not part of the lizard are removed and the point density is reduced from approximately $600,000$ to around $600$ points with the Subsampling method of the open-source application CloudCompare.
        For \gls{iou} calculation, we manually register the original 3D model to the full-resolution Point Cloud, aligning each rigid segment individually.
        Our system is trained on synthetic data and tested on real world data, which yielded a \gls{miou} of $0.526$ and an \gls{iou} of $0.530$.
        These scores were considerably lower than the $0.752$ and $0.759$ on synthetic data.
        Although the network accurately managed the overall reconstruction and segmentation, it struggled to precisely reconstruct the segments.

\input{plots/real_world_small_preview}

    \subsection{Variants and Ablations}
        \paragraph{Point Cloud Encoder Architecture and Loss Function}
            \label{sec:ablate:model_loss}
            The \gls{miou} and \gls{iou} of different variants of \glspl{pce} and loss functions are shown in \Cref{fig:ablate:model_loss}.

            \input{plots/reconstruction_barplots}

\input{plots/detail_view}
            
            The factor with the strongest influence on both \gls{miou} and \gls{iou} is the Point Cloud Encoder.
            PointNet++ significantly outperformed both Point Transformer and DeepSDF's autodecoder.
            PointNet++ produces high-frequency details present in the training data.
            The autodecoder of DeepSDF~\cite{parkDeepSDF2019} proved to be sensitive to both the type of object and the hyperparameters, and required considerably more epochs to train ($1000$ vs. $300$).
            
            \Cref{fig:pnpp_vs_transformer} shows that all of these variants are able to infer the overall structure of objects.

            The loss function has a smaller effect on the reconstruction quality.
            For PointNet++, adding the L1 component to the loss increases the \gls{miou} to $0.422$ compared to $0.361$ with CE loss alone for the Rope object.
            The benefit is less pronounced for other objects.
            When using only CE loss in combination with positional encoding, floating periodic artifacts far from the actual reconstruction can be observed.
            CE loss operates on class label probabilities and does not inherently enforce boundedness.
            In contrast, L1 loss penalizes large prediction errors and thus tends to encourage more bounded outputs.
            Our results of adding a cosine distance (CD) term to the loss also align with those of Lamb et al.~\cite{lamb2022deepjoin} that report a small benefit.
            However, we argue that the minor advantage in quality is outweighed by the additional computation and memory requirements.
            When using only the L1 loss, the reconstruction is not segmented.
            DeepSDF~\cite{parkDeepSDF2019}, see DeepSDF with L1 in \Cref{fig:pnpp_vs_transformer}, was outperformed by a large margin by PointNet++ with L1 loss in this task.
            Overall, CE+L1 loss produces more consistent results compared to either CE or L1 loss in isolation.
            In summary, the best variant uses a PointNet++ as Point Cloud Encoder and is trained on the CE+L1 loss with enabled Position Encoding.
        
        \paragraph{Query Point Generation}
            \label{sec:ablate:sampling}
            
            \begin{table}[tbh]
                \renewcommand{\arraystretch}{0.8}
                \caption{Reconstruction error, IoU, and mIoU using $3$ sampling methods. IoU and mIoU are averaged over all examples in the test data. The first row shows \textcolor{blue}{over-reconstruction} and \textcolor{orange}{under-reconstruction} for the Engine object.}
                \label{tab:ablate:sampling}
                \centering
                \setlength{\tabcolsep}{2pt} %
                \newcommand{\doublecell}[1]{\multicolumn{2}{c}{#1}}
                \begin{tabular}{l c c c c c c}
                    \toprule
                     Method \tiny \textrightarrow & \doublecell{Volume Unif.} & \doublecell{Label Unif.} & \doublecell{SortSample} \\
                    \midrule
                    Engine
                    & \doublecell{\includegraphics[width=0.11\textwidth, margin=0pt 2pt 0pt 0pt,valign=m]{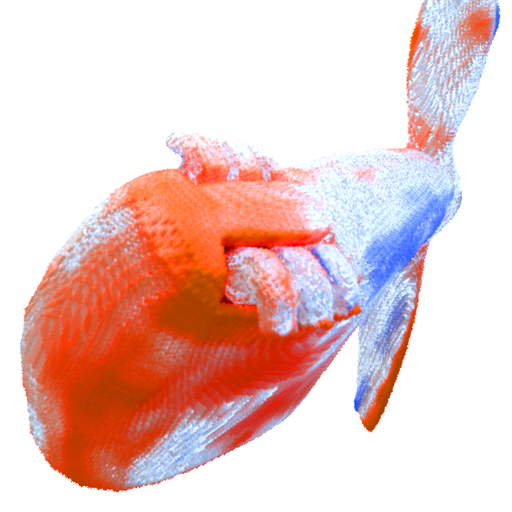}}
                    & \doublecell{\includegraphics[width=0.11\textwidth, margin=0pt 2pt 0pt 0pt,valign=m]{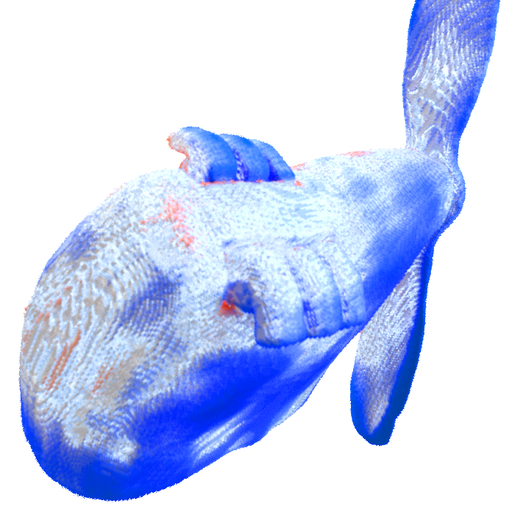}}
                    & \doublecell{\includegraphics[width=0.11\textwidth, margin=0pt 2pt 0pt 0pt,valign=m]{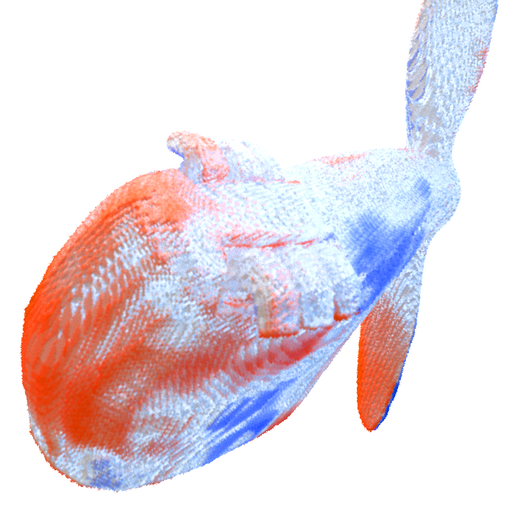}} \\
                    Engine & 0.926 & 0.698 & 0.915 & 0.686 & \textbf{0.931} & \textbf{0.729} \\
                    Human & 0.849 & 0.778 & 0.800 & 0.713 & \textbf{0.894} & \textbf{0.837} \\
                    Lizard & 0.672 & 0.676 & 0.679 & 0.681 & \textbf{0.752} & \textbf{0.749} \\
                    \midrule
                    Object \tiny \textuparrow & IoU & mIoU & IoU & mIoU & IoU & mIoU \\
                    \bottomrule
                \end{tabular}
            \end{table}
            
            We investigate the effect of different sampling methods to generate Query Points.
            \textit{Volume Uniform} sampling draws points uniformly within the extended bounding box of each segment.
            $256$ samples are drawn for each segment.
            \textit{Label Uniform} sampling generates points uniformly inside the segment, and then again uniformly outside the segment within the extended bounding box.
            $256$ samples are drawn for each segment, $128$ samples inside and $128$ samples outside.
           \input{plots/sampling_methods_comparison}

           The different methods are illustrated in \cref{fig:sampling_methods_comparison}.
           In contrast to Volume and Label Uniform, SortSample does not cause local or global biases.
            
            \cref{tab:ablate:sampling} shows \gls{iou} and \gls{miou} for Engine, Human, and Lizard using all $3$ sampling methods.
            SortSample results in datasets that promote improved reconstruction quality.
            The improvement is most noticable for more complex objects such as the Lizard.
            Volume Uniform sampling outperforms Label Uniform sampling as it avoids introducing a local bias in point density at the boundaries.
            The rendered reconstruction shows that Label Uniform sampling results in a \textit{ballooning} effect (over-reconstruction) of structures.

    \paragraph{Positional Encoding}
    The results of using a frequency-based encoding are shown in \cref{tab:ablate:positionalencoding}.
    The use of positional encoding improves the quality of reconstructions by an average of $2$\%.
    The results indicate that objects with more intricate details such as Lizard benefit from the positional encoding, while less detailed objects such as Human and Robot do not.

    \begin{table}[ht]
    \centering
    \caption{mIoU with positional encoding (PE) and without.}
    \label{tab:ablate:positionalencoding}
    \begin{tabular}{lccccc}
        \toprule
        Scene & Shapes & Rope & Human & Lizard & Robot \\
        \toprule
        PE & \textbf{0.791} & \textbf{0.375} & \textbf{0.837} & \textbf{0.749} & \textbf{0.864} \\
        Without & 0.772 & 0.354 & 0.835 & 0.737 & 0.860 \\
        \bottomrule
    \end{tabular}
\end{table}

\section{Discussion and Conclusion}
    In this work, we present a system for registered and segmented 3D reconstruction of deformable objects from single-view point clouds.
    We introduce a simple sampling method for generating Query Points without hyperparameter tuning.
    This sampling method produces suitable training data regardless of the surface-area-to-volume ratio of object segments.
    Furthermore, we present a system for learning segmented objects implicitly through a multi-class occupancy function.
    The system supports using an arbitrary neural network as a point cloud encoder.
    The system is evaluated on synthetic data of a novel suite of $10$ deformable objects and data from a real-world experiment.
    The system is able to reconstruct occluded and high-frequency features.
    Segments are clearly defined and there are only minimal segmentation errors.
    Trained on synthetic data, the system is able to reconstruct real-world objects with some degradation in performance (\gls{miou} of $0.526$ versus $0.752$).
    PointNet++ considerably outperforms Point Transformer and the autodecoder of DeepSDF as Point Cloud Encoders.
    Combining cross-entropy loss (Occupancy) with L1 distance loss (\gls{sdf}) is superior to either loss function in isolation.
    Positional encoding benefits objects with high frequency details, but we argue that the evaluated objects did not contain enough details to make full use of it.
    
    A major limitation of the proposed method is the requirement for watertight meshes for occupancy testing using ray casting.
    This prohibits the use of some publicly available mesh datasets.
    In addition, realistic deformations must be modeled by hand or simulated.
    Future work will address the ability of a single model to generalize across many objects.

    \ifwacvfinal
    \paragraph{Acknowledgements}
        This contribution is supported by the Helmholtz Association under the joint research school "HIDSS4Health – Helmholtz Information and Data Science School for Health" and the Helmholtz Association's Initiative and Networking Fund on the HAICORE@KIT partition.
        We wish to acknowledge Vaisakh Shaj and Jan-Philipp Kaiser from the Karlsruhe Institute of Technology for their support.
    \fi

{\small
\bibliographystyle{ieee_fullname}
\bibliography{main}
}

\ifwacvfinal
\clearpage
\input{appendix}
\fi

\end{document}

%% file: plots/reconstructions_table.tex
\begin{table*}
    \centering
    \renewcommand{\arraystretch}{0.8}
    \begin{tabular}{p{1.8cm}cccccc}
        \toprule
        \textbf{Object} & \textbf{Point Cloud} & \textbf{Reconstruction} & \textbf{Reconst. Error} & \textbf{Segmentation} & \textbf{Seg. Error} & \textbf{Reference} \\ 
        \midrule
        Human
            & \includegraphics[width=0.09\textwidth, margin=0pt 0pt 0pt 0pt,valign=m]{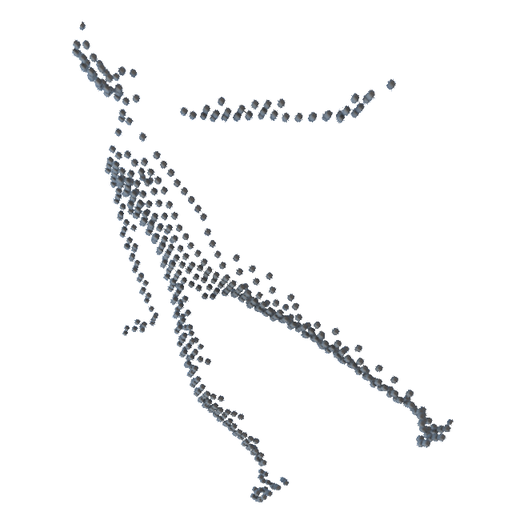}
            & \includegraphics[width=0.09\textwidth, margin=0pt 0pt 0pt 0pt,valign=m]{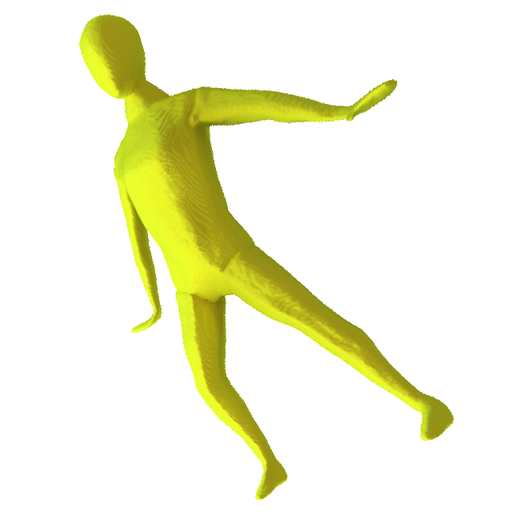}
            & \includegraphics[width=0.09\textwidth, margin=0pt 0pt 0pt 0pt,valign=m]{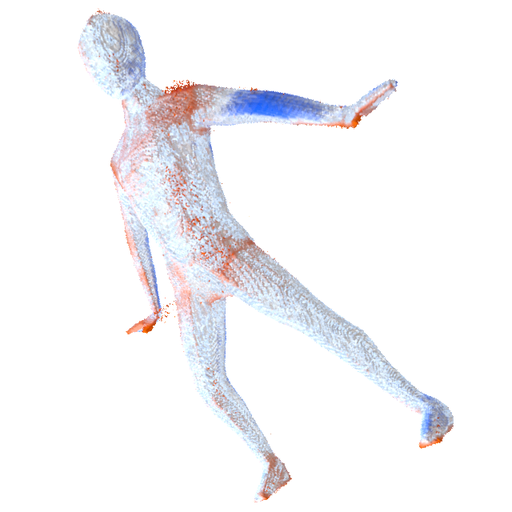}
            & \includegraphics[width=0.09\textwidth, margin=0pt 0pt 0pt 0pt,valign=m]{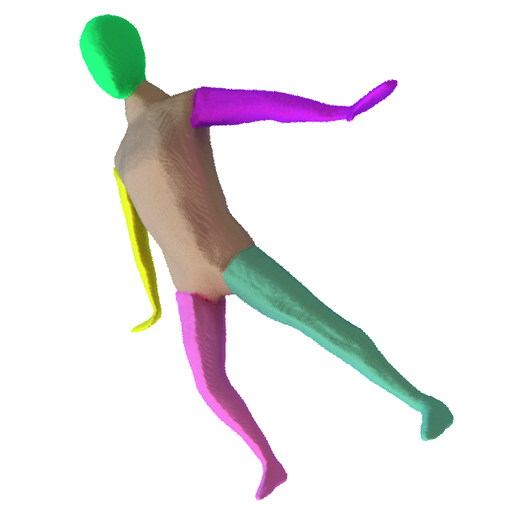}
            & \includegraphics[width=0.09\textwidth, margin=0pt 0pt 0pt 0pt,valign=m]{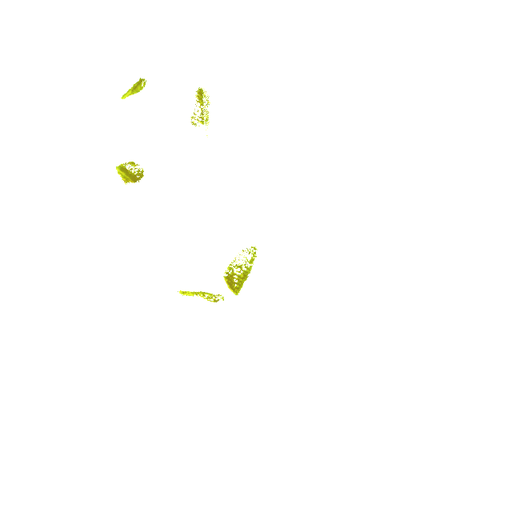}
            & \includegraphics[width=0.09\textwidth, margin=0pt 0pt 0pt 0pt,valign=m]{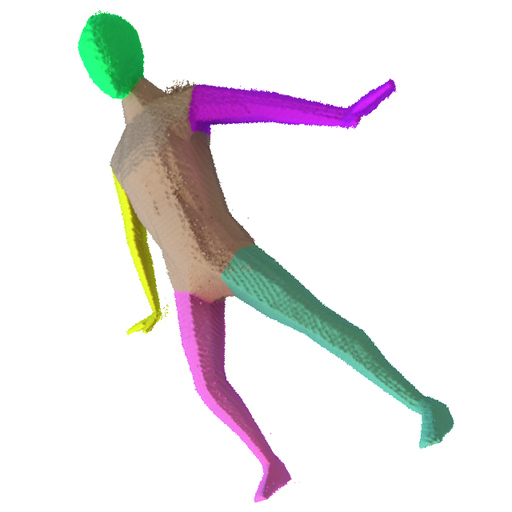} \\
            & & \multicolumn{2}{c}{\small IoU: 0.894} & \multicolumn{2}{c}{\small mIoU: 0.837} & \\
        Robot
            & \includegraphics[width=0.09\textwidth, margin=0pt 0pt 0pt 0pt,valign=m]{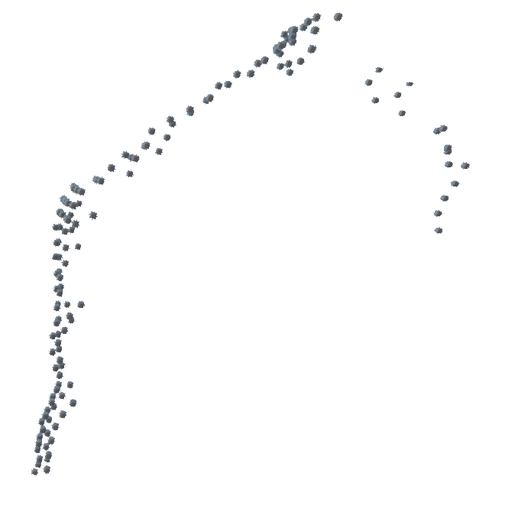}
            & \includegraphics[width=0.09\textwidth, margin=0pt 0pt 0pt 0pt,valign=m]{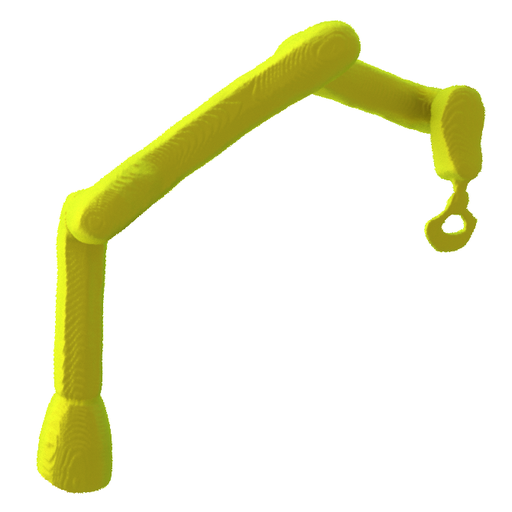}
            & \includegraphics[width=0.09\textwidth, margin=0pt 0pt 0pt 0pt,valign=m]{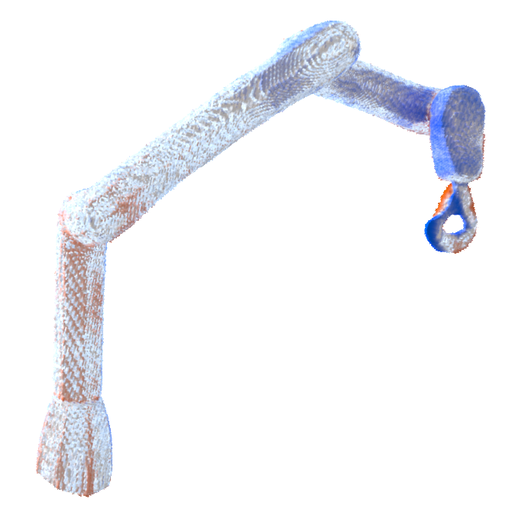}
            & \includegraphics[width=0.09\textwidth, margin=0pt 0pt 0pt 0pt,valign=m]{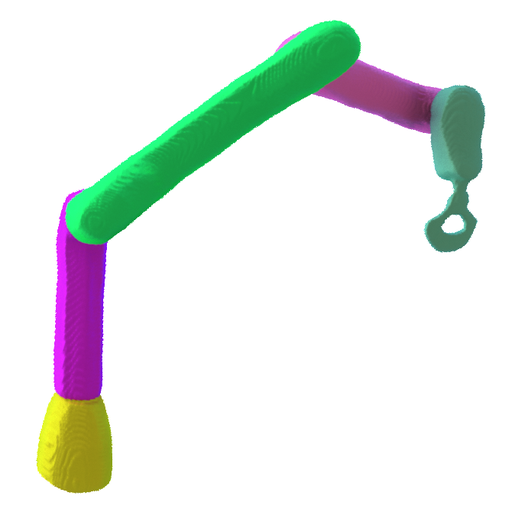}
            & \includegraphics[width=0.09\textwidth, margin=0pt 0pt 0pt 0pt,valign=m]{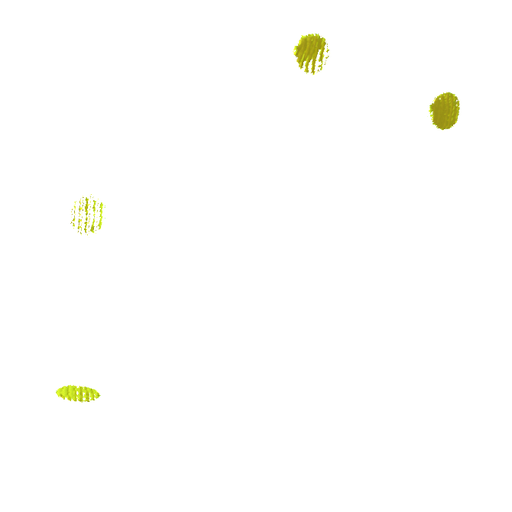}
            & \includegraphics[width=0.09\textwidth, margin=0pt 0pt 0pt 0pt,valign=m]{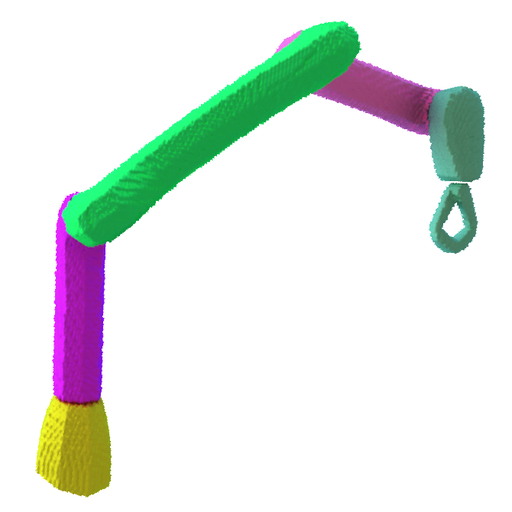} \\
            & & \multicolumn{2}{c}{\small IoU: 0.882} & \multicolumn{2}{c}{\small mIoU: 0.864} & \\
        Stanford Bunny
            & \includegraphics[width=0.09\textwidth, margin=0pt 0pt 0pt 0pt,valign=m]{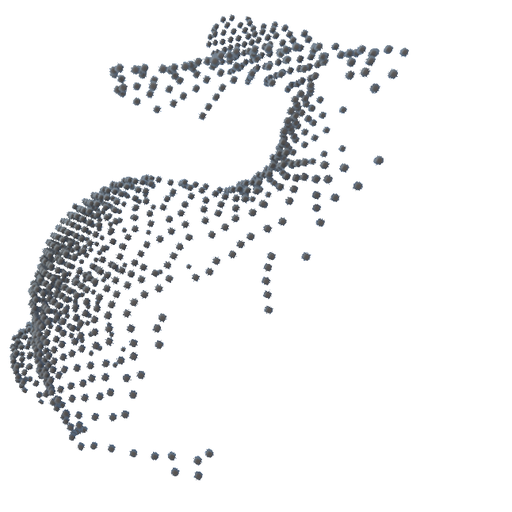}
            & \includegraphics[width=0.09\textwidth, margin=0pt 0pt 0pt 0pt,valign=m]{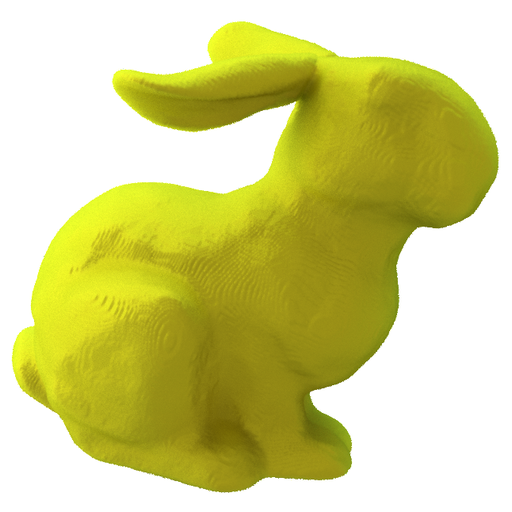}
            & \includegraphics[width=0.09\textwidth, margin=0pt 0pt 0pt 0pt,valign=m]{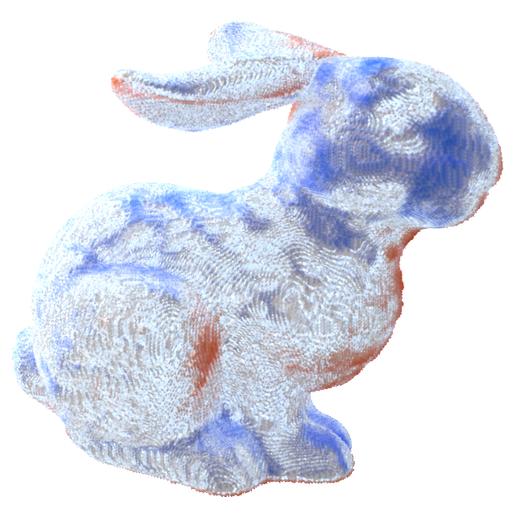}
            & \includegraphics[width=0.09\textwidth, margin=0pt 0pt 0pt 0pt,valign=m]{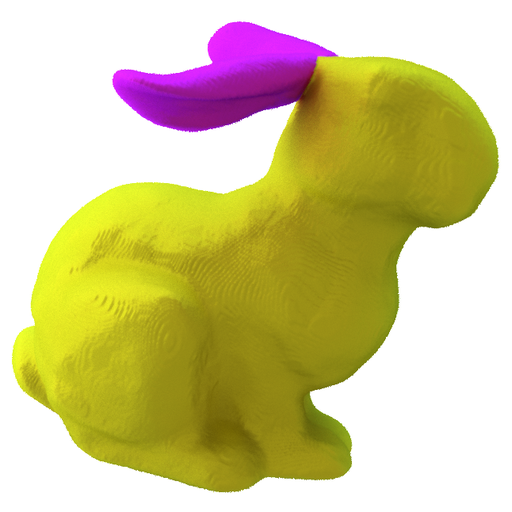}
            & \includegraphics[width=0.09\textwidth, margin=0pt 0pt 0pt 0pt,valign=m]{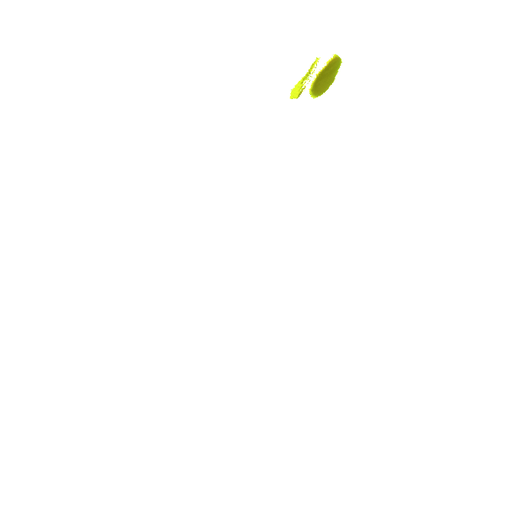}
            & \includegraphics[width=0.09\textwidth, margin=0pt 0pt 0pt 0pt,valign=m]{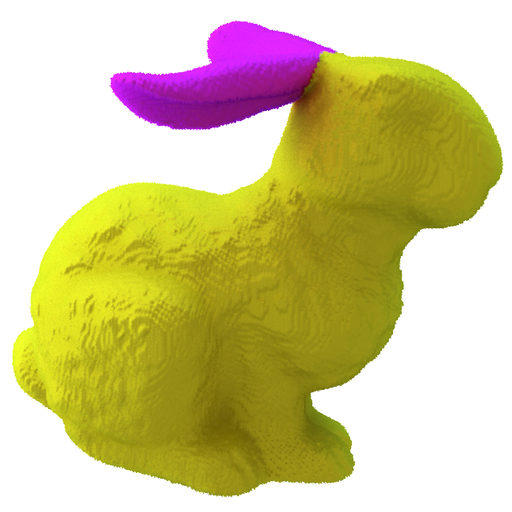} \\
            & & \multicolumn{2}{c}{\small IoU: 0.971} & \multicolumn{2}{c}{\small mIoU: 0.917} & \\
        Lizard
            & \includegraphics[width=0.09\textwidth, margin=0pt 0pt 0pt 0pt,valign=m]{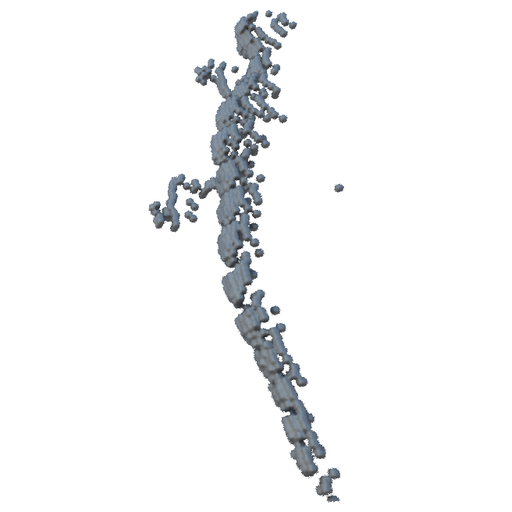}
            & \includegraphics[width=0.09\textwidth, margin=0pt 0pt 0pt 0pt,valign=m]{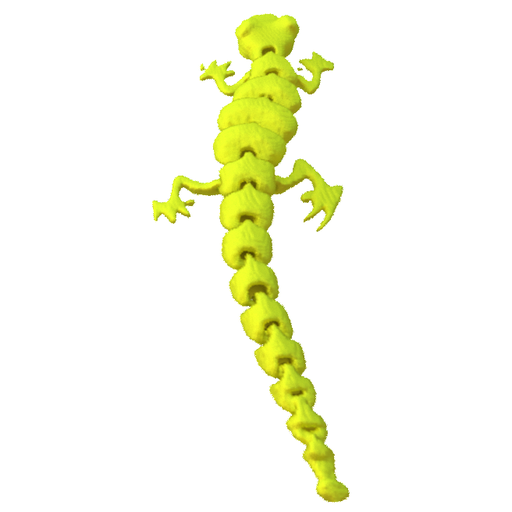}
            & \includegraphics[width=0.09\textwidth, margin=0pt 0pt 0pt 0pt,valign=m]{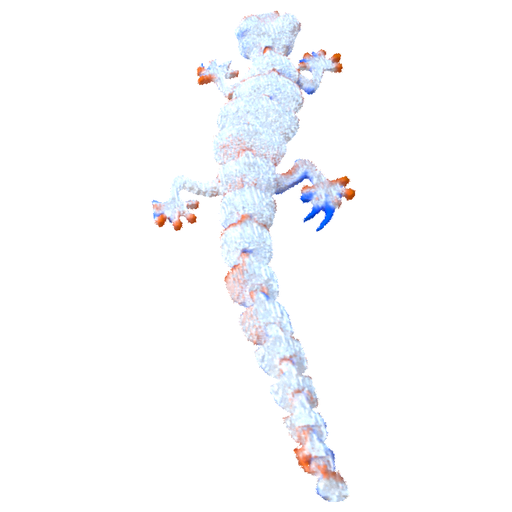}
            & \includegraphics[width=0.09\textwidth, margin=0pt 0pt 0pt 0pt,valign=m]{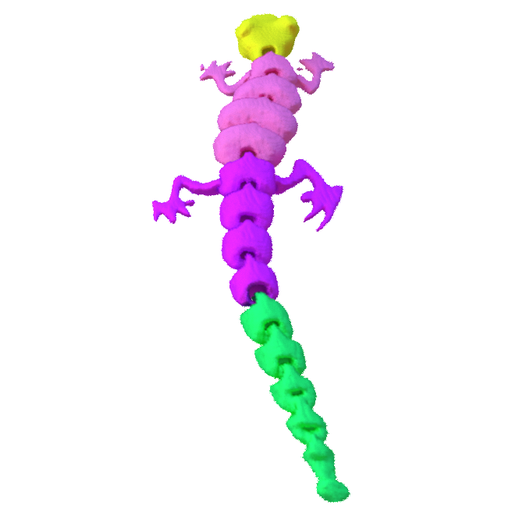}
            & \includegraphics[width=0.09\textwidth, margin=0pt 0pt 0pt 0pt,valign=m]{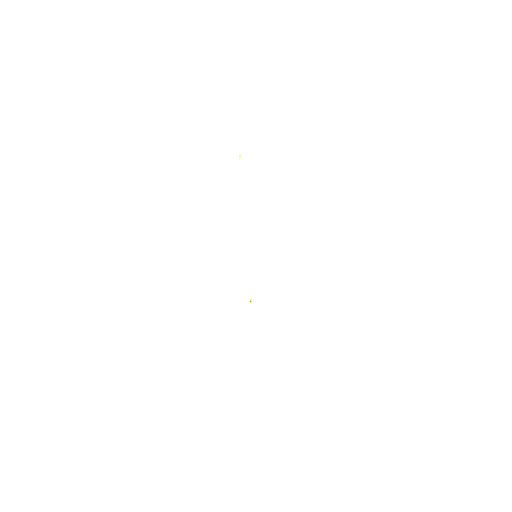}
            & \includegraphics[width=0.09\textwidth, margin=0pt 0pt 0pt 0pt,valign=m]{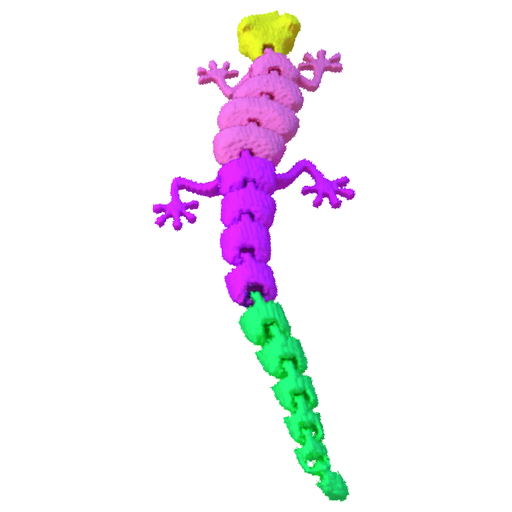} \\
            & & \multicolumn{2}{c}{\small IoU: 0.752} & \multicolumn{2}{c}{\small mIoU: 0.749} & \\
        Rope
            & \includegraphics[width=0.09\textwidth, margin=0pt 0pt 0pt 0pt,valign=m]{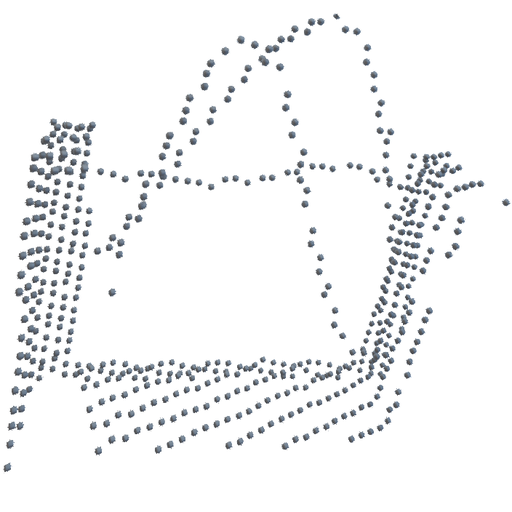}
            & \includegraphics[width=0.09\textwidth, margin=0pt 0pt 0pt 0pt,valign=m]{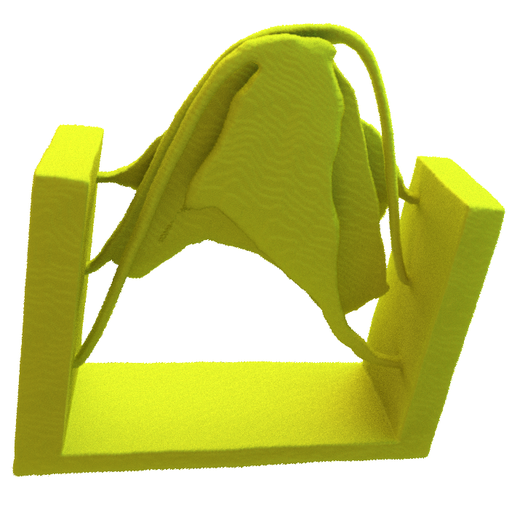}
            & \includegraphics[width=0.09\textwidth, margin=0pt 0pt 0pt 0pt,valign=m]{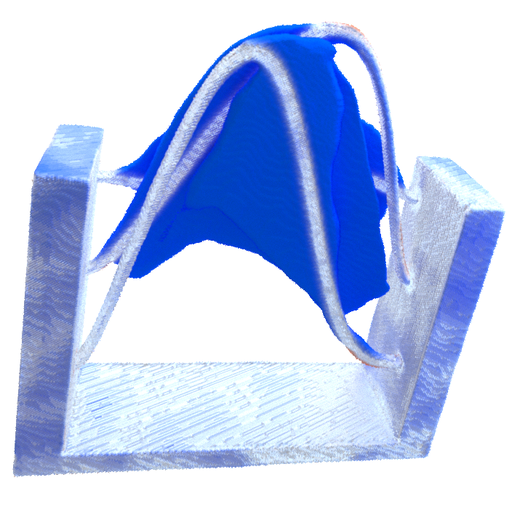}
            & \includegraphics[width=0.09\textwidth, margin=0pt 0pt 0pt 0pt,valign=m]{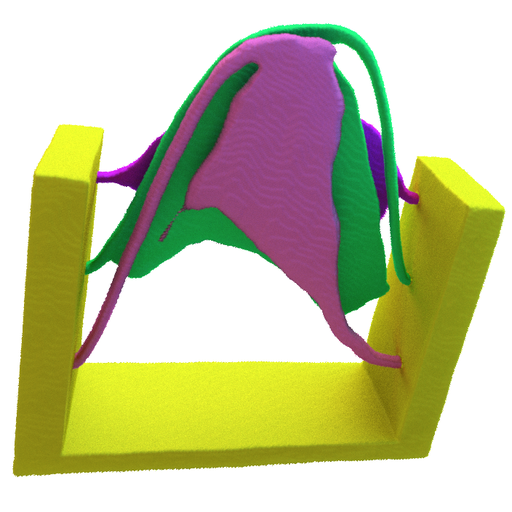}
            & \includegraphics[width=0.09\textwidth, margin=0pt 0pt 0pt 0pt,valign=m]{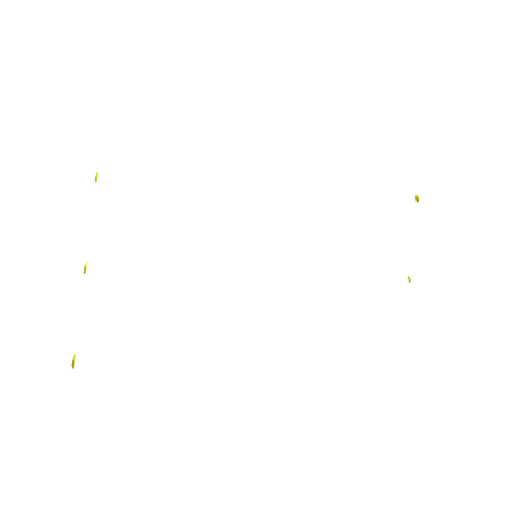}
            & \includegraphics[width=0.09\textwidth, margin=0pt 0pt 0pt 0pt,valign=m]{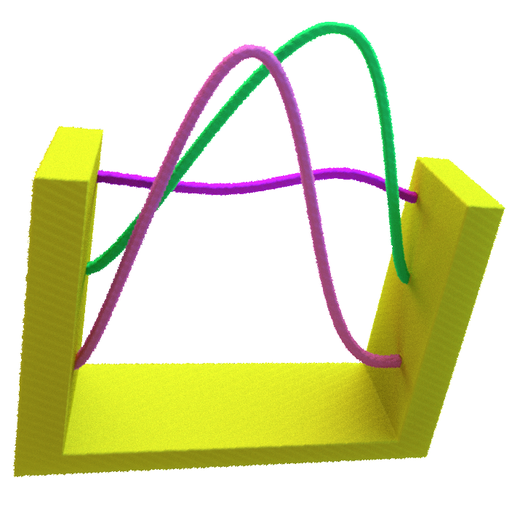} \\
            & & \multicolumn{2}{c}{\small IoU: 0.815} & \multicolumn{2}{c}{\small \small mIoU: 0.375} & \\
        \bottomrule
    \end{tabular}
    \captionof{figure}{
        Examples of reconstructions generated by conditioning on the input \textbf{Point Cloud}.
        The \textbf{Reconstruction} is the sum of all points that are not classified as empty space.
        \textbf{Reconstruction Error} identifies \textcolor{blue}{over-reconstruction} and \textcolor{orange}{under-reconstruction} when compared with the reference.
        \textbf{Segmentation} colors each predicted class. \textbf{Segmentation Error} between predicted \textbf{Segmentation} and \textbf{Reference}.
        IoU and mIoU values are averaged over all examples in the test data, not just the rendered examples.
        Remaining objects are listed in Appendix A.
    }
    \label{fig:reconstructions}
\end{table*}

%% file: plots/real_world_small_preview.tex
\begin{table}[tbh]

    \centering
    \begin{tabular}{cccc}
        \multicolumn{4}{c}{\includegraphics[width=0.4\textwidth]{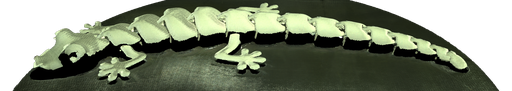}} \\
        \textbf{PC} & \textbf{Segmentation} & \textbf{Reference} & \textbf{Re. Error}\\ 
            \rotatebox{180}{\includegraphics[width=0.09\textwidth, margin=0pt 0pt 0pt 0pt,valign=m]{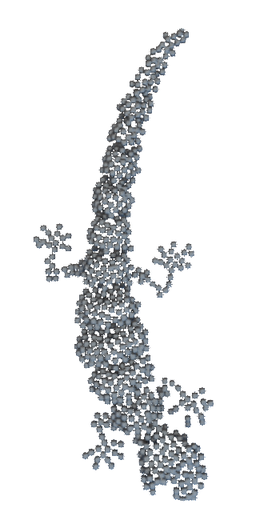}}
            & \rotatebox{180}{\includegraphics[width=0.09\textwidth, margin=0pt 0pt 0pt 0pt,valign=m]{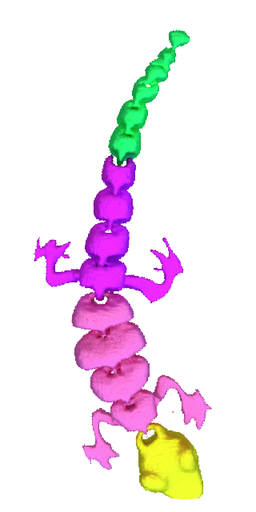}}
            & \rotatebox{180}{\includegraphics[width=0.09\textwidth, margin=0pt 0pt 0pt 0pt,valign=m]{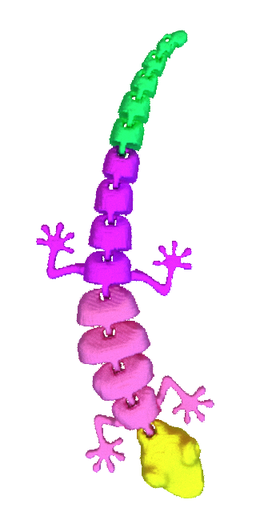}}
            & \rotatebox{180}{\includegraphics[width=0.09\textwidth, margin=0pt 0pt 0pt 0pt,valign=m]{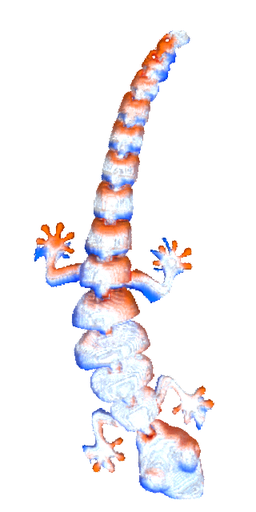}}\\ 
    \end{tabular}
    \captionof{figure}{
    Example of a reconstruction generated by using a sub-sampled version \textbf{PC} of a real-world point cloud (top) as input to the network trained on synthetic data. The lizard is approximately 30 cm long. \textbf{Re. Error} identifies \textcolor{blue}{over-reconstruction} and \textcolor{orange}{under-reconstruction} when compared with the reference. All reconstructions are shown in Appendix D.
    }
    \label{fig:reconstructions_real_world_small}
\end{table}

%% file: plots/reconstruction_barplots.tex
\definecolor{CE}{RGB}{230, 159, 0}
\definecolor{L1}{RGB}{49, 104, 142}
\definecolor{CEL1}{RGB}{53, 183, 121}
\definecolor{CEL1CD}{RGB}{68, 1, 84}
\pgfplotsset{
    legend image blank/.style={
        legend image code/.code={%
            \draw [#1] (0cm, -0.1cm) rectangle (0.15cm, 0.2cm);
        }
    },
}
\pgfplotsset{
    legend image pattern/.style={
        legend image code/.code={%
            \draw [#1, pattern=north west lines] (0cm, -0.1cm) rectangle (0.15cm, 0.2cm);
        }
    },
}
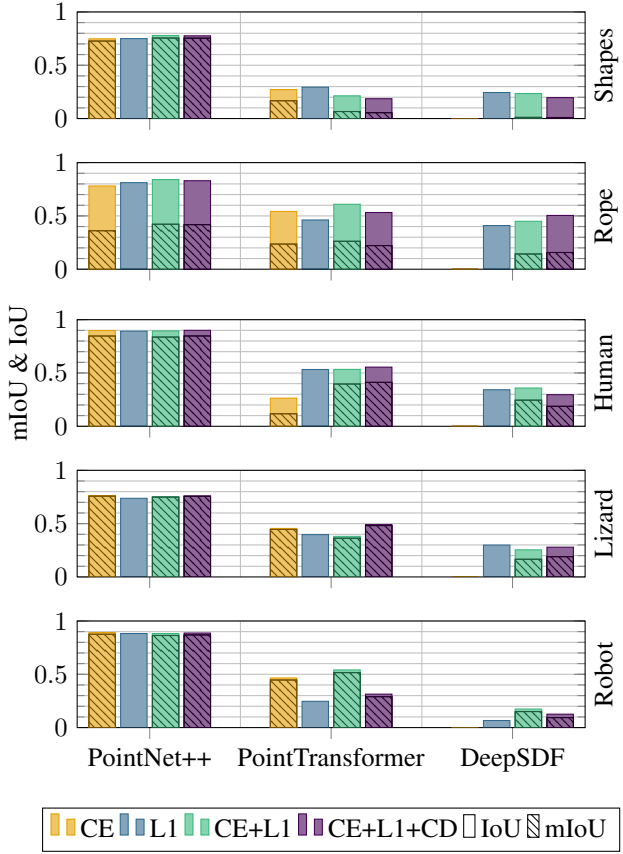
\begin{figure}[tbh]
    \ifnographics
    \else
    \centering
    \begin{tikzpicture}
        \begin{axis}[
            ybar,
            name=sceneRobot,
            height=3cm,
            width=\columnwidth,
            enlarge x limits=0.2,
            legend style={at={(0.5,-0.75)}, anchor=north,legend columns=-1},
            symbolic x coords={PointNet++, PointTransformer, DeepSDF},
            xtick=data,
            minor x tick num=1,
            xminorgrids,
            yminorgrids,
            xtick pos=left,
            ytick distance={0.5},
            ymax=1.0,
            ymin=0.0,
            minor ytick={0.1, 0.2, ..., 1.0}, 
            minor x tick style = {opacity=0},
            ]
            \addplot[draw=CE, fill=CE!60] coordinates {(PointNet++, 0.8919) (PointTransformer, 0.4655) (DeepSDF, 0.0)}; \addlegendentry{CE}
            \addplot[draw=L1, fill=L1!60] coordinates {(PointNet++, 0.8826) (PointTransformer, 0.2466) (DeepSDF, 0.06655)}; \addlegendentry{L1}
            \addplot[draw=CEL1, fill=CEL1!60] coordinates {(PointNet++, 0.8823) (PointTransformer, 0.5408) (DeepSDF, 0.1739)}; \addlegendentry{CE+L1}
            \addplot[draw=CEL1CD, fill=CEL1CD!60] coordinates {(PointNet++, 0.887) (PointTransformer, 0.3139) (DeepSDF, 0.1259)}; \addlegendentry{CE+L1+CD}
            \addlegendimage{legend image blank}
            \addlegendentry{IoU}
            \addlegendimage{legend image pattern}
            \addlegendentry{mIoU}
        \end{axis}
        
        \begin{axis}[
            ybar,
            at=(sceneRobot.center),
            anchor=center,
            name=sceneRobotb,
            height=3cm,
            width=\columnwidth,
            enlarge x limits=0.2,
            symbolic x coords={PointNet++, PointTransformer, DeepSDF},
            ymax=1.0,
            ymin=0.0,
            axis line style={draw=none},
            tick style={draw=none},
            xticklabel style={draw=none},
            yticklabel style={draw=none},
            yticklabels={,,},
            xticklabels={,,},
            ]
            \addplot[draw=CE!50!black, fill=CE!60, postaction={pattern=north west lines, pattern color=CE!50!black}] coordinates
            {(PointNet++, 0.8755) (PointTransformer, 0.4462) (DeepSDF, 0.0)};
            
            \addplot[draw=L1!50!black, fill=L1!60, postaction={pattern=north west lines, pattern color=L1!50!black}] coordinates
            {(PointNet++, 0.0) (PointTransformer, 0.0) (DeepSDF, 0.0)};
            
            \addplot[draw=CEL1!50!black, fill=CEL1!60, postaction={pattern=north west lines, pattern color=CEL1!50!black}] coordinates
            {(PointNet++, 0.864) (PointTransformer, 0.5156) (DeepSDF, 0.1504)};
            
            \addplot[draw=CEL1CD!50!black, fill=CEL1CD!60, postaction={pattern=north west lines, pattern color=CEL1CD!50!black}] coordinates 
            {(PointNet++, 0.8691) (PointTransformer, 0.2901) (DeepSDF, 0.09389)};
        \end{axis}
        
        \begin{axis}[
            ybar,
            name=sceneLizard,
            at=(sceneRobot.above north west),
            anchor=below south west,
            yshift=-1em,
            height=3cm,
            width=\columnwidth,
            enlarge x limits=0.2,
            symbolic x coords={PointNet++, PointTransformer, DeepSDF},
            xtick=data,
            minor x tick num=1,
            xminorgrids,
            yminorgrids,
            xtick pos=left,
            ytick distance={0.5},
            ymax=1.0,
            ymin=0.0,
            minor ytick={0.1, 0.2, ..., 1.0}, 
            xticklabel style = {opacity=0},
            minor x tick style = {opacity=0},
            ]
            \addplot[draw=CE, fill=CE!60] coordinates {(PointNet++, 0.7635) (PointTransformer, 0.4552) (DeepSDF, 0.0)};
            \addplot[draw=L1, fill=L1!60] coordinates {(PointNet++, 0.7376) (PointTransformer, 0.3969) (DeepSDF, 0.2986)};
            \addplot[draw=CEL1, fill=CEL1!60] coordinates {(PointNet++, 0.7522) (PointTransformer, 0.3809) (DeepSDF, 0.2542)};
            \addplot[draw=CEL1CD, fill=CEL1CD!60] coordinates {(PointNet++, 0.7598) (PointTransformer, 0.4912) (DeepSDF, 0.2793)};
        \end{axis}
        
        \begin{axis}[
            ybar,
            at=(sceneLizard.center),
            anchor=center,
            name=sceneLizardb,
            height=3cm,
            width=\columnwidth,
            enlarge x limits=0.2,
            symbolic x coords={PointNet++, PointTransformer, DeepSDF},
            ymax=1.0,
            ymin=0.0,
            axis line style={draw=none},
            tick style={draw=none},
            xticklabel style={draw=none},
            yticklabel style={draw=none},
            yticklabels={,,},
            xticklabels={,,},
            ]
            \addplot[draw=CE!50!black, fill=CE!60, postaction={pattern=north west lines, pattern color=CE!50!black}] coordinates
            {(PointNet++, 0.7577) (PointTransformer, 0.4455) (DeepSDF, 0.0)};
            
            \addplot[draw=L1!50!black, fill=L1!60, postaction={pattern=north west lines, pattern color=L1!50!black}] coordinates
            {(PointNet++, 0.0) (PointTransformer, 0.0) (DeepSDF, 0.0)};
            
            \addplot[draw=CEL1!50!black, fill=CEL1!60, postaction={pattern=north west lines, pattern color=CEL1!50!black}] coordinates
            {(PointNet++, 0.7487) (PointTransformer, 0.3634) (DeepSDF, 0.1658)};
            
            \addplot[draw=CEL1CD!50!black, fill=CEL1CD!60, postaction={pattern=north west lines, pattern color=CEL1CD!50!black}] coordinates 
            {(PointNet++, 0.7563) (PointTransformer, 0.4818) (DeepSDF, 0.1901)};
        \end{axis}
        
        \begin{axis}[
            ybar,
            name=sceneHuman,
            at=(sceneLizard.above north west),
            anchor=below south west,
            yshift=-1em,
            height=3cm,
            width=\columnwidth,
            enlarge x limits=0.2,
            symbolic x coords={PointNet++, PointTransformer, DeepSDF},
            xtick=data,
            minor x tick num=1,
            xminorgrids,
            yminorgrids,
            xtick pos=left,
            ytick distance={0.5},
            ymax=1.0,
            ymin=0.0,
            minor ytick={0.1, 0.2, ..., 1.0}, 
            xticklabel style = {opacity=0},
            minor x tick style = {opacity=0},
            y label style={at={(axis description cs:-0.08,0.5)},anchor=south},
            ylabel={mIoU \& IoU},
            ]
            \addplot[draw=CE, fill=CE!60] coordinates {(PointNet++, 0.8986) (PointTransformer, 0.264) (DeepSDF, 0.0)};
            \addplot[draw=L1, fill=L1!60] coordinates {(PointNet++, 0.892) (PointTransformer, 0.5328) (DeepSDF, 0.3427)};
            \addplot[draw=CEL1, fill=CEL1!60] coordinates {(PointNet++, 0.8942) (PointTransformer, 0.5338) (DeepSDF, 0.359)};
            \addplot[draw=CEL1CD, fill=CEL1CD!60] coordinates {(PointNet++, 0.9002) (PointTransformer, 0.555) (DeepSDF, 0.2964)};
        \end{axis}
        
        \begin{axis}[
            ybar,
            at=(sceneHuman.center),
            anchor=center,
            name=sceneHumanb,
            height=3cm,
            width=\columnwidth,
            enlarge x limits=0.2,
            symbolic x coords={PointNet++, PointTransformer, DeepSDF},
            ymax=1.0,
            ymin=0.0,
            axis line style={draw=none},
            tick style={draw=none},
            xticklabel style={draw=none},
            yticklabel style={draw=none},
            yticklabels={,,},
            xticklabels={,,},
            ]
            \addplot[draw=CE!50!black, fill=CE!60, postaction={pattern=north west lines, pattern color=CE!50!black}] coordinates
            {(PointNet++, 0.8467) (PointTransformer, 0.1173) (DeepSDF, 0.0)};
            
            \addplot[draw=L1!50!black, fill=L1!60, postaction={pattern=north west lines, pattern color=L1!50!black}] coordinates
            {(PointNet++, 0.0) (PointTransformer, 0.0) (DeepSDF, 0.0)};
            
            \addplot[draw=CEL1!50!black, fill=CEL1!60, postaction={pattern=north west lines, pattern color=CEL1!50!black}] coordinates
            {(PointNet++, 0.8368) (PointTransformer, 0.3967) (DeepSDF, 0.245)};
            
            \addplot[draw=CEL1CD!50!black, fill=CEL1CD!60, postaction={pattern=north west lines, pattern color=CEL1CD!50!black}] coordinates 
            {(PointNet++, 0.8473) (PointTransformer, 0.412) (DeepSDF, 0.1875)};
        \end{axis}

        \begin{axis}[
            ybar,
            name=sceneRope,
            at=(sceneHuman.above north west),
            anchor=below south west,
            yshift=-1em,
            height=3cm,
            width=\columnwidth,
            enlarge x limits=0.2,
            symbolic x coords={PointNet++, PointTransformer, DeepSDF},
            xtick=data,
            minor x tick num=1,
            xminorgrids,
            yminorgrids,
            xtick pos=left,
            ytick distance={0.5},
            ymax=1.0,
            ymin=0.0,
            minor ytick={0.1, 0.2, ..., 1.0}, 
            xticklabel style = {opacity=0},
            minor x tick style = {opacity=0},
            ]
            \addplot[draw=CE, fill=CE!60] coordinates {(PointNet++, 0.781) (PointTransformer, 0.542) (DeepSDF, 0.0)};
            \addplot[draw=L1, fill=L1!60] coordinates {(PointNet++, 0.812) (PointTransformer, 0.462) (DeepSDF, 0.4095)};
            \addplot[draw=CEL1, fill=CEL1!60] coordinates {(PointNet++, 0.841) (PointTransformer, 0.609) (DeepSDF, 0.4496)};
            \addplot[draw=CEL1CD, fill=CEL1CD!60] coordinates {(PointNet++, 0.830) (PointTransformer, 0.532) (DeepSDF, 0.5045)};
        \end{axis}
        
        \begin{axis}[
            ybar,
            at=(sceneRope.center),
            anchor=center,
            name=sceneRopeb,
            height=3cm,
            width=\columnwidth,
            enlarge x limits=0.2,
            symbolic x coords={PointNet++, PointTransformer, DeepSDF},
            ymax=1.0,
            ymin=0.0,
            axis line style={draw=none},
            tick style={draw=none},
            xticklabel style={draw=none},
            yticklabel style={draw=none},
            yticklabels={,,},
            xticklabels={,,},
            ]
            \addplot[draw=CE!50!black, fill=CE!60, postaction={pattern=north west lines, pattern color=CE!50!black}] coordinates
            {(PointNet++, 0.361) (PointTransformer, 0.236) (DeepSDF, 0.0)};
            
            \addplot[draw=L1!50!black, fill=L1!60, postaction={pattern=north west lines, pattern color=L1!50!black}] coordinates
            {(PointNet++, 0.0) (PointTransformer, 0.0) (DeepSDF, 0.0)};
            
            \addplot[draw=CEL1!50!black, fill=CEL1!60, postaction={pattern=north west lines, pattern color=CEL1!50!black}] coordinates
            {(PointNet++, 0.422) (PointTransformer, 0.262) (DeepSDF, 0.1429)};
            
            \addplot[draw=CEL1CD!50!black, fill=CEL1CD!60, postaction={pattern=north west lines, pattern color=CEL1CD!50!black}] coordinates 
            {(PointNet++, 0.418) (PointTransformer, 0.222) (DeepSDF, 0.1569)};
        \end{axis}

        \begin{axis}[
            ybar,
            name=sceneShapes,
            at=(sceneRope.above north west),
            anchor=below south west,
            yshift=-1em,
            height=3cm,
            width=\columnwidth,
            enlarge x limits=0.2,
            symbolic x coords={PointNet++, PointTransformer, DeepSDF},
            xtick=data,
            minor x tick num=1,
            xminorgrids,
            yminorgrids,
            xtick pos=left,
            ytick distance={0.5},
            ymax=1.0,
            ymin=0.0,
            minor ytick={0.1, 0.2, ..., 1.0}, 
            xticklabel style = {opacity=0},
            minor x tick style = {opacity=0},
            ]
            \addplot[draw=CE, fill=CE!60] coordinates {(PointNet++, 0.748) (PointTransformer, 0.272) (DeepSDF, 0.0)};
            \addplot[draw=L1, fill=L1!60] coordinates {(PointNet++, 0.750) (PointTransformer, 0.294) (DeepSDF, 0.2443)};
            \addplot[draw=CEL1, fill=CEL1!60] coordinates {(PointNet++, 0.779) (PointTransformer, 0.213) (DeepSDF, 0.2348)};
            \addplot[draw=CEL1CD, fill=CEL1CD!60] coordinates {(PointNet++, 0.776) (PointTransformer, 0.186) (DeepSDF, 0.1963)};
        \end{axis}
        
        \begin{axis}[
            ybar,
            at=(sceneShapes.center),
            anchor=center,
            name=sceneShapesb,
            height=3cm,
            width=\columnwidth,
            enlarge x limits=0.2,
            symbolic x coords={PointNet++, PointTransformer, DeepSDF},
            ymax=1.0,
            ymin=0.0,
            axis line style={draw=none},
            tick style={draw=none},
            xticklabel style={draw=none},
            yticklabel style={draw=none},
            yticklabels={,,},
            xticklabels={,,},
            ]
            \addplot[draw=CE!50!black, fill=CE!60, postaction={pattern=north west lines, pattern color=CE!50!black}] coordinates
            {(PointNet++, 0.727) (PointTransformer, 0.167) (DeepSDF, 0.0)};
            
            \addplot[draw=L1!50!black, fill=L1!60, postaction={pattern=north west lines, pattern color=L1!50!black}] coordinates
            {(PointNet++, 0.0) (PointTransformer, 0.0) (DeepSDF, 0.0)};
            
            \addplot[draw=CEL1!50!black, fill=CEL1!60, postaction={pattern=north west lines, pattern color=CEL1!50!black}] coordinates
            {(PointNet++, 0.755) (PointTransformer, 0.066) (DeepSDF, 0.01113)};
            
            \addplot[draw=CEL1CD!50!black, fill=CEL1CD!60, postaction={pattern=north west lines, pattern color=CEL1CD!50!black}] coordinates 
            {(PointNet++, 0.754) (PointTransformer, 0.055) (DeepSDF, 0.009942)};
        \end{axis}

        \node [anchor = north, rotate=90] at (sceneRobot.east) {Robot};
        \node [anchor = north, rotate=90] at (sceneLizard.east) {Lizard};
        \node [anchor = north, rotate=90] at (sceneHuman.east) {Human};
        \node [anchor = north, rotate=90] at (sceneRope.east) {Rope};
        \node [anchor = north, rotate=90] at (sceneShapes.east) {Shapes};
    
    \end{tikzpicture}
    \fi
    \caption{
        IoU and mIoU for each model architecture and loss function on selected scenes. It is not possible to train the DeepSDF with only CE loss, so these values are omitted. The L1 loss does not optimize class probabilities required for mIoU, so these values are also omitted.
    }
    \label{fig:ablate:model_loss}
\end{figure}

%% file: plots/detail_view.tex
\begin{figure}
    \centering
    \begin{tikzpicture}[align=center]
        \node (lizard PN) {\includegraphics[width=0.3\columnwidth]{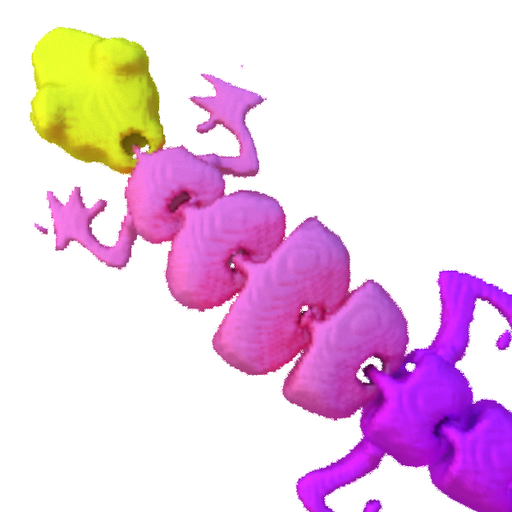}};
        \node[anchor=west, right = 0 of lizard PN] (lizard TF) {\includegraphics[width=0.3\columnwidth]{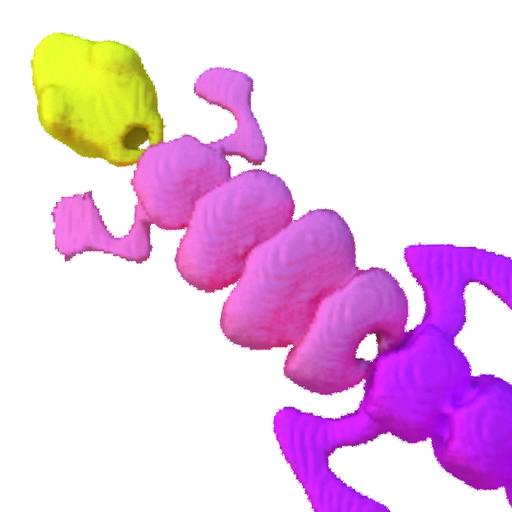}};
        \node[anchor=west, right = 0 of lizard TF] (lizard DSDF) {\includegraphics[width=0.3\columnwidth]{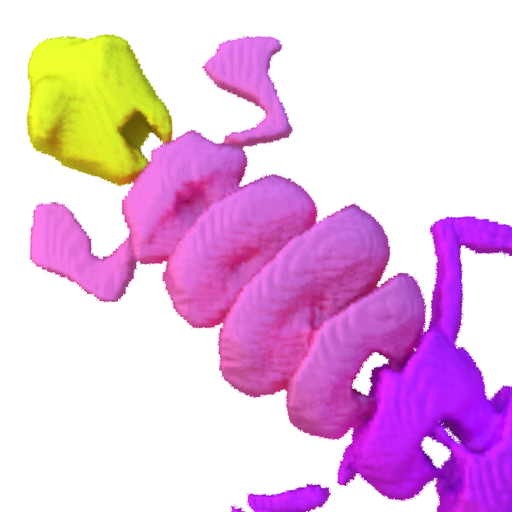}};
        
        \node[anchor=south west, align=center] at ($(lizard PN.south west) + (0.1, 0.1)$) {mIoU: \\ 0.749};
        \node[anchor=south west, align=center] at ($(lizard TF.south west) + (0.1, 0.1)$) {mIoU: \\ 0.363};
        \node[anchor=south west, align=center] at ($(lizard DSDF.south west) + (0.1, 0.1)$) {mIoU: \\ 0.166};

        \node[anchor=south, above = -0.3cm of lizard PN] {\bf PointNet++\strut};
        \node[anchor=south, above = -0.3cm of lizard TF] {\bf PointTransformer\strut};
        \node[anchor=south, above = -0.3cm of lizard DSDF] {\bf DeepSDF\strut};

        \node[anchor=north, below = 0 of lizard PN] (human PN) {\includegraphics[width=0.3\columnwidth]{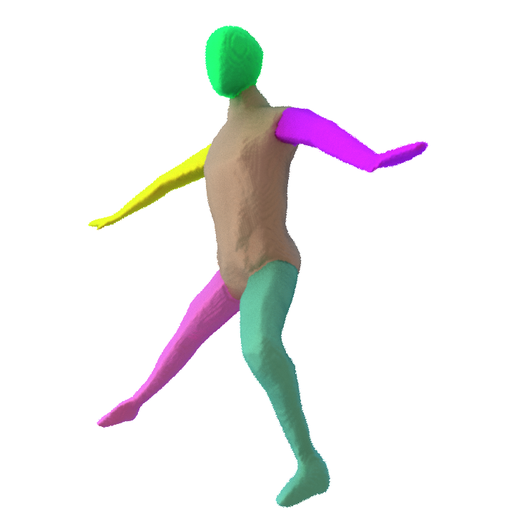}};
        \node[anchor=west, right = 0 of human PN] (human TF) {\includegraphics[width=0.3\columnwidth]{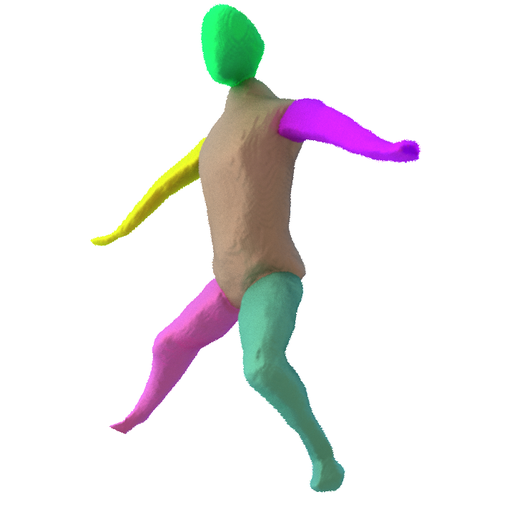}};
        \node[anchor=west, right = 0 of human TF] (human DSDF) {\includegraphics[width=0.3\columnwidth]{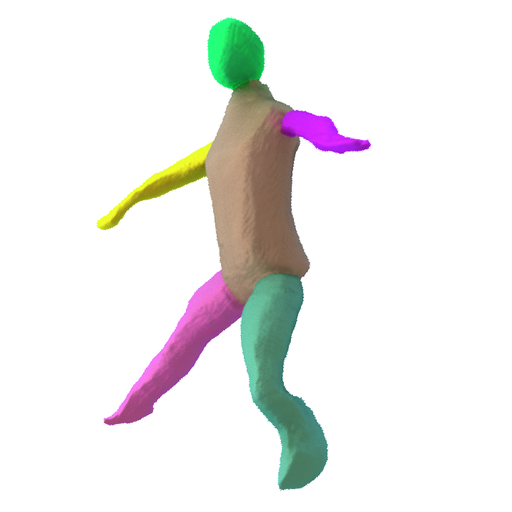}};

        \node[anchor=south west, align=center] at ($(human PN.south west) + (0.1, 0.1)$) {0.837};
        \node[anchor=south west, align=center] at ($(human TF.south west) + (0.1, 0.1)$) {0.397};
        \node[anchor=south west, align=center] at ($(human DSDF.south west) + (0.1, 0.1)$) {0.245};
    \end{tikzpicture}
    \caption{
        Comparison of reconstructions of the Lizard (top row) and Human (bottom row) scenes provided by different models trained with CE+L1 loss.
        The significant difference in mIoU between the models is due to the high-frequency details, which PointNet++ is better able to capture.
        All models are able to infer the overall structure of the scenes.
    }
    \label{fig:pnpp_vs_transformer}
\end{figure}
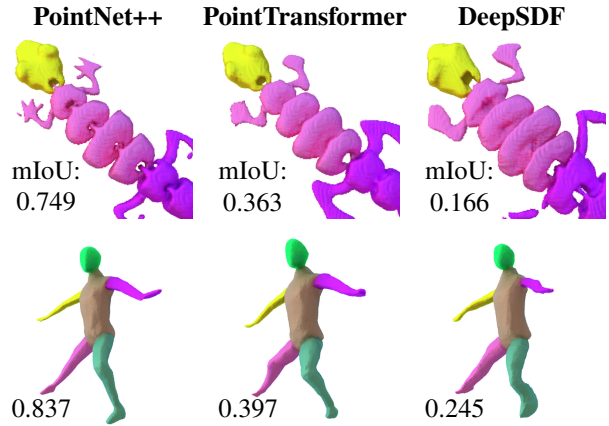

%% file: plots/sampling_methods_comparison.tex
\input{plots/points/points_n=256_k=256_sampling=BODYFORM_shape=SIMPLE.tex}
\input{plots/points/points_n=256_k=256_sampling=BODYFORM_shape=SMALL.tex}
\input{plots/points/points_n=256_k=256_sampling=BODYFORM_shape=THIN.tex}

\input{plots/points/points_n=256_k=256_sampling=SSC_ALL_shape=SIMPLE.tex}
\input{plots/points/points_n=256_k=256_sampling=SSC_ALL_shape=SMALL.tex}
\input{plots/points/points_n=256_k=256_sampling=SSC_ALL_shape=THIN.tex}

\input{plots/points/points_n=256_k=512_sampling=UNIFORM_shape=SIMPLE.tex}
\input{plots/points/points_n=256_k=512_sampling=UNIFORM_shape=SMALL.tex}
\input{plots/points/points_n=256_k=512_sampling=UNIFORM_shape=THIN.tex}

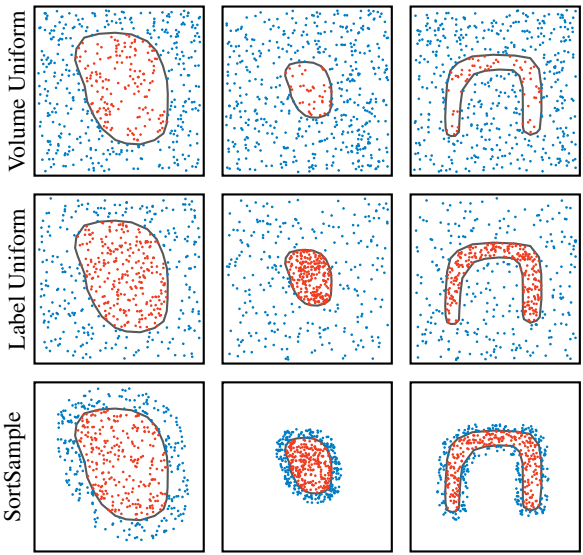
\begin{figure}[tb]
    \ifnographics
    \else
    \begin{tikzpicture}
        \pgfmathsetmacro{\cyanShift}{60} %
        \pgfmathsetmacro{\yellowShift}{10} %
        \pgfmathsetmacro{\grayShift}{70} %

        \draw[draw=black, thick] (0,2.5) rectangle (2.25, 4.75);
        \path[draw=gray!\grayShift!black, smooth, thick] (0.6665625, 3.6925) -- (0.57375, 3.9315625) -- (0.54, 4.0384375) -- (0.556875, 4.1875) -- (0.6609375, 4.3365625) -- (0.8578125, 4.3871875) -- (1.0715625, 4.40125) -- (1.2909375, 4.37875) -- (1.4990625, 4.28875) -- (1.6228125, 4.1565625) -- (1.7071875, 3.9625) -- (1.760625, 3.765625) -- (1.771875, 3.5884375) -- (1.7690625, 3.3521875) -- (1.7578125, 3.1946875) -- (1.715625, 3.0625) -- (1.5946875, 2.9725) -- (1.40625, 2.91625) -- (1.2178125, 2.9275) -- (1.074375, 2.9753125000000002) -- (0.894375, 3.1271875) -- (0.793125, 3.2875) -- (0.7003125, 3.4871875) -- cycle;
        \foreach \x\y in \sscallSimpleOutside{
            \draw[cyan!\cyanShift!blue, fill=cyan!\cyanShift!blue] (\x, \y) circle (0.01);
        }
        \foreach \x\y in \sscallSimpleInside{
            \draw[yellow!\yellowShift!red, fill=yellow!\yellowShift!red] (\x, \y) circle (0.01);
        }
        \node[anchor=south, rotate=90] at (0, 3.625) {\small SortSample};

        \draw[draw=black, thick] (0,5.0) rectangle (2.25, 7.25);
        \path[draw=gray!\grayShift!black, smooth, thick] (0.6665625, 6.1925) -- (0.57375, 6.4315625) -- (0.54, 6.5384375) -- (0.556875, 6.6875) -- (0.6609375, 6.8365625) -- (0.8578125, 6.8871875) -- (1.0715625, 6.90125) -- (1.2909375, 6.87875) -- (1.4990625, 6.78875) -- (1.6228125, 6.6565625) -- (1.7071875, 6.4625) -- (1.760625, 6.265625) -- (1.771875, 6.0884374999999995) -- (1.7690625, 5.8521875) -- (1.7578125, 5.6946875) -- (1.715625, 5.5625) -- (1.5946875, 5.4725) -- (1.40625, 5.41625) -- (1.2178125, 5.4275) -- (1.074375, 5.4753125) -- (0.894375, 5.6271875) -- (0.793125, 5.7875) -- (0.7003125, 5.9871875) -- cycle;
        \foreach \x\y in \bodyformSimpleOutside{
            \draw[cyan!\cyanShift!blue, fill=cyan!\cyanShift!blue] (\x, \y) circle (0.01);
        }
        \foreach \x\y in \bodyformSimpleInside{
            \draw[yellow!\yellowShift!red, fill=yellow!\yellowShift!red] (\x, \y) circle (0.01);
        }
        \node[anchor=south, rotate=90] at (0, 6.125) {\small Label Uniform};

        \draw[draw=black, thick] (0,7.5) rectangle (2.25, 9.75);
        \path[draw=gray!\grayShift!black, smooth, thick] (0.6665625, 8.692499999999999) -- (0.57375, 8.9315625) -- (0.54, 9.0384375) -- (0.556875, 9.1875) -- (0.6609375, 9.3365625) -- (0.8578125, 9.3871875) -- (1.0715625, 9.401250000000001) -- (1.2909375, 9.37875) -- (1.4990625, 9.28875) -- (1.6228125, 9.1565625) -- (1.7071875, 8.9625) -- (1.760625, 8.765625) -- (1.771875, 8.5884375) -- (1.7690625, 8.3521875) -- (1.7578125, 8.1946875) -- (1.715625, 8.0625) -- (1.5946875, 7.9725) -- (1.40625, 7.91625) -- (1.2178125, 7.9275) -- (1.074375, 7.9753125) -- (0.894375, 8.1271875) -- (0.793125, 8.2875) -- (0.7003125, 8.4871875) -- cycle;
        \foreach \x\y in \uniformSimpleOutside{
            \draw[cyan!\cyanShift!blue, fill=cyan!\cyanShift!blue] (\x, \y) circle (0.01);
        }
        \foreach \x\y in \uniformSimpleInside{
            \draw[yellow!\yellowShift!red, fill=yellow!\yellowShift!red] (\x, \y) circle (0.01);
        }
        \node[anchor=south, rotate=90] at (0, 8.625) {\small Volume Uniform};

        \draw[draw=black, thick] (2.5,2.5) rectangle (4.75, 4.75);
        \path[draw=gray!\grayShift!black, smooth, thick] (3.394375, 3.65875) -- (3.349375, 3.776875) -- (3.3325, 3.8303125) -- (3.3409375, 3.90625) -- (3.3915625, 3.979375) -- (3.49, 4.0046875) -- (3.596875, 4.0131250000000005) -- (3.7065625, 4.001875) -- (3.810625, 3.956875) -- (3.8725, 3.8893750000000002) -- (3.9146875000000003, 3.79375) -- (3.9428125, 3.6953125) -- (3.9484375, 3.6053125) -- (3.9456249999999997, 3.4871875) -- (3.94, 3.4084375) -- (3.9203125, 3.34375) -- (3.8584375, 3.29875) -- (3.765625, 3.270625) -- (3.67, 3.27625) -- (3.5996875, 3.29875) -- (3.5096875, 3.3746875) -- (3.4590625, 3.45625) -- (3.41125, 3.5546875) -- cycle;
        \foreach \x\y in \sscallSmallOutside{
            \draw[cyan!\cyanShift!blue, fill=cyan!\cyanShift!blue] (\x, \y) circle (0.01);
        }
        \foreach \x\y in \sscallSmallInside{
            \draw[yellow!\yellowShift!red, fill=yellow!\yellowShift!red] (\x, \y) circle (0.01);
        }

        \draw[draw=black, thick] (2.5,5.0) rectangle (4.75, 7.25);
        \path[draw=gray!\grayShift!black, smooth, thick] (3.394375, 6.1587499999999995) -- (3.349375, 6.276875) -- (3.3325, 6.3303125) -- (3.3409375, 6.40625) -- (3.3915625, 6.479375) -- (3.49, 6.5046875) -- (3.596875, 6.5131250000000005) -- (3.7065625, 6.501875) -- (3.810625, 6.456875) -- (3.8725, 6.389375) -- (3.9146875000000003, 6.29375) -- (3.9428125, 6.1953125) -- (3.9484375, 6.1053125) -- (3.9456249999999997, 5.9871875) -- (3.94, 5.9084375) -- (3.9203125, 5.84375) -- (3.8584375, 5.79875) -- (3.765625, 5.770625) -- (3.67, 5.77625) -- (3.5996875, 5.79875) -- (3.5096875, 5.8746875) -- (3.4590625, 5.95625) -- (3.41125, 6.0546875) -- cycle;
        \foreach \x\y in \bodyformSmallOutside{
            \draw[cyan!\cyanShift!blue, fill=cyan!\cyanShift!blue] (\x, \y) circle (0.01);
        }
        \foreach \x\y in \bodyformSmallInside{
            \draw[yellow!\yellowShift!red, fill=yellow!\yellowShift!red] (\x, \y) circle (0.01);
        }

        \draw[draw=black, thick] (2.5,7.5) rectangle (4.75, 9.75);
        \path[draw=gray!\grayShift!black, smooth, thick] (3.394375, 8.65875) -- (3.349375, 8.776875) -- (3.3325, 8.8303125) -- (3.3409375, 8.90625) -- (3.3915625, 8.979375000000001) -- (3.49, 9.0046875) -- (3.596875, 9.013125) -- (3.7065625, 9.001875) -- (3.810625, 8.956875) -- (3.8725, 8.889375) -- (3.9146875000000003, 8.79375) -- (3.9428125, 8.6953125) -- (3.9484375, 8.6053125) -- (3.9456249999999997, 8.4871875) -- (3.94, 8.4084375) -- (3.9203125, 8.34375) -- (3.8584375, 8.29875) -- (3.765625, 8.270625) -- (3.67, 8.27625) -- (3.5996875, 8.29875) -- (3.5096875, 8.3746875) -- (3.4590625, 8.45625) -- (3.41125, 8.5546875) -- cycle;
        \foreach \x\y in \uniformSmallOutside{
            \draw[cyan!\cyanShift!blue, fill=cyan!\cyanShift!blue] (\x, \y) circle (0.01);
        }
        \foreach \x\y in \uniformSmallInside{
            \draw[yellow!\yellowShift!red, fill=yellow!\yellowShift!red] (\x, \y) circle (0.01);
        }

        \draw[draw=black, thick] (5.0,2.5) rectangle (7.25, 4.75);
        \path[draw=gray!\grayShift!black, smooth, thick] (5.465625, 3.1384375) -- (5.46, 3.270625) -- (5.4684375, 3.47875) -- (5.4853125, 3.7234375) -- (5.5303125, 3.900625) -- (5.6259375, 3.9990625) -- (5.82, 4.08625) -- (6.095625, 4.10875) -- (6.3881250000000005, 4.0975) -- (6.590625, 4.06375) -- (6.7059375, 3.9175) -- (6.7368749999999995, 3.6925) -- (6.7396875000000005, 3.36625) -- (6.7171875, 3.11875) -- (6.675, 3.0709375) -- (6.590625, 3.0428125) -- (6.50625, 3.090625) -- (6.4921875, 3.1525) -- (6.4921875, 3.4253125) -- (6.489375, 3.67) -- (6.4471875, 3.8274999999999997) -- (6.36, 3.9034375) -- (6.1743749999999995, 3.911875) -- (5.9915625, 3.895) -- (5.8425, 3.85) -- (5.7609375, 3.743125) -- (5.6878125, 3.63625) -- (5.6625, 3.4421875) -- (5.6484375, 3.2228125) -- (5.6428125, 3.0821875) -- (5.5921875, 3.02875) -- (5.5359375, 3.0259375) -- (5.4853125, 3.05125) -- cycle;
        \foreach \x\y in \sscallThinOutside{
            \draw[cyan!\cyanShift!blue, fill=cyan!\cyanShift!blue] (\x, \y) circle (0.01);
        }
        \foreach \x\y in \sscallThinInside{
            \draw[yellow!\yellowShift!red, fill=yellow!\yellowShift!red] (\x, \y) circle (0.01);
        }
        
        \draw[draw=black, thick] (5.0,5.0) rectangle (7.25, 7.25);
        \path[draw=gray!\grayShift!black, smooth, thick] (5.465625, 5.6384375) -- (5.46, 5.770625) -- (5.4684375, 5.97875) -- (5.4853125, 6.2234375) -- (5.5303125, 6.400625) -- (5.6259375, 6.4990625) -- (5.82, 6.58625) -- (6.095625, 6.60875) -- (6.3881250000000005, 6.5975) -- (6.590625, 6.56375) -- (6.7059375, 6.4175) -- (6.7368749999999995, 6.1925) -- (6.7396875000000005, 5.86625) -- (6.7171875, 5.61875) -- (6.675, 5.5709375) -- (6.590625, 5.5428125) -- (6.50625, 5.590625) -- (6.4921875, 5.6525) -- (6.4921875, 5.9253125) -- (6.489375, 6.17) -- (6.4471875, 6.3275) -- (6.36, 6.4034375) -- (6.1743749999999995, 6.411875) -- (5.9915625, 6.395) -- (5.8425, 6.35) -- (5.7609375, 6.243125) -- (5.6878125, 6.13625) -- (5.6625, 5.9421875) -- (5.6484375, 5.7228125) -- (5.6428125, 5.5821875) -- (5.5921875, 5.5287500000000005) -- (5.5359375, 5.5259374999999995) -- (5.4853125, 5.55125) -- cycle;
        \foreach \x\y in \bodyformThinOutside{
            \draw[cyan!\cyanShift!blue, fill=cyan!\cyanShift!blue] (\x, \y) circle (0.01);
        }
        \foreach \x\y in \bodyformThinInside{
            \draw[yellow!\yellowShift!red, fill=yellow!\yellowShift!red] (\x, \y) circle (0.01);
        }
        
        \draw[draw=black, thick] (5.0,7.5) rectangle (7.25, 9.75);
        \path[draw=gray!\grayShift!black, smooth, thick] (5.465625, 8.1384375) -- (5.46, 8.270625) -- (5.4684375, 8.47875) -- (5.4853125, 8.7234375) -- (5.5303125, 8.900625) -- (5.6259375, 8.9990625) -- (5.82, 9.08625) -- (6.095625, 9.10875) -- (6.3881250000000005, 9.0975) -- (6.590625, 9.06375) -- (6.7059375, 8.9175) -- (6.7368749999999995, 8.692499999999999) -- (6.7396875000000005, 8.36625) -- (6.7171875, 8.11875) -- (6.675, 8.0709375) -- (6.590625, 8.0428125) -- (6.50625, 8.090625) -- (6.4921875, 8.1525) -- (6.4921875, 8.4253125) -- (6.489375, 8.67) -- (6.4471875, 8.8275) -- (6.36, 8.903437499999999) -- (6.1743749999999995, 8.911875) -- (5.9915625, 8.895) -- (5.8425, 8.85) -- (5.7609375, 8.743125) -- (5.6878125, 8.63625) -- (5.6625, 8.4421875) -- (5.6484375, 8.2228125) -- (5.6428125, 8.0821875) -- (5.5921875, 8.02875) -- (5.5359375, 8.0259375) -- (5.4853125, 8.05125) -- cycle;
        \foreach \x\y in \uniformThinOutside{
            \draw[cyan!\cyanShift!blue, fill=cyan!\cyanShift!blue] (\x, \y) circle (0.01);
        }
        \foreach \x\y in \uniformThinInside{
            \draw[yellow!\yellowShift!red, fill=yellow!\yellowShift!red] (\x, \y) circle (0.01);
        }     
    \end{tikzpicture}
    \fi
    \caption{Sample distributions using different sampling methods on 2D toy scenes. Only SortSample results in a dataset that is unbiased between red and blue classes both globally and locally near object boundaries.}
    \label{fig:sampling_methods_comparison}
\end{figure}

%% file: plots/points/points_n=256_k=256_sampling=BODYFORM_shape=SIMPLE.tex
\def\bodyformSimpleOutside{1.505/5.292, 2.189/5.341, 2.101/6.888, 0.238/5.813, 0.778/5.398, 0.306/6.417, 0.617/5.230, 0.283/5.602, 2.185/6.419, 0.216/7.197, 0.162/6.215, 2.178/6.799, 0.343/5.331, 0.278/5.320, 0.376/6.903, 2.090/5.132, 1.853/6.798, 1.799/5.281, 0.335/5.259, 1.770/6.188, 2.182/5.661, 1.962/5.834, 0.302/5.979, 1.029/5.204, 1.139/5.446, 0.442/5.166, 1.975/5.667, 2.051/6.821, 1.891/6.283, 0.853/5.629, 0.386/5.420, 0.154/6.103, 0.244/5.659, 0.309/5.659, 1.533/7.026, 0.632/7.113, 0.505/5.674, 0.179/5.481, 0.128/6.503, 0.876/5.173, 0.337/5.767, 2.101/5.690, 1.080/5.253, 0.133/5.570, 1.383/5.175, 0.373/6.265, 1.694/5.286, 0.125/5.547, 0.503/5.946, 0.397/5.985, 2.124/6.586, 0.344/6.811, 1.780/5.199, 0.453/5.800, 1.214/5.208, 0.542/5.541, 0.603/6.322, 1.755/7.069, 0.842/6.975, 1.196/5.068, 0.660/5.470, 1.135/7.109, 1.697/7.035, 1.776/6.770, 0.654/6.099, 0.922/5.501, 2.193/5.564, 0.645/5.077, 0.455/5.657, 0.687/5.124, 2.003/7.117, 0.238/6.355, 2.053/5.458, 0.207/6.703, 0.661/5.410, 0.473/5.636, 2.128/5.438, 0.885/7.069, 2.144/6.616, 2.057/5.879, 2.129/6.620, 0.193/6.823, 1.596/6.989, 1.839/5.328, 0.559/5.353, 2.190/6.682, 0.530/6.706, 2.084/7.051, 0.128/5.908, 0.206/5.167, 1.946/5.708, 0.152/6.040, 0.187/5.553, 0.113/6.900, 0.726/5.864, 0.375/6.217, 0.064/6.517, 1.294/7.074, 1.992/5.106, 0.315/5.241, 0.762/5.477, 2.183/5.693, 2.007/7.190, 1.917/5.480, 0.995/6.999, 0.221/5.632, 1.743/6.690, 0.998/6.997, 0.098/6.692, 1.843/7.046, 2.093/7.157, 2.077/6.265, 0.598/6.199, 0.760/5.331, 0.766/5.769, 0.536/5.671, 0.224/5.299, 1.226/7.146, 0.638/5.720, 1.540/6.885, 0.295/5.672, 0.159/5.211, 0.709/5.412, 2.073/5.474, 0.245/6.428, 0.097/5.989, 1.068/7.111, 1.174/6.973, 1.560/6.920, 0.162/6.377, 1.804/6.512, 0.846/7.067, 1.523/5.354, 0.314/5.993, 1.433/7.093, 1.351/5.150, 0.724/5.518, 0.700/7.055, 1.760/5.479, 2.131/6.965, 1.672/6.755, 2.005/6.166, 0.952/5.275, 0.550/6.104, 0.404/7.051, 2.136/6.451, 2.134/6.942, 0.109/6.352, 0.177/6.754, 2.038/6.688, 0.491/6.166, 1.999/5.442, 1.109/5.337, 1.871/6.738, 0.382/6.973, 0.469/6.281, 2.162/5.819, 1.892/6.686, 0.403/7.027, 0.549/7.023, 1.323/5.111, 2.125/5.949, 1.641/5.413, 0.162/7.000, 0.659/6.162, 1.775/6.345, 2.127/7.037, 0.401/6.222, 0.571/6.266, 1.609/6.990, 0.192/7.005, 0.538/5.869, 1.794/6.258, 0.794/7.019, 1.591/6.708, 1.123/5.363, 0.524/5.647, 2.061/5.986, 2.132/6.285, 0.944/5.206, 2.059/7.116, 1.716/6.577, 0.086/5.894, 1.466/6.834, 2.032/6.629, 0.982/6.954, 1.978/5.263, 0.625/5.703, 0.579/7.168, 2.097/5.091, 1.494/7.076, 1.570/5.315, 0.515/5.108, 2.002/6.657, 0.645/5.479, 0.696/5.274, 0.358/6.316, 0.091/6.062, 0.092/6.808, 1.926/5.595, 2.001/7.045, 2.107/5.721, 0.433/6.493, 1.210/7.015, 1.715/7.170, 1.225/5.350, 1.645/5.104, 1.924/6.356, 1.406/7.061, 0.125/5.656, 0.193/5.697, 1.864/6.644, 0.098/6.617, 0.592/6.746, 1.499/7.009, 2.033/6.430, 0.585/5.429, 0.538/5.160, 1.870/5.112, 0.086/5.749, 1.571/7.096, 1.978/7.118, 0.180/6.031, 0.936/7.170, 0.641/5.251, 0.434/6.788, 1.223/7.139, 1.084/7.004, 2.052/5.498, 1.787/6.731, 0.134/5.285, 0.410/7.096, 0.587/7.096, 0.409/6.219, 0.711/7.126, 1.503/5.217, 1.570/7.087, 0.741/5.083, 0.403/5.730, 0.822/7.006, 0.404/5.837, 0.716/5.381, 2.019/6.051, 0.930/7.063, 1.532/5.410, 0.493/6.946, 2.065/6.014, 1.724/6.567, 0.664/6.005, 2.057/6.264, 0.838/6.893, 1.082/5.434, 2.017/6.417, 0.601/6.316, 1.221/7.061, 1.471/6.911
}
\def\bodyformSimpleInside{1.495/5.814, 1.140/5.710, 1.186/5.708, 1.122/6.846, 0.625/6.308, 1.530/6.549, 1.407/5.486, 1.497/6.627, 1.626/6.214, 1.753/5.883, 0.697/6.595, 1.066/6.356, 1.524/5.456, 0.999/6.780, 0.993/6.203, 1.345/5.907, 1.540/6.561, 0.738/6.618, 0.894/6.878, 1.065/5.509, 1.059/6.022, 0.955/5.591, 1.248/5.847, 1.201/5.506, 1.102/5.589, 1.149/6.673, 0.790/6.289, 1.197/5.917, 0.869/6.021, 1.143/5.722, 0.754/6.355, 1.177/6.322, 1.157/5.491, 1.413/6.004, 0.973/5.805, 1.266/5.841, 1.682/5.972, 1.507/6.248, 0.941/6.329, 1.223/6.819, 0.765/6.130, 1.393/5.551, 1.010/5.738, 1.481/6.627, 0.671/6.714, 1.002/6.512, 1.591/5.716, 1.357/6.192, 1.351/6.344, 1.708/6.132, 1.052/5.628, 0.979/6.541, 1.396/6.086, 1.240/6.762, 1.493/5.486, 0.930/5.896, 1.225/6.081, 1.448/6.335, 1.398/5.587, 1.035/6.599, 0.981/6.343, 1.177/6.838, 1.114/5.982, 0.722/6.778, 1.163/6.476, 1.015/6.495, 1.220/6.834, 0.922/6.302, 1.357/6.685, 0.860/6.351, 1.103/6.518, 1.103/5.796, 1.730/6.068, 0.732/6.187, 1.113/5.904, 1.142/6.850, 1.192/5.673, 1.366/5.696, 1.243/6.512, 1.024/6.881, 1.342/6.026, 0.933/6.089, 0.858/5.768, 0.890/6.656, 1.086/6.366, 1.311/6.474, 1.542/5.835, 1.617/5.885, 0.778/6.140, 1.239/6.172, 0.974/6.133, 1.386/6.513, 1.570/5.884, 1.424/6.077, 1.429/5.643, 1.555/6.342, 1.743/5.958, 1.350/5.780, 1.538/6.718, 0.992/5.665, 1.419/5.704, 1.660/5.843, 1.308/6.106, 0.639/6.589, 1.150/6.550, 1.441/5.930, 1.398/5.694, 1.715/6.328, 1.653/5.667, 1.687/5.933, 1.397/6.746, 1.434/6.551, 1.253/6.359, 0.907/5.888, 1.551/5.771, 1.523/5.649, 1.420/6.275, 1.385/6.203, 1.670/5.724, 1.400/6.148, 0.745/6.715, 0.901/6.619, 0.823/6.259, 1.627/6.291, 0.948/5.593, 1.500/6.254, 1.232/6.226, 1.150/5.925, 1.413/6.599, 1.504/6.616, 1.546/6.616, 1.456/6.227, 0.940/6.588, 1.640/6.605, 1.724/5.823, 1.550/5.667, 0.774/6.662, 1.220/5.641, 1.433/6.004, 1.608/6.622, 0.729/6.717, 1.060/6.357, 1.086/6.399, 1.680/5.644, 1.451/6.465, 1.101/5.951, 0.772/6.738, 0.700/6.228, 1.057/5.773, 1.561/5.653, 1.601/5.539, 1.224/6.379, 0.755/5.988, 1.291/5.994, 0.821/5.960, 1.270/6.586, 0.693/6.825, 1.351/6.572, 1.255/5.709, 0.924/6.373, 1.317/5.859, 1.586/6.407, 0.751/6.380, 1.134/6.567, 0.796/5.843, 0.809/6.325, 1.209/5.695, 1.564/6.206, 0.941/5.700, 1.354/6.006, 1.247/5.554, 1.611/5.777, 1.651/5.891, 1.489/5.792, 0.576/6.690, 1.674/6.188, 1.578/6.407, 1.447/6.172, 0.604/6.574, 1.172/6.185, 0.973/6.469, 1.503/6.401, 0.696/6.157, 1.210/6.606, 1.667/5.638, 0.947/5.629, 1.162/6.786, 1.177/6.865, 1.186/6.748, 1.053/6.490, 0.981/5.820, 1.362/6.624, 1.234/5.544, 0.922/6.335, 1.004/6.758, 1.629/5.503, 1.264/6.432, 1.101/5.915, 1.478/6.609, 1.468/6.143, 1.713/6.273, 1.048/6.051, 1.299/6.271, 1.471/6.402, 1.500/6.426, 1.345/5.503, 1.609/5.889, 0.683/6.393, 0.846/6.158, 1.455/6.150, 1.108/5.858, 1.404/6.621, 0.944/5.807, 1.383/6.784, 0.850/6.405, 1.207/6.184, 1.716/6.045, 1.728/6.046, 1.160/6.568, 0.691/6.172, 1.011/6.458, 0.714/6.511, 1.129/6.074, 0.909/6.561, 0.878/6.249, 0.999/6.817, 0.952/6.688, 1.277/6.535, 0.761/6.024, 1.612/5.613, 1.429/5.593, 1.520/6.237, 1.062/6.097, 1.637/5.731, 1.559/6.287, 1.034/5.991, 1.201/5.515, 1.475/5.601, 1.689/6.086, 0.901/6.470, 1.169/6.156, 0.910/6.867, 0.952/5.643, 1.416/6.081, 1.384/6.361, 0.808/6.217, 1.163/6.133, 1.689/5.794, 0.678/6.779, 1.080/6.521, 1.125/6.122, 1.090/6.772, 1.104/6.253, 0.927/5.797, 0.711/6.311, 1.278/6.016
}

%% file: plots/points/points_n=256_k=256_sampling=BODYFORM_shape=SMALL.tex
\def\bodyformSmallOutside{3.101/6.858, 2.854/5.405, 4.586/5.274, 3.801/5.351, 4.691/6.866, 3.853/5.717, 4.119/5.424, 3.728/5.173, 4.543/7.028, 4.480/6.121, 4.258/5.986, 4.555/6.573, 3.211/6.251, 3.345/5.948, 3.265/6.822, 3.696/6.784, 4.035/5.734, 4.455/6.895, 3.928/5.705, 2.646/6.779, 3.116/6.072, 4.029/5.782, 2.904/5.747, 4.003/5.692, 4.350/5.789, 3.540/6.797, 3.209/6.606, 2.787/5.159, 3.437/6.831, 4.304/5.242, 2.805/5.963, 3.843/6.860, 4.092/6.695, 3.125/6.428, 4.298/5.554, 3.458/6.548, 3.135/6.853, 4.579/6.934, 3.920/5.271, 4.444/6.959, 4.498/6.310, 4.413/5.479, 2.693/5.999, 4.563/6.401, 3.192/6.310, 4.053/6.198, 4.009/6.342, 3.366/5.085, 4.497/5.066, 2.767/5.310, 3.719/6.755, 4.330/6.313, 4.640/7.068, 2.865/6.429, 2.948/5.406, 4.628/6.397, 3.789/5.401, 3.136/6.032, 3.523/5.750, 3.238/6.691, 3.372/5.055, 4.394/7.192, 3.634/5.541, 3.046/7.092, 2.941/7.143, 4.555/6.247, 2.863/7.029, 3.132/5.546, 2.595/6.324, 4.578/5.222, 2.747/5.288, 3.160/7.084, 4.304/6.143, 3.246/6.211, 3.164/6.129, 2.741/6.640, 4.143/6.351, 3.098/5.645, 4.108/7.064, 3.504/5.292, 4.486/5.356, 3.697/5.530, 4.157/6.596, 2.774/5.639, 2.950/5.669, 2.725/5.728, 3.437/5.688, 4.370/5.219, 3.900/5.471, 3.073/5.330, 3.958/6.287, 2.825/7.099, 4.299/5.906, 3.508/6.869, 3.973/6.817, 4.603/5.754, 3.117/5.533, 3.180/5.247, 3.102/6.276, 3.844/5.447, 3.214/6.915, 4.512/6.935, 2.725/6.439, 2.593/5.129, 3.207/5.655, 3.688/5.147, 2.698/6.773, 4.102/5.069, 3.950/6.970, 3.340/5.368, 3.094/6.864, 4.210/6.249, 3.402/6.069, 4.690/6.885, 4.477/6.625, 4.425/6.553, 3.238/6.417, 3.054/6.485, 2.858/5.916, 3.069/5.683, 4.255/6.980, 2.640/6.042, 4.476/5.690, 3.006/6.048, 3.050/5.980, 4.129/5.105, 3.129/6.412, 3.866/5.456, 2.579/6.179, 3.393/5.351, 4.342/5.307, 4.585/7.117, 4.532/5.861, 4.276/5.188, 4.009/5.505, 4.385/5.959, 3.941/6.263, 3.501/7.175, 3.274/6.298, 4.651/6.208, 3.334/6.870, 4.444/7.118, 3.290/5.190, 3.163/6.276, 4.038/7.163, 3.829/7.017, 4.284/5.966, 3.794/5.478, 4.695/5.512, 3.421/5.793, 4.477/6.644, 3.817/6.741, 3.038/6.674, 3.107/6.752, 3.093/7.128, 4.161/6.035, 3.341/5.064, 4.296/5.618, 3.916/6.832, 4.415/5.479, 3.181/5.754, 2.597/6.997, 2.672/6.351, 2.616/6.383, 4.302/6.837, 3.632/5.436, 3.358/5.193, 2.832/6.583, 3.308/6.866, 4.497/5.304, 4.247/6.737, 3.190/5.532, 3.367/6.130, 4.502/6.689, 3.039/6.707, 3.156/5.435, 4.044/6.814, 2.904/6.177, 2.561/6.292, 2.704/5.319, 2.691/6.395, 2.868/5.830, 3.276/6.751, 3.988/6.609, 3.039/6.895, 4.066/6.046, 4.217/5.384, 3.384/6.027, 3.866/6.506, 2.885/6.702, 3.234/5.451, 3.335/5.641, 3.453/5.930, 2.754/6.486, 2.609/5.435, 3.069/5.224, 4.400/6.952, 2.878/5.367, 2.791/6.141, 4.620/6.293, 2.856/5.981, 3.192/5.420, 2.783/6.768, 2.590/6.707, 3.017/6.387, 3.309/6.145, 4.622/5.849, 3.467/6.922, 3.520/5.396, 2.843/6.562, 4.241/5.293, 3.382/6.662, 2.739/5.546, 4.662/6.534, 2.799/5.808, 4.356/5.660, 4.544/7.036, 2.918/6.748, 2.888/5.151, 4.310/6.603, 3.952/5.187, 4.261/6.285, 4.236/6.842, 4.174/6.000, 3.209/6.295, 4.675/7.128, 2.967/6.883, 4.340/6.630, 3.389/7.150, 2.998/6.895, 2.703/5.098, 2.585/6.286, 3.481/6.911, 2.816/5.679, 3.826/5.250, 3.386/6.083, 3.104/6.966, 4.124/5.469, 4.348/5.630, 3.305/6.007, 2.725/6.883, 4.108/5.996, 2.669/6.302, 2.984/5.306, 4.569/5.348, 2.747/7.168, 2.759/6.610, 3.742/6.669, 3.483/6.586, 3.566/6.673, 3.505/5.357, 3.322/5.534, 4.696/7.049, 2.960/6.913, 3.536/5.467, 2.720/5.240
}
\def\bodyformSmallInside{3.757/5.899, 3.425/6.148, 3.595/5.891, 3.625/6.154, 3.734/6.135, 3.845/6.190, 3.528/6.254, 3.534/6.336, 3.841/6.108, 3.873/6.096, 3.660/6.386, 3.566/6.355, 3.758/6.332, 3.644/5.843, 3.813/6.396, 3.782/6.334, 3.758/6.376, 3.803/6.460, 3.533/6.152, 3.540/6.442, 3.779/5.829, 3.774/5.911, 3.901/6.122, 3.744/5.876, 3.479/6.278, 3.852/6.396, 3.838/6.409, 3.539/6.279, 3.834/6.287, 3.536/5.861, 3.542/5.954, 3.847/6.410, 3.556/6.120, 3.544/6.230, 3.699/6.290, 3.651/6.498, 3.573/6.252, 3.592/6.296, 3.646/6.096, 3.862/6.204, 3.414/6.062, 3.790/6.360, 3.445/6.458, 3.670/6.412, 3.528/6.172, 3.395/6.399, 3.602/5.949, 3.724/6.328, 3.507/5.907, 3.522/5.880, 3.570/6.064, 3.775/6.300, 3.755/5.917, 3.655/5.842, 3.399/6.250, 3.753/6.186, 3.576/6.095, 3.431/6.121, 3.772/6.398, 3.433/6.022, 3.818/5.795, 3.607/6.043, 3.493/6.428, 3.853/6.165, 3.546/6.193, 3.726/6.266, 3.540/6.468, 3.647/6.192, 3.734/6.248, 3.571/6.494, 3.875/6.328, 3.552/6.318, 3.735/5.903, 3.545/5.995, 3.595/6.414, 3.881/5.925, 3.792/6.293, 3.656/6.101, 3.410/6.293, 3.858/5.800, 3.771/5.921, 3.519/6.269, 3.614/6.354, 3.810/5.940, 3.515/6.231, 3.694/5.909, 3.856/6.302, 3.644/6.102, 3.583/6.482, 3.924/6.259, 3.640/6.332, 3.620/6.214, 3.561/6.510, 3.624/5.957, 3.512/5.936, 3.490/6.069, 3.711/5.845, 3.785/6.212, 3.922/5.889, 3.895/6.323, 3.731/5.828, 3.503/6.114, 3.671/5.782, 3.600/6.408, 3.944/6.173, 3.904/6.003, 3.715/6.183, 3.543/6.507, 3.600/6.353, 3.636/5.858, 3.782/5.816, 3.592/6.499, 3.423/6.309, 3.760/5.977, 3.652/6.449, 3.795/6.291, 3.546/6.395, 3.614/6.500, 3.750/5.957, 3.637/6.284, 3.565/6.242, 3.752/6.380, 3.939/5.976, 3.495/6.023, 3.731/5.914, 3.791/6.014, 3.735/6.417, 3.784/6.014, 3.551/6.434, 3.680/6.164, 3.507/5.896, 3.550/6.251, 3.476/6.488, 3.527/5.876, 3.574/6.076, 3.726/5.951, 3.547/5.880, 3.384/6.351, 3.737/6.443, 3.899/5.913, 3.876/5.841, 3.683/6.349, 3.592/6.229, 3.649/5.801, 3.939/6.033, 3.755/5.985, 3.446/6.199, 3.607/6.095, 3.399/6.391, 3.617/5.991, 3.621/6.173, 3.521/6.110, 3.382/6.208, 3.829/5.887, 3.445/6.468, 3.631/6.400, 3.927/5.900, 3.683/6.283, 3.642/6.224, 3.889/5.822, 3.842/6.170, 3.930/6.180, 3.801/5.886, 3.779/6.276, 3.757/6.278, 3.761/5.824, 3.691/6.192, 3.715/6.469, 3.865/5.807, 3.756/6.373, 3.731/5.808, 3.570/6.157, 3.494/6.476, 3.573/6.146, 3.905/6.196, 3.658/5.784, 3.674/6.110, 3.670/6.505, 3.585/6.197, 3.843/5.822, 3.719/5.859, 3.582/6.497, 3.836/6.282, 3.847/6.405, 3.587/6.279, 3.793/5.921, 3.351/6.351, 3.913/6.171, 3.723/6.479, 3.816/6.338, 3.828/5.896, 3.682/6.288, 3.804/5.877, 3.419/6.326, 3.858/5.808, 3.491/6.072, 3.658/6.039, 3.845/6.357, 3.604/6.275, 3.733/6.247, 3.625/6.483, 3.783/6.432, 3.751/6.267, 3.686/6.132, 3.796/6.425, 3.738/5.885, 3.570/6.340, 3.498/6.362, 3.830/5.861, 3.893/5.831, 3.485/6.378, 3.445/5.993, 3.585/5.855, 3.374/6.219, 3.840/6.108, 3.881/5.826, 3.737/6.329, 3.785/6.252, 3.879/6.083, 3.884/5.853, 3.891/5.852, 3.418/6.410, 3.679/5.860, 3.549/6.429, 3.883/6.027, 3.715/5.902, 3.668/6.331, 3.702/5.951, 3.682/6.248, 3.851/5.841, 3.944/6.050, 3.836/5.864, 3.430/6.039, 3.560/6.247, 3.857/6.205, 3.920/6.223, 3.428/6.051, 3.807/6.043, 3.535/6.342, 3.714/5.919, 3.869/5.903, 3.898/6.253, 3.852/6.337, 3.718/6.359, 3.453/6.110, 3.712/6.000, 3.411/6.429, 3.592/6.061, 3.601/6.088, 3.619/5.893, 3.862/6.054, 3.535/6.252, 3.752/6.019, 3.848/6.098, 3.607/6.007, 3.677/5.938
}

%% file: plots/points/points_n=256_k=256_sampling=BODYFORM_shape=THIN.tex
\def\bodyformThinOutside{6.400/5.181, 6.520/7.050, 5.668/5.531, 5.205/5.451, 5.508/5.192, 5.749/6.100, 6.115/5.874, 6.130/6.672, 7.186/6.635, 6.359/5.458, 5.505/6.384, 5.507/5.295, 5.886/6.062, 5.692/5.581, 6.430/5.617, 7.168/6.564, 5.664/7.142, 7.013/6.619, 5.405/5.355, 6.146/7.198, 6.102/5.626, 5.097/5.487, 6.754/5.691, 7.083/5.107, 6.462/5.208, 6.043/6.849, 7.192/5.342, 6.829/7.099, 5.406/6.129, 6.730/6.682, 5.271/5.540, 5.346/6.307, 5.750/5.920, 6.804/6.556, 7.198/6.636, 6.148/5.780, 5.267/6.654, 5.713/6.935, 5.760/5.382, 7.136/6.315, 5.094/5.237, 5.717/5.777, 6.627/5.145, 6.243/6.015, 6.738/6.364, 7.134/6.458, 5.796/7.130, 6.445/5.242, 5.759/5.542, 6.757/5.607, 5.610/5.064, 5.079/5.941, 6.641/6.844, 6.108/5.157, 5.811/5.828, 5.434/5.431, 7.047/6.686, 6.871/5.350, 5.789/5.127, 6.905/5.980, 5.650/5.156, 6.033/6.326, 7.025/5.653, 6.139/5.232, 6.155/6.328, 6.108/5.717, 5.430/7.131, 7.193/6.617, 6.922/5.522, 5.729/7.025, 6.362/5.861, 5.859/5.890, 5.504/5.187, 6.862/6.121, 5.531/6.604, 5.139/6.069, 5.953/7.153, 6.005/5.996, 6.411/5.208, 5.893/6.236, 5.618/5.403, 6.902/6.556, 6.553/5.160, 5.085/6.546, 6.410/6.292, 6.105/5.312, 6.314/6.221, 6.159/5.405, 5.346/5.710, 5.275/5.613, 6.888/6.925, 7.060/5.179, 6.890/5.904, 7.166/6.519, 7.011/7.168, 5.595/6.859, 6.015/5.148, 5.857/6.907, 6.876/6.630, 6.796/5.105, 6.291/6.321, 5.879/5.392, 7.193/6.424, 5.944/5.124, 5.402/5.090, 5.411/6.521, 5.818/6.764, 6.132/5.081, 6.347/6.396, 6.237/6.350, 7.122/6.214, 6.864/6.719, 5.377/5.310, 6.593/5.107, 6.818/6.891, 7.133/5.813, 6.479/5.699, 5.231/5.555, 7.111/5.648, 6.817/6.617, 5.422/5.668, 5.285/6.647, 6.033/6.071, 6.020/6.305, 5.974/6.266, 5.700/6.692, 5.290/6.354, 7.137/6.683, 5.213/6.895, 5.432/6.414, 5.394/5.923, 6.415/5.698, 5.850/5.411, 5.526/7.105, 5.734/5.432, 5.329/5.530, 5.644/6.833, 7.004/5.862, 6.410/6.039, 5.378/6.254, 7.053/6.925, 6.229/5.632, 5.266/5.410, 5.629/5.077, 6.020/5.804, 5.190/5.391, 5.426/6.085, 6.706/6.974, 5.883/5.291, 7.124/7.069, 5.995/5.059, 5.625/5.456, 6.320/5.770, 7.125/5.064, 6.976/7.003, 5.540/6.699, 7.003/5.532, 6.947/6.205, 6.086/6.225, 6.807/6.296, 5.805/6.770, 5.326/6.540, 7.071/6.920, 6.367/5.260, 6.349/5.178, 5.202/5.270, 6.684/6.970, 5.143/6.149, 5.214/5.863, 7.099/5.676, 6.056/5.978, 6.610/5.509, 5.884/6.922, 6.787/5.653, 7.124/5.325, 5.448/5.971, 6.992/5.876, 7.073/7.077, 6.140/6.721, 5.504/6.855, 6.685/5.232, 7.014/6.753, 7.182/6.284, 6.185/6.897, 5.158/5.388, 6.302/7.029, 5.094/6.177, 5.133/5.647, 6.393/6.951, 6.299/6.829, 5.301/5.674, 5.098/5.081, 6.665/5.520, 5.478/6.164, 6.051/5.321, 6.310/6.321, 5.793/7.022, 6.194/5.216, 6.240/6.925, 5.832/5.930, 7.113/6.027, 7.163/5.397, 6.676/5.517, 6.535/5.302, 5.835/6.801, 6.605/5.460, 7.112/6.592, 6.113/5.187, 6.009/5.356, 6.850/5.435, 5.725/6.684, 5.253/5.776, 5.346/6.307, 5.413/5.323, 5.096/6.493, 6.911/5.712, 6.725/6.905, 6.370/5.390, 7.070/5.518, 6.379/5.073, 6.383/6.781, 6.190/6.802, 6.994/5.927, 5.161/5.298, 5.975/5.581, 5.493/7.041, 6.066/5.108, 5.380/5.695, 5.073/6.769, 6.908/6.537, 6.859/5.258, 5.125/6.945, 5.658/6.592, 6.842/5.345, 6.314/5.696, 6.002/5.530, 5.207/5.289, 7.005/6.289, 5.170/6.908, 5.756/5.418, 6.284/6.744, 6.016/5.692, 5.936/5.943, 6.992/6.028, 6.891/5.881, 5.774/6.015, 5.605/5.050, 5.212/7.170, 5.642/6.686, 5.363/6.241, 6.247/5.401, 5.895/5.458, 5.506/6.767, 5.400/6.876, 7.026/5.879, 6.747/7.048
}
\def\bodyformThinInside{5.641/6.004, 6.317/6.594, 5.865/6.550, 6.514/5.809, 6.458/6.439, 5.772/6.376, 6.552/5.984, 6.629/5.707, 6.702/5.863, 6.218/6.599, 6.586/5.996, 6.567/5.881, 6.665/6.317, 6.522/5.754, 5.474/5.880, 6.468/6.260, 6.142/6.555, 5.506/5.541, 6.649/6.472, 5.657/6.416, 6.398/6.564, 6.126/6.510, 6.671/6.214, 5.544/6.137, 5.525/5.867, 6.704/6.367, 6.173/6.548, 6.725/6.086, 5.866/6.415, 5.929/6.483, 5.606/6.351, 6.502/6.196, 5.667/6.406, 5.635/6.044, 6.624/6.359, 6.091/6.533, 6.017/6.596, 6.665/6.113, 6.591/5.659, 6.290/6.527, 5.793/6.515, 5.997/6.439, 6.243/6.422, 6.699/5.847, 6.455/6.570, 6.523/6.175, 6.658/6.024, 6.459/6.376, 6.095/6.578, 6.490/6.399, 6.033/6.478, 5.608/5.802, 6.569/6.412, 6.497/6.048, 5.481/5.720, 5.597/5.534, 6.102/6.562, 6.068/6.597, 6.562/6.449, 6.628/6.054, 5.617/6.315, 6.303/6.525, 6.292/6.502, 5.728/6.446, 6.574/5.997, 6.502/5.949, 5.617/5.952, 6.250/6.490, 6.312/6.561, 6.500/6.394, 5.820/6.415, 6.564/6.124, 6.653/5.696, 6.312/6.471, 6.106/6.528, 6.520/6.548, 5.632/6.173, 6.647/6.386, 5.827/6.333, 6.501/5.698, 6.701/6.102, 6.072/6.509, 5.925/6.397, 5.590/6.140, 5.555/6.240, 5.684/6.311, 5.607/6.152, 5.513/5.583, 5.524/6.215, 6.535/6.233, 6.698/5.886, 6.612/5.740, 6.627/6.370, 6.196/6.538, 5.541/6.080, 6.691/5.914, 6.536/6.472, 6.651/5.879, 6.738/6.039, 5.782/6.466, 5.496/5.584, 5.573/6.055, 6.505/5.974, 5.768/6.285, 6.610/5.983, 6.483/6.423, 5.625/5.564, 5.628/6.367, 5.674/6.464, 5.491/6.136, 6.637/6.292, 5.469/5.879, 5.510/5.930, 5.548/6.262, 6.521/6.389, 5.604/5.572, 6.624/6.265, 5.506/5.703, 5.519/6.002, 5.708/6.203, 6.026/6.575, 5.830/6.537, 6.731/5.877, 5.491/5.795, 6.514/5.689, 5.654/6.068, 5.859/6.466, 6.656/5.856, 5.851/6.436, 5.803/6.473, 6.674/5.681, 5.777/6.545, 5.781/6.414, 6.360/6.410, 5.604/6.101, 5.987/6.538, 6.625/6.184, 5.765/6.288, 5.754/6.305, 5.543/5.902, 6.565/6.423, 6.003/6.543, 6.707/5.981, 6.109/6.442, 5.693/6.287, 5.519/6.274, 6.635/5.666, 5.634/6.439, 5.593/5.960, 6.552/5.775, 6.431/6.468, 5.647/6.201, 6.610/5.826, 6.735/5.988, 6.654/6.132, 5.610/5.964, 6.574/6.174, 5.550/5.863, 6.556/5.984, 5.646/6.468, 5.946/6.445, 5.801/6.552, 6.409/6.374, 5.631/6.324, 5.513/5.669, 5.639/6.287, 6.658/5.992, 6.643/5.580, 6.379/6.403, 6.555/5.816, 6.271/6.416, 5.500/5.558, 6.607/6.416, 6.517/6.381, 5.807/6.350, 6.617/6.257, 6.520/6.014, 6.348/6.404, 5.495/6.177, 6.497/6.453, 5.531/5.775, 6.674/6.423, 6.573/5.581, 6.424/6.390, 6.332/6.591, 6.146/6.605, 6.582/5.565, 6.715/6.221, 6.589/6.389, 5.756/6.304, 6.246/6.510, 6.060/6.431, 6.674/5.794, 6.639/5.797, 5.738/6.363, 5.576/6.103, 6.529/6.432, 6.457/6.525, 6.529/5.700, 5.677/6.481, 6.551/6.304, 6.531/6.456, 5.934/6.417, 6.355/6.427, 5.940/6.509, 6.565/5.955, 6.078/6.515, 6.506/6.414, 6.572/5.774, 6.369/6.596, 6.447/6.430, 6.646/6.230, 5.502/5.982, 5.561/5.883, 6.526/6.209, 6.128/6.489, 6.683/6.019, 6.499/6.520, 6.053/6.435, 6.568/5.574, 6.582/5.725, 6.530/5.693, 6.458/6.570, 6.492/6.156, 6.291/6.554, 6.643/5.895, 6.580/6.454, 5.659/6.466, 6.672/5.642, 5.588/5.833, 5.508/5.948, 6.643/6.121, 6.648/6.468, 6.354/6.555, 6.658/6.023, 6.524/6.487, 5.572/5.845, 6.660/6.121, 5.656/6.307, 5.772/6.321, 6.698/6.237, 6.498/5.791, 5.868/6.408, 6.014/6.517, 6.664/5.718, 6.566/5.824, 6.115/6.594, 6.420/6.455, 6.557/6.387, 5.599/6.337, 6.587/5.739, 6.514/5.656, 6.406/6.578, 5.550/5.808, 6.132/6.547, 5.616/6.121
}

%% file: plots/points/points_n=256_k=256_sampling=SSC_ALL_shape=SIMPLE.tex
\def\sscallSimpleOutside{0.777/3.316, 0.663/3.692, 1.131/2.952, 0.580/3.902, 1.472/2.928, 1.659/4.094, 1.646/4.127, 1.750/3.843, 1.337/4.370, 1.555/4.245, 1.193/4.400, 1.756/3.830, 1.745/3.872, 1.651/4.125, 0.914/4.405, 0.547/4.203, 1.740/3.908, 1.347/4.378, 1.494/2.918, 0.538/3.965, 1.752/3.894, 0.958/4.419, 0.775/3.267, 0.764/3.286, 1.790/3.239, 1.794/3.696, 1.791/3.206, 0.595/3.786, 0.779/3.247, 0.765/3.267, 0.600/3.763, 1.806/3.643, 1.498/4.332, 0.573/4.285, 1.372/2.875, 0.621/3.691, 1.465/4.351, 0.642/3.572, 0.607/3.723, 1.477/4.347, 1.750/3.976, 0.583/4.305, 0.709/4.396, 1.150/2.902, 1.296/2.876, 0.513/3.962, 0.521/3.938, 1.441/4.368, 1.752/3.989, 1.824/3.560, 1.331/4.419, 1.773/3.067, 0.704/3.351, 1.362/2.860, 0.794/4.431, 0.553/4.284, 1.790/3.103, 1.398/2.857, 0.553/3.816, 0.552/3.817, 0.670/4.403, 0.922/4.457, 1.827/3.756, 0.608/3.635, 1.826/3.193, 0.767/4.436, 1.789/3.927, 0.493/4.218, 0.481/3.983, 0.538/4.290, 1.829/3.172, 1.345/2.844, 0.711/3.281, 1.399/4.416, 0.661/4.422, 1.815/3.887, 0.502/3.877, 1.783/4.009, 0.485/4.241, 0.830/4.473, 1.777/4.033, 0.771/3.148, 1.788/4.011, 0.631/4.426, 0.464/3.966, 0.728/4.452, 1.855/3.220, 1.432/2.822, 0.632/3.400, 1.824/3.072, 0.652/4.438, 0.772/3.132, 0.462/4.223, 1.726/2.943, 0.847/4.490, 1.044/2.871, 1.692/2.907, 0.862/4.500, 0.453/4.233, 0.814/3.047, 0.571/3.563, 1.861/3.843, 0.615/3.387, 0.445/4.233, 1.038/2.857, 1.822/2.998, 0.599/3.406, 1.888/3.233, 1.268/4.511, 1.858/3.079, 1.865/3.883, 1.262/2.793, 1.906/3.595, 0.557/3.529, 1.733/4.233, 1.906/3.603, 0.549/4.415, 1.220/4.524, 1.331/4.511, 1.907/3.631, 0.623/4.472, 0.800/3.020, 1.754/2.912, 0.473/4.318, 0.410/3.973, 1.454/4.466, 0.446/3.856, 1.910/3.235, 1.655/2.835, 1.461/2.776, 1.593/4.406, 0.410/3.947, 1.912/3.773, 1.192/2.777, 0.756/4.518, 0.498/3.704, 0.446/3.837, 0.536/3.526, 1.876/3.937, 0.584/4.473, 1.925/3.654, 0.430/3.866, 1.502/4.462, 1.928/3.321, 0.599/3.321, 0.547/4.456, 0.396/3.942, 0.442/3.808, 1.160/4.561, 1.643/4.381, 1.132/4.565, 0.997/4.567, 0.435/3.805, 0.838/2.945, 1.488/4.487, 0.602/3.268, 0.393/4.268, 0.373/4.204, 0.506/4.443, 1.909/3.944, 0.371/4.229, 1.909/3.039, 0.354/3.991, 0.562/3.330, 0.985/4.591, 0.384/4.282, 0.476/3.614, 1.630/2.773, 0.382/4.288, 1.873/2.934, 1.721/4.349, 1.970/3.676, 1.261/2.721, 0.476/3.587, 1.493/4.515, 1.967/3.226, 1.978/3.433, 0.599/3.208, 0.553/3.307, 0.801/2.928, 1.986/3.574, 1.386/4.573, 1.916/2.973, 1.962/3.871, 0.943/2.796, 0.675/4.571, 1.994/3.482, 0.659/4.569, 0.365/4.308, 0.410/4.373, 0.322/4.147, 1.363/2.688, 0.817/2.891, 0.588/3.180, 1.750/4.360, 0.312/3.991, 0.404/3.717, 0.322/4.206, 0.361/4.320, 2.002/3.301, 1.943/2.992, 1.610/4.501, 1.727/2.773, 1.337/2.679, 1.814/4.307, 1.439/4.580, 1.564/2.708, 1.961/3.962, 0.457/3.447, 0.385/4.373, 1.515/4.553, 1.618/2.720, 1.750/2.776, 1.208/4.640, 2.021/3.377, 1.727/2.758, 1.576/2.701, 1.716/4.432, 1.171/4.649, 1.108/2.693, 0.321/3.874, 1.959/4.026, 0.356/4.354, 0.461/3.385, 1.421/2.655, 1.897/4.184, 1.996/3.901, 2.004/3.102, 2.031/3.323, 2.035/3.447, 0.423/3.532, 0.460/4.512, 0.733/4.631, 0.326/3.830, 1.997/3.062, 1.887/4.224, 1.979/3.001, 1.757/4.410, 1.942/4.101, 1.779/4.387, 0.507/3.258, 0.436/3.418, 1.224/2.653, 2.038/3.266, 0.343/3.762, 1.576/2.679, 1.485/4.595, 0.317/4.324, 0.846/4.665, 0.425/3.442, 1.933/4.148, 0.807/2.831, 1.260/4.667, 1.887/4.277, 1.849/2.800, 1.962/2.907
}
\def\sscallSimpleInside{0.647/4.314, 0.985/3.054, 1.764/3.621, 0.813/4.369, 1.012/3.037, 0.842/4.375, 1.732/3.140, 1.631/4.117, 1.573/4.196, 0.781/3.338, 1.618/4.141, 0.633/4.274, 0.711/3.503, 0.824/3.265, 1.354/2.934, 1.614/4.138, 0.691/4.325, 0.972/4.375, 0.838/3.253, 1.647/4.048, 1.726/3.810, 1.674/3.059, 1.374/4.318, 1.737/3.265, 1.386/4.308, 0.579/4.111, 0.783/3.384, 1.738/3.620, 0.886/3.202, 1.034/3.054, 1.695/3.111, 0.925/4.356, 0.598/3.975, 0.638/4.236, 1.725/3.713, 0.946/4.353, 1.365/4.303, 0.775/4.322, 1.710/3.786, 1.074/3.033, 0.889/4.343, 0.614/3.962, 1.155/3.001, 1.499/2.996, 0.755/4.309, 0.713/3.712, 1.348/4.298, 0.893/3.228, 1.577/3.025, 0.733/3.629, 1.594/4.082, 0.811/3.389, 0.692/3.790, 1.702/3.265, 0.807/4.309, 0.919/3.207, 1.077/3.056, 0.621/4.004, 1.222/4.313, 0.662/3.904, 1.699/3.580, 1.347/2.993, 0.931/3.206, 0.697/4.252, 0.873/3.308, 1.361/2.997, 1.684/3.704, 1.574/4.068, 1.603/3.999, 1.040/3.111, 0.642/4.166, 1.148/4.311, 0.628/4.084, 0.652/4.178, 1.430/4.228, 1.680/3.278, 1.122/3.051, 0.851/3.376, 1.424/3.015, 1.504/4.150, 0.919/3.263, 1.573/3.074, 1.142/4.297, 0.783/3.581, 0.902/3.297, 1.059/4.301, 0.700/4.220, 0.768/3.692, 1.099/4.296, 0.705/4.217, 1.638/3.173, 0.820/3.490, 0.717/4.224, 0.883/3.357, 0.668/4.152, 1.652/3.407, 1.519/4.093, 1.539/3.081, 1.043/3.159, 1.460/4.151, 0.982/4.271, 1.634/3.241, 1.497/3.077, 1.566/3.111, 0.898/3.369, 0.760/3.823, 1.608/3.171, 0.680/4.049, 0.966/3.270, 0.994/4.257, 1.537/3.103, 1.579/3.903, 0.947/3.305, 0.825/3.598, 0.706/4.153, 1.626/3.598, 0.890/4.240, 0.824/3.678, 0.829/3.658, 0.762/3.874, 0.707/4.028, 1.467/4.094, 1.463/4.097, 1.058/4.243, 1.235/3.092, 0.778/3.856, 1.355/3.083, 1.105/4.234, 0.773/3.877, 0.753/3.929, 0.713/4.083, 1.602/3.457, 1.074/4.231, 1.033/3.232, 1.404/3.093, 1.320/4.179, 1.540/3.914, 1.581/3.221, 1.266/3.105, 1.157/3.141, 0.943/3.400, 1.375/4.142, 1.338/4.156, 1.576/3.736, 1.407/3.112, 0.923/3.459, 1.368/4.136, 1.543/3.173, 0.770/4.155, 1.576/3.460, 0.949/3.419, 0.755/4.017, 0.767/3.987, 1.562/3.707, 1.568/3.455, 0.900/3.540, 1.564/3.651, 1.559/3.692, 0.788/4.156, 1.306/3.130, 1.343/3.129, 1.380/3.128, 1.558/3.455, 1.532/3.191, 1.305/4.139, 0.825/3.884, 1.544/3.266, 0.913/3.550, 0.864/3.799, 1.363/4.104, 1.175/3.189, 1.540/3.608, 1.537/3.528, 1.115/3.247, 0.908/4.156, 1.533/3.402, 1.028/4.158, 0.802/4.008, 0.856/3.870, 0.816/3.981, 1.432/3.975, 1.253/3.176, 0.922/3.672, 1.520/3.375, 1.023/4.145, 0.821/3.997, 0.925/3.723, 0.998/3.465, 0.869/4.120, 0.891/3.838, 0.917/3.779, 0.811/4.071, 1.375/3.186, 1.228/3.206, 0.987/3.550, 1.178/4.103, 1.023/3.483, 1.050/4.106, 1.082/3.381, 0.962/3.715, 1.473/3.369, 1.440/3.769, 1.231/4.068, 1.062/3.454, 0.873/4.066, 1.453/3.388, 1.212/3.276, 0.931/3.895, 0.976/3.812, 1.003/3.697, 1.379/3.880, 1.250/3.270, 1.416/3.753, 0.933/3.970, 0.945/3.944, 1.245/4.014, 1.125/3.423, 1.412/3.427, 1.046/3.630, 1.187/3.360, 1.011/3.822, 1.300/3.291, 1.329/3.903, 1.298/3.960, 1.058/4.026, 1.379/3.728, 1.389/3.368, 1.305/3.933, 0.969/3.971, 1.371/3.715, 1.103/4.003, 1.367/3.675, 1.031/3.878, 1.356/3.624, 1.309/3.337, 1.070/3.789, 1.334/3.720, 1.341/3.636, 1.321/3.702, 1.151/3.948, 1.112/3.724, 1.064/3.904, 1.224/3.457, 1.309/3.695, 1.315/3.448, 1.119/3.770, 1.286/3.759, 1.179/3.558, 1.284/3.717, 1.095/3.920, 1.120/3.845, 1.125/3.896, 1.264/3.503, 1.171/3.831, 1.176/3.850, 1.196/3.762
}

%% file: plots/points/points_n=256_k=256_sampling=SSC_ALL_shape=SMALL.tex
\def\sscallSmallOutside{3.656/4.007, 3.400/3.983, 3.517/3.369, 3.917/3.796, 3.942/3.404, 3.858/3.294, 3.392/3.650, 3.446/3.472, 3.928/3.771, 3.909/3.825, 3.341/3.917, 3.935/3.369, 3.652/4.014, 3.629/4.017, 3.587/4.021, 3.326/3.847, 3.428/3.501, 3.383/3.663, 3.907/3.844, 3.940/3.750, 3.390/3.615, 3.493/3.379, 3.694/4.017, 3.929/3.333, 3.881/3.904, 3.896/3.878, 3.854/3.934, 3.459/3.426, 3.882/3.906, 3.385/3.622, 3.964/3.544, 3.947/3.745, 3.368/3.976, 3.560/3.310, 3.965/3.472, 3.821/3.974, 3.963/3.689, 3.745/3.252, 3.327/3.782, 3.886/3.909, 3.956/3.728, 3.938/3.332, 3.920/3.316, 3.815/3.262, 3.386/3.580, 3.514/4.031, 3.339/3.739, 3.414/3.492, 3.801/3.989, 3.514/3.339, 3.960/3.385, 3.792/3.251, 3.681/3.249, 3.380/3.587, 3.698/4.030, 3.957/3.362, 3.904/3.896, 3.886/3.924, 3.980/3.559, 3.724/4.031, 3.578/4.046, 3.361/3.997, 3.465/3.380, 3.752/4.023, 3.986/3.603, 3.680/4.043, 3.518/4.046, 3.882/3.267, 3.678/3.236, 3.338/3.697, 3.894/3.928, 3.348/3.988, 3.623/3.249, 3.371/3.550, 3.364/3.583, 3.290/3.852, 3.893/3.936, 3.302/3.928, 3.456/3.375, 3.784/3.229, 3.391/4.028, 3.975/3.752, 3.515/4.054, 3.354/3.624, 3.996/3.605, 3.313/3.739, 3.338/3.987, 3.946/3.302, 3.886/3.258, 3.927/3.892, 3.361/4.019, 3.967/3.325, 3.373/3.521, 3.759/4.037, 3.948/3.853, 3.933/3.888, 3.677/3.221, 3.682/3.220, 3.601/3.241, 4.000/3.675, 3.621/3.234, 3.349/4.016, 3.999/3.699, 3.660/4.064, 3.650/3.223, 3.974/3.325, 3.834/4.010, 3.544/4.069, 4.006/3.486, 3.770/3.211, 3.513/3.295, 3.538/3.274, 3.274/3.854, 3.903/3.255, 3.385/3.468, 3.291/3.945, 4.006/3.692, 3.977/3.312, 4.003/3.723, 3.781/3.207, 3.455/3.340, 4.010/3.667, 3.994/3.758, 3.406/3.416, 3.970/3.836, 3.411/4.055, 3.859/3.228, 4.016/3.538, 3.514/3.283, 3.944/3.904, 3.461/3.326, 4.016/3.459, 3.263/3.851, 3.273/3.928, 3.340/3.557, 3.665/4.079, 3.261/3.839, 3.916/3.250, 3.336/3.573, 4.022/3.601, 4.023/3.608, 3.297/3.708, 3.496/3.290, 4.023/3.572, 3.989/3.314, 4.000/3.347, 3.688/3.200, 3.263/3.894, 3.736/3.196, 3.551/4.087, 3.326/3.607, 3.265/3.921, 3.992/3.809, 3.931/3.255, 3.274/3.760, 3.291/3.713, 3.393/3.412, 3.282/3.731, 3.918/3.243, 4.016/3.732, 3.946/3.929, 3.703/4.085, 4.014/3.746, 3.686/4.087, 3.301/3.994, 3.695/4.087, 3.941/3.253, 3.981/3.283, 3.678/4.092, 3.973/3.885, 3.904/3.221, 4.037/3.591, 4.038/3.593, 3.597/3.206, 3.241/3.835, 4.039/3.537, 3.627/4.103, 3.438/4.088, 3.980/3.879, 3.380/3.408, 3.939/3.242, 3.353/3.465, 3.422/4.085, 3.387/4.076, 3.310/4.028, 3.980/3.885, 3.364/3.435, 3.239/3.852, 3.332/3.503, 3.611/4.108, 4.004/3.298, 3.952/3.247, 3.766/4.082, 3.364/3.433, 3.500/3.258, 3.989/3.275, 3.970/3.914, 4.042/3.671, 3.329/4.055, 3.855/3.194, 4.045/3.463, 3.244/3.930, 4.043/3.689, 3.362/3.428, 3.886/4.024, 3.985/3.889, 3.583/3.199, 3.331/3.490, 3.389/3.375, 3.386/3.380, 4.048/3.664, 3.923/3.216, 3.269/3.987, 3.337/4.072, 4.054/3.480, 3.648/4.116, 3.955/3.235, 3.250/3.732, 3.297/3.584, 4.054/3.669, 3.377/4.090, 3.384/3.367, 3.359/4.086, 3.250/3.971, 3.330/3.468, 3.273/3.661, 3.665/4.121, 3.905/4.025, 3.286/3.618, 3.221/3.800, 4.027/3.299, 3.809/3.162, 3.217/3.851, 4.049/3.359, 3.764/4.107, 3.276/3.639, 3.482/4.125, 3.441/3.278, 3.945/3.991, 3.354/3.397, 3.227/3.949, 4.029/3.288, 3.838/3.165, 3.348/3.406, 3.945/3.994, 3.766/3.148, 4.072/3.512, 3.689/3.150, 3.364/4.102, 3.330/4.090, 4.038/3.299, 3.630/3.157, 3.222/3.760, 4.076/3.598, 3.805/4.099, 3.930/3.192
}
\def\sscallSmallInside{3.914/3.797, 3.900/3.330, 3.595/3.306, 3.946/3.508, 3.564/3.333, 3.931/3.733, 3.710/3.998, 3.408/3.604, 3.943/3.520, 3.943/3.584, 3.882/3.860, 3.905/3.807, 3.349/3.797, 3.578/3.327, 3.941/3.610, 3.451/3.492, 3.399/3.676, 3.869/3.879, 3.933/3.691, 3.506/3.400, 3.843/3.307, 3.908/3.772, 3.757/3.285, 3.815/3.300, 3.818/3.302, 3.538/3.372, 3.911/3.369, 3.355/3.815, 3.812/3.932, 3.928/3.468, 3.930/3.603, 3.411/3.673, 3.544/3.372, 3.438/3.546, 3.928/3.555, 3.429/3.583, 3.421/3.631, 3.927/3.620, 3.440/3.971, 3.734/3.967, 3.357/3.841, 3.364/3.812, 3.617/3.320, 3.889/3.351, 3.461/3.516, 3.580/3.985, 3.921/3.578, 3.915/3.433, 3.402/3.945, 3.553/3.376, 3.545/3.383, 3.855/3.864, 3.460/3.967, 3.917/3.614, 3.387/3.767, 3.371/3.814, 3.507/3.441, 3.439/3.602, 3.856/3.338, 3.893/3.373, 3.912/3.514, 3.422/3.951, 3.558/3.384, 3.595/3.976, 3.440/3.628, 3.591/3.359, 3.417/3.712, 3.658/3.966, 3.898/3.408, 3.381/3.891, 3.905/3.541, 3.671/3.321, 3.904/3.638, 3.489/3.495, 3.864/3.357, 3.396/3.780, 3.430/3.692, 3.899/3.470, 3.419/3.937, 3.451/3.946, 3.402/3.911, 3.431/3.940, 3.409/3.916, 3.831/3.345, 3.886/3.411, 3.743/3.326, 3.836/3.837, 3.429/3.932, 3.504/3.495, 3.584/3.391, 3.437/3.720, 3.418/3.772, 3.873/3.405, 3.850/3.372, 3.878/3.437, 3.412/3.892, 3.518/3.488, 3.503/3.521, 3.404/3.872, 3.405/3.875, 3.872/3.416, 3.480/3.574, 3.431/3.761, 3.874/3.635, 3.779/3.350, 3.868/3.695, 3.477/3.616, 3.408/3.846, 3.479/3.600, 3.520/3.500, 3.596/3.402, 3.805/3.853, 3.415/3.865, 3.806/3.364, 3.862/3.421, 3.506/3.547, 3.514/3.532, 3.837/3.768, 3.563/3.446, 3.800/3.848, 3.861/3.664, 3.864/3.515, 3.605/3.406, 3.549/3.924, 3.859/3.467, 3.530/3.509, 3.435/3.796, 3.686/3.917, 3.585/3.925, 3.558/3.462, 3.790/3.852, 3.833/3.764, 3.594/3.420, 3.455/3.751, 3.521/3.533, 3.729/3.896, 3.609/3.922, 3.854/3.675, 3.475/3.708, 3.724/3.893, 3.497/3.619, 3.478/3.706, 3.509/3.574, 3.850/3.466, 3.774/3.856, 3.821/3.769, 3.444/3.878, 3.596/3.442, 3.794/3.810, 3.701/3.381, 3.469/3.765, 3.711/3.381, 3.731/3.874, 3.762/3.850, 3.511/3.896, 3.741/3.865, 3.693/3.390, 3.538/3.558, 3.638/3.418, 3.830/3.479, 3.489/3.885, 3.528/3.588, 3.624/3.890, 3.545/3.558, 3.461/3.828, 3.654/3.413, 3.822/3.480, 3.563/3.886, 3.782/3.788, 3.683/3.876, 3.596/3.886, 3.661/3.415, 3.817/3.463, 3.537/3.879, 3.623/3.876, 3.739/3.408, 3.812/3.588, 3.806/3.680, 3.810/3.493, 3.666/3.866, 3.524/3.714, 3.647/3.867, 3.807/3.523, 3.675/3.424, 3.807/3.561, 3.788/3.425, 3.771/3.767, 3.698/3.848, 3.598/3.517, 3.797/3.622, 3.737/3.815, 3.553/3.636, 3.493/3.850, 3.544/3.695, 3.499/3.823, 3.782/3.690, 3.570/3.588, 3.540/3.722, 3.561/3.631, 3.766/3.737, 3.770/3.716, 3.588/3.851, 3.702/3.436, 3.567/3.624, 3.605/3.532, 3.780/3.654, 3.575/3.600, 3.567/3.642, 3.766/3.703, 3.773/3.674, 3.775/3.657, 3.725/3.799, 3.775/3.624, 3.539/3.771, 3.564/3.708, 3.724/3.787, 3.583/3.646, 3.541/3.826, 3.532/3.815, 3.578/3.685, 3.597/3.599, 3.558/3.764, 3.753/3.670, 3.741/3.701, 3.709/3.778, 3.582/3.719, 3.730/3.720, 3.599/3.667, 3.607/3.808, 3.564/3.793, 3.648/3.800, 3.599/3.710, 3.606/3.694, 3.733/3.508, 3.674/3.519, 3.583/3.774, 3.725/3.669, 3.709/3.724, 3.639/3.588, 3.678/3.526, 3.593/3.767, 3.644/3.782, 3.618/3.784, 3.676/3.752, 3.611/3.764, 3.708/3.588, 3.648/3.642, 3.626/3.749, 3.673/3.583, 3.698/3.581, 3.651/3.687, 3.667/3.618, 3.670/3.716, 3.685/3.656, 3.658/3.735, 3.653/3.726
}

%% file: plots/points/points_n=256_k=256_sampling=SSC_ALL_shape=THIN.tex
\def\sscallThinOutside{6.604/3.047, 5.499/3.780, 6.491/3.195, 6.500/4.080, 6.393/3.873, 6.717/3.853, 6.696/3.091, 5.532/3.025, 5.674/3.500, 5.462/3.429, 5.654/3.247, 6.366/4.103, 5.456/3.295, 6.485/3.625, 5.909/4.099, 6.745/3.441, 5.689/3.591, 6.387/4.106, 6.744/3.702, 6.733/3.194, 5.685/3.543, 6.740/3.740, 6.346/3.894, 5.516/3.886, 6.448/3.783, 5.633/3.057, 5.564/3.016, 5.449/3.263, 6.057/4.117, 5.471/3.061, 6.750/3.347, 6.297/4.113, 6.660/3.997, 6.751/3.634, 5.744/4.067, 6.485/4.096, 6.726/3.875, 6.170/3.897, 5.451/3.145, 5.894/4.107, 5.747/3.696, 5.776/3.737, 5.703/3.629, 5.449/3.148, 6.755/3.495, 6.752/3.711, 6.493/4.098, 6.737/3.818, 6.142/3.891, 6.736/3.130, 5.976/4.117, 6.165/4.125, 5.621/4.018, 5.460/3.647, 6.587/3.023, 6.759/3.505, 6.759/3.522, 6.746/3.205, 6.599/4.084, 5.868/4.113, 6.138/4.130, 6.706/3.955, 6.202/3.887, 6.469/3.322, 6.387/4.122, 6.536/3.046, 6.467/3.457, 6.241/4.128, 5.439/3.169, 5.696/3.494, 5.720/4.070, 6.484/4.108, 6.466/3.207, 6.764/3.599, 5.761/4.089, 6.387/4.125, 5.706/3.562, 6.719/3.078, 6.023/3.869, 6.043/3.871, 6.052/4.135, 5.713/3.594, 6.095/4.139, 5.477/3.815, 5.845/3.802, 6.469/3.112, 6.659/3.032, 5.724/3.632, 5.600/2.997, 6.459/3.424, 6.032/4.137, 6.464/3.125, 6.457/3.516, 5.441/3.589, 5.680/3.114, 6.383/3.834, 5.704/3.468, 6.454/3.442, 6.594/4.102, 5.683/3.106, 5.438/3.631, 6.514/4.118, 5.707/3.460, 5.694/3.275, 5.533/3.964, 5.418/3.265, 5.703/4.080, 5.630/4.049, 5.461/3.016, 6.588/4.109, 5.697/3.287, 5.749/3.648, 5.708/3.451, 6.712/3.982, 6.661/4.047, 5.822/4.132, 6.484/3.050, 6.445/3.514, 5.663/3.037, 6.627/4.093, 5.579/4.020, 6.662/4.052, 5.417/3.144, 6.443/3.460, 5.609/4.046, 6.776/3.780, 6.708/3.996, 5.451/3.797, 5.726/3.525, 6.791/3.495, 5.699/3.160, 6.255/4.156, 6.739/3.060, 6.435/3.542, 6.437/3.173, 5.696/3.064, 6.536/4.131, 5.702/3.118, 6.240/3.851, 6.735/3.975, 5.767/4.127, 6.486/4.140, 5.856/3.771, 6.430/3.664, 6.783/3.804, 6.431/3.233, 6.429/3.606, 6.683/3.009, 6.787/3.782, 6.667/3.003, 6.125/4.170, 5.838/3.741, 6.791/3.245, 6.265/4.165, 6.422/3.676, 6.189/4.169, 6.432/3.129, 5.403/3.458, 5.443/3.822, 6.425/3.574, 5.720/4.114, 6.426/3.288, 5.773/4.138, 6.140/4.174, 6.790/3.796, 5.747/3.570, 5.875/4.159, 6.424/3.184, 5.776/4.142, 6.091/4.178, 6.785/3.853, 6.782/3.093, 6.615/4.129, 5.731/3.415, 5.410/3.653, 6.567/2.976, 5.398/3.485, 6.477/3.025, 5.452/3.883, 6.421/3.278, 6.423/3.134, 6.134/4.179, 5.716/3.113, 6.152/3.837, 6.459/3.036, 5.463/3.930, 6.416/3.597, 6.568/4.143, 5.758/3.595, 6.793/3.837, 5.439/2.992, 5.734/3.366, 5.737/3.413, 6.413/3.638, 6.414/3.235, 6.359/3.800, 5.398/3.601, 6.817/3.599, 5.568/2.948, 6.369/3.790, 5.402/3.667, 6.412/3.160, 5.722/3.064, 6.408/3.595, 6.821/3.446, 6.187/4.188, 5.719/3.048, 5.385/3.116, 5.387/3.511, 5.820/3.681, 6.765/3.045, 6.777/3.058, 6.752/3.029, 6.406/3.337, 6.811/3.188, 6.815/3.232, 5.451/3.939, 5.767/3.564, 6.403/3.456, 6.672/4.104, 6.732/4.028, 5.457/3.955, 5.451/3.945, 5.638/2.949, 6.831/3.466, 5.909/3.774, 5.383/3.586, 6.666/4.117, 6.350/3.788, 5.488/3.992, 5.774/4.169, 5.451/3.952, 5.378/3.551, 5.753/3.344, 5.860/4.187, 5.642/4.113, 6.482/2.992, 5.367/3.410, 5.391/3.753, 6.497/4.180, 5.588/2.929, 6.392/3.192, 6.336/3.792, 6.465/2.999, 5.734/3.039, 6.757/3.012, 6.603/4.165, 5.383/3.717, 5.359/3.300, 6.814/3.892, 6.000/3.790, 5.753/3.225, 6.765/4.011, 5.357/3.288, 5.606/2.925, 5.432/3.936
}
\def\sscallThinInside{5.975/4.098, 6.142/4.106, 6.434/4.088, 6.460/3.788, 5.602/3.045, 6.735/3.433, 6.734/3.538, 6.118/3.912, 6.706/3.114, 5.638/3.093, 5.641/3.188, 5.577/3.939, 6.461/4.078, 5.468/3.272, 6.706/3.118, 6.271/3.916, 6.720/3.748, 6.730/3.405, 6.422/4.082, 6.502/3.510, 5.478/3.441, 5.642/3.314, 6.549/4.059, 6.313/4.088, 6.140/4.094, 5.644/3.358, 6.107/3.919, 5.487/3.543, 5.969/3.902, 6.215/4.090, 5.493/3.079, 6.203/4.091, 6.011/4.088, 5.497/3.687, 6.587/3.061, 6.504/3.663, 5.480/3.377, 6.240/4.087, 5.491/3.566, 6.426/4.075, 6.018/4.084, 5.632/3.243, 6.510/3.162, 6.538/3.094, 6.719/3.447, 5.576/3.918, 5.492/3.515, 6.479/3.794, 6.678/3.917, 5.614/3.086, 6.038/3.923, 6.604/4.008, 6.517/3.290, 6.009/3.922, 5.493/3.476, 6.374/3.924, 6.466/3.846, 5.822/3.869, 5.658/3.626, 5.671/3.988, 5.631/3.387, 5.868/3.887, 6.709/3.629, 6.056/3.930, 6.635/3.088, 5.637/3.473, 5.510/3.657, 5.632/3.441, 5.620/3.252, 6.522/3.431, 5.495/3.381, 5.577/3.904, 6.524/3.354, 5.621/3.299, 6.478/4.049, 5.660/3.978, 6.522/3.680, 5.503/3.465, 5.761/4.021, 5.499/3.352, 6.526/3.604, 5.722/4.003, 6.033/4.067, 5.502/3.391, 5.558/3.859, 5.981/3.931, 5.649/3.967, 6.045/4.065, 6.688/3.759, 5.643/3.602, 5.606/3.161, 6.277/4.060, 5.835/4.045, 5.681/3.701, 6.535/3.265, 6.578/4.010, 6.092/4.065, 5.525/3.674, 6.652/3.915, 6.584/4.000, 5.606/3.261, 6.076/3.949, 6.177/3.958, 5.609/3.337, 6.543/3.138, 6.522/4.028, 5.845/4.042, 5.617/3.457, 6.674/3.809, 5.679/3.706, 6.077/4.060, 6.469/4.035, 6.674/3.172, 6.540/3.187, 6.539/3.587, 5.892/4.043, 5.515/3.402, 6.621/3.945, 6.688/3.385, 5.705/3.979, 6.006/4.049, 5.531/3.611, 6.543/3.628, 5.626/3.572, 6.304/4.047, 6.654/3.891, 6.058/4.049, 6.608/3.108, 5.561/3.790, 6.665/3.170, 6.552/3.370, 6.678/3.540, 6.214/3.971, 6.590/3.965, 6.128/4.046, 6.677/3.536, 5.541/3.630, 5.586/3.227, 5.622/3.905, 5.525/3.228, 6.674/3.635, 5.587/3.280, 6.217/3.975, 6.625/3.123, 6.557/3.270, 5.700/3.769, 5.952/3.952, 5.768/3.862, 5.917/3.943, 6.332/4.032, 5.578/3.173, 6.671/3.467, 5.539/3.513, 6.559/3.661, 5.534/3.174, 6.380/4.027, 6.563/3.419, 5.922/3.948, 6.338/3.976, 6.173/3.984, 6.564/3.452, 5.595/3.482, 5.573/3.143, 6.536/3.778, 6.656/3.736, 5.571/3.756, 5.585/3.416, 6.531/3.806, 6.592/3.939, 6.393/4.019, 6.451/3.927, 5.538/3.258, 6.649/3.764, 6.660/3.585, 6.180/3.990, 5.974/4.020, 6.230/4.025, 6.447/4.006, 5.957/3.968, 6.657/3.665, 5.783/3.981, 5.895/3.950, 5.594/3.548, 5.755/3.870, 6.351/4.016, 5.686/3.936, 5.566/3.237, 5.585/3.496, 6.212/4.020, 6.578/3.153, 6.647/3.730, 5.764/3.969, 6.576/3.388, 6.622/3.142, 5.581/3.476, 5.977/4.012, 6.628/3.846, 5.559/3.511, 6.041/3.988, 6.633/3.796, 6.081/3.993, 5.561/3.253, 5.564/3.313, 6.136/4.017, 5.758/3.960, 6.556/3.770, 6.190/4.002, 5.584/3.745, 6.399/3.985, 6.647/3.521, 6.577/3.696, 5.570/3.623, 6.160/4.003, 6.599/3.904, 6.116/4.015, 6.562/3.950, 6.233/4.003, 5.832/3.988, 5.842/3.949, 6.573/3.725, 6.587/3.216, 6.543/3.970, 6.386/3.996, 5.700/3.823, 5.803/3.940, 6.060/4.001, 6.638/3.561, 6.590/3.640, 5.578/3.604, 5.747/3.939, 6.627/3.725, 6.624/3.252, 5.592/3.680, 5.847/3.982, 5.926/3.987, 5.668/3.798, 6.524/3.904, 6.601/3.309, 5.867/3.979, 5.636/3.757, 5.771/3.940, 5.773/3.941, 5.665/3.875, 5.619/3.739, 6.608/3.216, 6.609/3.226, 5.623/3.789, 6.607/3.774, 6.595/3.736, 6.568/3.842, 6.613/3.417, 6.617/3.495, 6.491/3.956, 6.557/3.885, 5.630/3.775, 6.513/3.938
}

%% file: plots/points/points_n=256_k=512_sampling=UNIFORM_shape=SIMPLE.tex
\def\uniformSimpleOutside{1.379/7.903, 0.284/8.342, 0.356/8.060, 0.171/7.798, 1.844/7.946, 1.552/9.336, 0.713/8.144, 0.624/7.712, 2.000/8.476, 2.095/8.476, 1.946/8.301, 0.270/8.869, 0.517/7.644, 0.686/9.424, 0.199/8.306, 1.907/9.355, 2.138/8.961, 0.953/9.696, 0.333/8.181, 0.515/8.542, 1.702/7.600, 1.850/7.843, 0.748/7.848, 0.772/9.663, 0.588/8.725, 0.493/9.108, 2.138/9.463, 0.266/8.395, 0.799/9.654, 0.258/7.651, 0.789/7.570, 1.664/7.643, 1.119/7.934, 0.103/8.031, 1.235/7.907, 1.049/7.817, 0.861/8.133, 0.226/9.431, 2.181/8.965, 0.463/8.896, 1.130/7.784, 0.683/8.062, 2.162/8.054, 0.131/8.881, 1.899/8.031, 1.824/8.096, 2.166/7.664, 0.702/9.473, 1.826/8.479, 0.917/7.789, 1.583/9.488, 1.028/7.953, 1.897/9.383, 1.906/7.651, 1.897/9.112, 1.536/9.667, 2.177/8.367, 2.188/9.347, 1.777/8.308, 2.137/9.156, 2.123/9.296, 0.852/8.154, 0.349/8.769, 0.512/7.844, 0.252/9.343, 0.937/9.477, 0.319/7.899, 1.977/8.424, 1.792/9.616, 1.712/9.162, 1.375/7.592, 1.889/9.204, 2.035/9.626, 1.969/8.115, 0.178/8.296, 0.991/7.582, 2.100/8.383, 1.589/9.525, 0.627/8.229, 0.589/8.641, 0.322/9.300, 0.723/7.581, 1.087/9.535, 1.414/7.729, 1.449/7.906, 1.772/8.792, 0.629/9.486, 2.142/8.395, 1.918/9.411, 1.415/9.662, 0.386/7.628, 1.779/9.137, 0.226/9.315, 0.367/9.315, 0.267/9.000, 1.167/9.451, 0.371/9.387, 1.090/7.787, 2.099/7.623, 0.105/7.713, 0.212/7.913, 0.748/7.866, 0.425/8.428, 1.505/9.399, 0.264/8.820, 0.140/9.438, 0.487/8.390, 2.148/8.691, 0.260/9.430, 1.829/9.188, 1.537/9.386, 0.432/8.658, 0.464/7.884, 0.902/8.029, 1.749/9.441, 1.652/9.339, 0.299/9.688, 0.066/9.464, 2.087/9.275, 1.789/8.893, 2.163/7.930, 1.982/8.359, 2.055/8.835, 0.446/8.328, 0.152/8.757, 0.958/9.633, 0.464/9.660, 0.190/7.840, 2.017/7.568, 0.205/9.393, 0.511/9.463, 1.869/8.416, 1.317/9.460, 1.893/7.952, 1.656/7.624, 0.159/9.582, 0.466/8.051, 1.918/8.985, 1.732/9.413, 1.284/9.599, 0.523/8.846, 2.014/8.346, 0.569/7.988, 1.976/9.534, 0.837/9.427, 0.892/7.627, 0.456/8.830, 0.506/9.226, 2.165/9.395, 0.873/8.092, 0.096/9.040, 0.161/8.947, 0.073/8.156, 0.171/9.627, 0.081/8.214, 1.030/9.555, 0.343/9.320, 0.179/9.070, 1.980/9.093, 1.155/7.813, 2.195/8.436, 0.461/7.824, 2.198/9.278, 0.196/7.718, 0.830/8.218, 0.221/9.321, 0.815/7.841, 0.343/8.008, 1.035/7.865, 0.207/8.462, 1.647/7.952, 1.298/7.829, 0.214/9.001, 0.645/8.379, 2.072/7.903, 1.504/9.433, 1.230/7.834, 1.384/7.879, 0.099/9.155, 2.070/8.090, 1.947/9.646, 1.933/9.049, 1.896/7.710, 0.150/9.658, 0.309/9.119, 1.911/8.252, 2.040/8.834, 0.695/7.758, 0.594/8.861, 0.117/8.986, 2.050/8.036, 0.928/7.564, 1.291/9.539, 0.517/8.502, 0.415/7.625, 1.698/9.442, 1.770/9.540, 0.702/8.325, 1.910/8.164, 1.888/9.271, 2.186/9.594, 1.387/7.887, 1.861/9.260, 0.302/8.451, 1.862/8.510, 0.316/7.919, 0.566/8.900, 0.286/8.669, 1.784/9.648, 0.065/8.692, 1.828/8.623, 0.151/8.495, 0.840/8.148, 0.145/7.560, 2.009/8.490, 1.792/7.876, 1.946/8.253, 0.167/8.564, 0.421/9.329, 1.863/7.669, 2.177/7.713, 0.767/9.371, 1.163/9.680, 1.951/8.148, 0.763/8.236, 0.178/8.622, 0.178/7.960, 1.228/7.714, 2.045/9.664, 1.883/9.436, 1.990/9.692, 0.136/9.660, 1.318/7.593, 0.323/9.242, 0.647/8.637, 1.880/9.061, 0.409/7.796, 0.421/8.907, 1.593/9.587, 1.962/9.145, 2.184/8.684, 1.185/9.672, 1.594/9.433, 2.087/9.075, 0.053/8.862, 0.389/9.461, 1.822/9.675, 0.612/7.936, 1.994/9.512, 0.553/9.631, 1.374/9.350, 0.290/7.798, 0.441/7.782, 1.428/7.723, 0.786/7.601, 0.639/9.586, 2.125/8.480, 1.957/9.513, 0.858/7.986, 1.156/7.839, 2.095/9.474, 1.814/9.354, 0.430/8.971, 1.973/8.211, 0.220/7.614, 1.070/9.646, 0.415/9.510, 0.180/8.331, 0.308/7.837, 0.185/7.951, 1.824/8.909, 0.766/8.332, 0.237/8.348, 0.725/7.968, 0.402/7.774, 1.808/8.064, 2.053/8.521, 1.272/9.579, 0.311/8.729, 2.137/9.358, 0.677/8.480, 2.165/9.506, 1.617/9.248, 0.127/9.100, 2.106/9.500, 0.192/8.986, 1.984/9.609, 0.398/8.155, 1.711/9.564, 0.061/8.460, 0.223/9.029, 0.600/9.515, 0.583/9.574, 0.206/7.955, 1.673/7.843, 1.132/9.582, 0.380/9.040, 0.246/7.720, 2.050/7.763, 1.804/8.380, 2.001/7.721, 0.370/9.615, 2.093/9.546, 2.198/9.610, 0.467/8.600, 0.480/9.328, 0.977/7.839, 0.360/9.612, 0.749/7.733, 0.708/8.358, 0.530/9.478, 1.852/7.748, 0.510/8.533, 0.606/8.015, 1.882/9.348, 1.004/7.785, 0.983/9.510, 1.815/8.428, 2.043/9.667, 1.991/8.919, 0.139/8.651, 0.852/7.667, 1.972/7.778, 2.146/7.616, 1.701/9.672, 0.959/7.713, 0.159/9.597, 1.907/7.639, 1.981/8.992, 2.052/9.066, 1.852/8.052, 0.079/9.471, 2.065/8.624, 1.788/7.606, 1.768/9.403, 0.334/9.539, 0.806/9.625, 0.276/9.488, 0.327/8.026, 1.947/8.838, 1.593/9.546, 0.576/8.857, 0.368/8.843, 1.144/9.531, 0.077/8.580, 0.161/9.044, 1.763/9.623, 0.987/9.689, 1.869/8.097, 0.247/8.213, 0.964/9.556, 0.668/8.516, 1.677/9.104, 0.775/7.664, 1.764/8.217, 1.204/9.599, 1.985/8.706
}
\def\uniformSimpleInside{1.014/8.108, 1.105/9.294, 0.793/8.992, 0.966/8.456, 1.558/9.019, 1.532/8.247, 0.764/8.871, 0.602/9.022, 1.508/9.135, 1.327/8.588, 1.371/8.276, 1.166/9.155, 1.640/8.086, 1.667/8.104, 0.835/9.125, 1.758/8.221, 1.337/9.101, 0.772/8.463, 1.447/8.851, 1.272/8.823, 1.070/9.307, 1.376/8.810, 0.677/8.925, 1.090/9.193, 1.272/8.097, 0.964/8.651, 0.827/8.847, 0.839/9.248, 1.118/9.244, 0.974/8.215, 0.874/8.811, 0.609/9.082, 0.983/9.065, 1.203/8.673, 1.277/9.214, 1.526/9.039, 1.000/8.352, 1.332/9.282, 1.244/8.123, 1.476/8.405, 1.162/9.004, 1.115/8.386, 0.589/9.198, 1.241/8.042, 0.868/8.393, 1.675/8.301, 1.155/8.487, 1.337/9.187, 1.026/8.450, 1.092/9.099, 1.324/7.985, 1.401/8.891, 1.725/8.352, 1.440/9.284, 0.871/8.857, 1.702/8.305, 0.678/9.153, 1.395/8.399, 1.311/8.784, 1.069/8.925, 1.700/8.134, 0.762/8.697, 1.539/8.733, 1.603/9.056, 0.702/8.757, 1.089/9.228, 0.960/8.309, 0.909/8.671, 1.048/8.424, 1.201/9.214, 1.567/9.162, 1.505/8.331, 1.593/9.165, 1.492/8.552, 1.708/8.220, 1.576/8.866, 0.748/8.831, 1.461/8.770, 0.941/8.213, 1.410/8.046, 1.337/8.729, 1.589/8.899, 0.949/8.369, 0.879/8.262, 1.626/9.053, 1.251/8.699, 1.286/8.985, 1.335/9.308, 1.099/9.007, 1.134/9.096, 1.568/8.374, 1.113/8.732, 0.874/9.049, 0.767/8.357, 1.371/7.970, 0.985/8.771, 1.024/8.442, 0.758/8.454, 1.349/8.128, 1.468/8.158, 1.346/9.166, 0.818/8.878, 1.209/8.656, 1.294/9.265, 1.010/9.077, 0.696/8.799, 1.086/9.150, 1.385/8.621, 1.199/9.266, 1.654/8.257, 0.838/8.618, 0.639/9.264, 1.532/9.220, 1.258/9.352, 1.525/8.639, 0.890/9.209, 1.227/9.074, 0.924/8.158, 0.904/8.546, 1.267/9.076, 0.848/8.359, 0.910/8.755, 1.003/8.315, 1.318/8.045, 0.959/8.138, 1.468/8.550, 1.030/8.133, 1.396/8.282, 1.453/8.855, 0.950/8.780, 0.560/9.075, 0.980/8.391, 1.016/8.589, 1.569/8.755, 1.383/8.812, 0.822/8.764, 1.586/8.959, 1.301/8.062, 1.334/8.891, 0.743/8.794, 0.922/8.398, 1.370/8.685, 0.826/8.753, 0.887/8.515, 1.369/7.978, 1.680/8.198, 0.934/8.475, 1.023/9.344, 1.274/8.911, 0.946/8.315, 0.711/8.962, 1.224/9.373, 1.444/8.380, 0.834/8.348, 1.418/8.720
}

%% file: plots/points/points_n=256_k=512_sampling=UNIFORM_shape=SMALL.tex
\def\uniformSmallOutside{4.126/9.240, 3.785/8.084, 4.667/8.225, 2.735/8.562, 4.313/7.897, 3.242/9.202, 3.607/9.503, 3.706/7.593, 3.435/9.398, 3.610/8.091, 4.082/8.301, 4.113/8.310, 2.555/9.523, 4.467/9.656, 4.456/8.223, 4.012/8.215, 3.019/8.714, 3.229/8.802, 2.933/7.577, 3.755/9.671, 3.343/7.677, 3.176/9.492, 2.812/7.865, 4.432/9.463, 2.704/9.444, 3.888/7.595, 4.612/9.655, 4.083/9.342, 4.188/8.158, 3.234/8.541, 3.038/8.174, 4.268/9.164, 4.321/8.854, 2.821/9.160, 2.843/7.837, 2.692/8.430, 3.737/7.632, 2.813/9.375, 2.704/7.594, 3.978/9.529, 2.679/9.218, 2.656/8.394, 3.756/8.199, 4.397/8.369, 3.148/8.671, 4.515/8.923, 4.143/9.461, 4.229/8.901, 4.543/7.595, 3.154/7.632, 4.034/8.735, 4.016/9.239, 2.695/7.660, 4.304/7.986, 3.181/8.995, 4.294/7.802, 4.312/7.619, 3.791/9.137, 2.754/9.479, 3.261/8.701, 4.594/9.615, 4.483/8.324, 4.577/9.084, 4.421/8.199, 4.425/8.360, 3.774/9.394, 4.400/8.181, 2.842/8.173, 3.774/9.144, 3.366/7.871, 4.360/7.627, 4.223/7.898, 4.216/9.213, 4.250/9.242, 4.206/8.956, 4.277/8.920, 4.153/9.482, 4.437/7.987, 4.346/7.572, 2.805/9.340, 4.576/8.936, 3.082/7.629, 3.750/9.248, 3.229/9.583, 2.835/8.001, 3.980/8.004, 4.102/7.940, 4.139/8.302, 2.847/8.245, 4.297/7.845, 2.931/8.126, 4.273/7.588, 4.204/9.259, 4.082/9.242, 3.923/9.092, 2.998/9.089, 4.056/8.661, 4.085/8.648, 2.879/9.223, 3.655/7.557, 2.597/8.632, 2.636/7.947, 4.496/8.588, 4.212/8.172, 3.457/7.751, 4.369/9.490, 3.645/7.694, 4.314/9.406, 4.304/9.394, 4.640/8.491, 4.034/9.656, 4.153/8.633, 3.377/9.242, 2.937/7.885, 4.288/8.252, 4.449/8.465, 2.686/8.484, 4.213/7.826, 4.560/9.251, 3.624/8.248, 4.163/8.499, 4.361/8.157, 3.484/7.996, 4.650/9.362, 3.137/8.165, 4.580/9.197, 4.313/8.110, 2.903/9.121, 4.346/9.464, 3.213/8.215, 4.328/9.136, 4.656/8.873, 3.005/7.835, 3.805/9.389, 3.246/9.472, 3.386/8.572, 3.271/8.819, 3.224/9.275, 3.973/7.757, 2.807/7.908, 2.989/8.517, 3.209/8.219, 4.253/8.762, 3.010/9.127, 2.737/9.009, 3.323/7.866, 3.869/9.545, 4.282/7.865, 3.221/8.007, 2.564/8.725, 4.511/9.578, 3.489/8.408, 4.285/7.893, 4.270/9.302, 3.522/8.222, 3.447/7.833, 3.768/9.188, 2.696/8.992, 2.984/8.089, 4.481/7.735, 2.734/8.669, 4.383/8.087, 4.682/8.802, 4.434/7.639, 3.073/8.459, 4.197/8.517, 3.819/7.962, 3.929/8.915, 3.298/9.186, 2.925/9.057, 3.790/7.954, 3.946/8.420, 3.289/7.608, 4.224/8.082, 4.456/7.628, 2.553/8.173, 3.683/7.616, 3.346/7.864, 3.073/8.537, 3.561/8.021, 3.885/7.643, 3.022/7.691, 3.892/7.823, 4.170/7.810, 3.717/7.734, 4.441/8.056, 3.948/8.715, 3.858/9.410, 4.530/8.854, 4.004/8.371, 2.830/8.903, 3.310/9.377, 4.507/8.933, 4.221/7.816, 3.655/7.609, 3.083/8.212, 4.678/8.964, 4.678/7.869, 3.384/8.328, 4.044/7.570, 4.455/8.343, 4.436/7.929, 3.791/9.212, 2.743/8.279, 3.701/7.788, 4.092/8.052, 4.698/8.048, 3.803/7.958, 3.211/8.677, 3.387/8.363, 2.651/7.888, 3.995/8.151, 4.483/9.018, 3.316/9.423, 3.519/8.113, 2.865/7.586, 4.472/7.633, 3.275/7.655, 3.254/9.076, 2.895/9.312, 4.078/8.561, 3.971/7.674, 4.415/8.204, 3.524/8.127, 4.025/8.384, 2.597/9.078, 3.455/9.432, 3.692/9.635, 2.957/8.325, 4.636/7.699, 2.733/8.354, 2.615/9.226, 4.333/9.598, 3.466/9.136, 4.540/8.479, 4.670/8.906, 3.433/9.692, 4.524/8.289, 4.004/7.857, 3.960/8.447, 4.422/8.469, 4.225/8.767, 4.082/8.745, 3.845/9.146, 4.689/7.813, 3.325/9.385, 2.822/8.011, 3.955/7.635, 4.370/7.859, 3.616/9.592, 2.981/7.917, 4.155/9.633, 4.538/8.521, 3.711/9.073, 2.810/8.203, 3.140/9.448, 2.659/8.131, 2.920/9.306, 2.862/8.879, 4.318/8.043, 4.685/8.917, 4.167/8.074, 2.729/7.578, 4.643/8.505, 2.593/8.426, 4.456/9.160, 3.730/7.783, 3.695/7.638, 4.305/7.919, 4.457/7.932, 4.329/8.658, 4.572/8.223, 3.592/9.018, 4.440/9.230, 2.611/8.300, 2.571/9.257, 3.199/8.419, 3.903/8.305, 4.244/8.281, 4.395/7.877, 3.575/9.415, 4.272/9.010, 4.332/7.849, 4.027/9.694, 2.583/7.819, 4.693/7.622, 3.821/8.203, 4.297/8.098, 3.308/8.453, 2.814/7.728, 4.473/8.512, 3.795/9.655, 4.443/9.636, 4.306/8.858, 4.195/8.755, 4.161/9.417, 3.653/8.156, 3.643/9.257, 2.764/8.111, 3.317/7.682, 3.712/9.629, 2.817/8.580, 4.158/8.775, 4.371/7.761, 2.569/8.881, 4.432/9.222, 4.001/9.035, 3.189/7.941, 2.938/9.624, 4.134/9.015, 4.104/8.086, 4.573/7.691, 3.607/7.900, 2.750/8.130, 3.060/9.544, 4.303/8.438, 3.376/9.334, 2.594/7.594, 3.757/9.489, 2.577/7.969, 3.056/8.251, 4.476/7.772, 3.354/9.656, 3.097/8.063, 2.572/8.833, 3.519/9.031, 2.659/8.727, 3.345/7.642, 4.083/8.244, 4.304/7.689, 2.802/8.358, 3.724/9.535, 4.261/9.157, 3.442/8.487, 2.590/8.430, 3.412/9.049, 3.776/9.291, 2.588/9.278, 3.859/7.973, 4.388/9.127, 3.127/9.292, 2.723/9.094, 2.762/9.025, 4.225/9.104, 3.161/8.059, 2.577/8.979, 3.944/9.420, 3.120/9.575, 2.821/9.604, 4.307/8.838, 4.218/9.156, 2.555/9.500, 2.989/9.158, 4.340/9.691, 3.554/9.367, 4.209/8.235, 3.767/9.616, 4.441/7.950, 3.315/7.880, 4.444/9.696, 4.677/8.440, 3.372/8.542, 4.212/8.844, 4.366/9.445, 4.555/7.984, 2.967/8.547, 3.284/8.363, 2.702/7.821, 2.746/9.455, 2.677/8.850, 3.703/7.922, 2.799/7.980, 3.546/8.269, 4.375/8.405, 4.007/9.055, 4.347/7.982, 3.023/8.776, 2.859/8.760, 4.360/8.030, 3.038/9.250, 3.571/9.210, 4.491/8.392, 2.956/7.939, 4.630/9.042, 4.124/9.634, 2.929/8.357, 4.684/9.331, 4.069/9.485, 4.117/8.479, 2.938/8.375, 4.464/8.013, 4.506/9.688, 2.606/8.270, 2.963/8.240, 4.518/9.589, 2.826/9.313, 4.210/9.434, 4.591/8.784, 4.568/7.621, 3.890/7.993, 2.947/8.232, 3.746/8.238, 4.244/9.522, 4.098/9.240, 3.190/8.137, 4.161/8.479, 4.321/8.575, 2.580/7.858, 4.311/7.700, 4.447/9.454, 3.994/8.266, 2.705/8.454, 4.018/8.175, 3.505/7.550, 3.623/9.376, 2.944/9.036, 3.868/9.183, 4.148/7.881, 4.535/7.589, 4.698/8.773, 4.585/7.847, 2.651/9.155, 2.970/8.951, 4.229/8.781, 4.698/9.334, 2.751/8.470, 3.958/8.921, 4.078/7.844, 4.685/8.808, 4.395/7.864, 3.498/9.162, 3.501/8.057, 2.716/7.717, 2.716/9.225, 4.255/8.472, 3.116/9.166, 3.630/9.281, 3.481/7.697, 2.818/8.944, 3.364/8.521, 4.465/8.480, 3.015/8.255, 3.355/9.152, 3.955/8.526, 2.938/8.783, 4.583/7.685, 3.952/8.056, 3.441/7.681, 2.563/7.607, 3.278/7.766, 3.511/8.026, 3.283/8.478, 4.234/7.663, 4.564/9.677, 3.411/9.225, 3.987/9.642, 4.658/9.546, 4.196/8.846, 4.170/8.789, 2.668/7.919, 4.097/9.477, 3.732/9.294, 4.665/9.411, 3.705/8.245, 3.338/8.072, 3.088/8.520, 2.609/9.427, 3.006/9.672, 3.477/9.533, 2.550/8.286, 4.019/8.976, 4.100/8.836, 2.681/9.430, 3.428/9.593
}
\def\uniformSmallInside{3.424/8.645, 3.432/8.825, 3.712/8.664, 3.844/8.915, 3.536/8.690, 3.792/8.557, 3.828/8.478, 3.720/8.481, 3.858/8.326, 3.567/8.474, 3.820/8.939, 3.680/8.311, 3.653/8.447, 3.798/8.304, 3.929/8.486, 3.700/8.682, 3.826/8.494, 3.715/8.392, 3.923/8.674, 3.408/8.684, 3.862/8.623, 3.947/8.617, 3.560/8.852, 3.849/8.702, 3.835/8.913, 3.645/8.624, 3.340/8.867, 3.928/8.433, 3.829/8.546, 3.600/8.902, 3.649/8.352, 3.393/8.925, 3.529/8.465, 3.661/8.330, 3.577/8.697, 3.672/8.633, 3.524/8.962, 3.624/8.641
}

%% file: plots/points/points_n=256_k=512_sampling=UNIFORM_shape=THIN.tex
\def\uniformThinOutside{6.928/8.746, 6.951/8.125, 7.088/8.035, 6.871/9.397, 7.017/7.847, 5.051/7.578, 6.177/8.560, 6.305/8.852, 6.316/9.631, 7.015/8.462, 6.852/9.315, 5.239/8.078, 5.299/9.147, 6.909/8.856, 6.072/8.392, 5.216/9.637, 5.221/9.439, 5.567/7.924, 6.202/8.820, 6.908/9.448, 5.859/9.658, 6.423/7.712, 6.141/8.638, 5.671/9.444, 7.041/8.057, 7.003/9.009, 6.225/9.515, 5.156/9.311, 7.062/9.302, 7.185/8.287, 6.313/8.041, 6.944/8.868, 5.950/8.731, 5.514/8.839, 7.168/8.070, 6.399/9.546, 6.931/7.640, 5.158/8.676, 5.936/8.091, 6.453/8.106, 5.616/9.286, 7.119/9.304, 5.231/9.288, 6.769/7.976, 6.521/9.162, 5.066/8.224, 6.172/8.463, 6.105/8.260, 5.492/8.771, 5.846/8.423, 5.120/9.005, 7.055/9.542, 6.843/7.735, 7.090/9.448, 5.581/7.912, 6.383/8.611, 5.426/8.977, 7.021/8.399, 6.166/9.141, 5.676/9.174, 7.100/7.677, 5.891/9.281, 6.747/8.287, 6.965/8.373, 5.717/9.101, 7.193/9.689, 5.124/7.586, 6.351/7.865, 6.654/7.823, 7.072/8.701, 6.754/7.558, 5.235/8.145, 5.436/8.533, 6.422/8.107, 6.696/9.461, 6.317/7.794, 5.335/8.675, 5.883/7.902, 5.699/9.336, 5.130/9.419, 5.436/8.475, 6.030/8.738, 6.427/9.696, 7.038/9.018, 5.953/9.538, 6.492/9.175, 6.014/8.191, 5.423/8.646, 5.560/8.941, 6.264/8.404, 7.101/9.350, 5.843/8.325, 5.103/8.118, 5.119/9.299, 5.418/9.519, 6.108/9.562, 6.914/8.945, 5.140/8.314, 7.142/7.948, 5.872/9.472, 6.994/8.130, 5.474/7.570, 6.736/9.390, 5.981/9.429, 5.230/9.598, 5.740/9.224, 6.188/7.841, 5.507/8.001, 6.177/8.119, 5.432/8.102, 7.129/8.263, 5.477/8.736, 6.093/9.486, 5.422/7.629, 5.681/8.365, 5.943/8.448, 6.746/8.712, 5.109/8.523, 6.869/7.650, 6.399/8.554, 5.441/8.880, 5.859/7.679, 5.933/8.414, 5.182/7.970, 6.923/8.953, 6.262/9.158, 6.353/8.640, 5.662/9.252, 5.751/8.042, 5.341/9.096, 5.406/8.216, 6.767/9.554, 5.554/9.594, 6.762/7.857, 5.240/8.243, 6.247/8.791, 7.033/8.024, 6.698/9.350, 5.257/9.465, 5.114/9.377, 5.997/7.976, 6.954/9.040, 6.746/8.533, 6.759/9.269, 6.255/8.601, 6.437/9.545, 6.364/8.696, 5.779/7.560, 6.839/9.009, 7.151/9.466, 5.673/9.370, 6.276/9.126, 7.106/7.796, 6.604/7.890, 5.606/9.509, 6.360/8.883, 5.944/8.115, 6.743/8.665, 6.534/9.557, 5.203/7.760, 7.020/7.895, 6.263/9.209, 6.081/9.429, 5.092/9.256, 6.147/8.830, 6.541/7.572, 6.243/9.324, 5.265/9.286, 6.992/8.552, 5.359/9.304, 5.405/7.569, 6.958/8.777, 5.115/8.822, 6.920/8.937, 5.459/9.335, 6.255/7.764, 7.003/9.286, 5.229/9.014, 6.162/7.738, 7.143/8.520, 6.923/9.104, 5.648/9.072, 6.551/9.428, 6.856/8.300, 5.398/8.829, 5.354/9.660, 6.011/8.326, 6.094/8.461, 6.289/9.200, 6.332/8.170, 7.096/9.142, 5.074/7.932, 6.408/8.201, 6.377/8.339, 6.217/7.610, 6.138/9.535, 6.745/7.745, 7.068/8.137, 5.219/8.662, 6.940/8.361, 5.544/7.708, 5.330/9.369, 5.808/8.165, 6.581/9.409, 6.977/8.875, 5.511/7.747, 6.988/7.941, 5.124/7.908, 6.292/9.430, 5.492/9.205, 6.860/9.023, 5.849/7.767, 6.148/8.128, 6.415/9.690, 5.777/8.450, 6.060/9.179, 5.285/8.741, 6.780/9.623, 5.997/7.854, 6.079/9.530, 5.810/8.181, 7.020/9.632, 6.955/7.724, 6.811/8.557, 6.479/8.377, 7.162/9.602, 5.720/9.596, 6.873/8.678, 7.122/9.530, 5.092/9.432, 6.518/9.418, 7.000/8.624, 6.088/7.845, 6.270/7.941, 7.024/7.878, 6.950/8.384, 6.593/7.692, 6.599/9.215, 6.458/8.633, 5.503/9.664, 6.148/7.647, 6.448/7.595, 6.521/7.986, 5.598/9.007, 7.023/9.603, 5.260/9.258, 5.812/7.735, 5.184/9.048, 6.784/8.026, 6.398/8.836, 5.733/7.592, 7.046/9.157, 6.378/9.238, 5.453/8.506, 5.504/9.097, 5.904/8.479, 7.053/9.477, 6.154/8.430, 7.163/8.499, 6.894/7.906, 6.756/7.712, 6.368/9.534, 5.345/9.149, 6.566/7.905, 5.880/8.843, 6.738/8.307, 6.756/9.557, 6.772/9.067, 5.785/9.489, 5.852/9.491, 6.959/8.006, 7.193/8.819, 6.742/9.465, 5.432/8.611, 5.763/8.274, 5.100/7.799, 5.476/9.662, 5.247/8.314, 6.333/7.661, 5.148/8.139, 5.065/7.790, 6.148/9.392, 6.436/8.481, 5.887/8.128, 7.021/8.931, 6.233/9.141, 5.461/9.586, 7.073/7.698, 6.661/9.177, 6.341/9.270, 6.628/9.662, 6.273/8.396, 6.302/8.552, 6.067/7.721, 6.448/8.319, 5.702/8.629, 5.444/7.990, 6.796/7.801, 6.331/9.488, 5.854/8.611, 5.379/9.228, 5.315/8.855, 6.072/8.790, 5.377/7.556, 5.224/9.523, 5.265/7.712, 5.114/7.934, 7.023/8.539, 5.607/9.578, 6.052/8.103, 6.167/9.381, 5.990/8.714, 5.975/7.612, 6.611/9.339, 6.489/9.423, 6.220/7.760, 5.690/8.262, 5.396/9.500, 6.048/7.980, 5.583/7.827, 6.808/8.979, 5.455/8.388, 6.197/8.609, 5.923/8.495, 5.989/8.182, 5.058/8.205, 5.766/9.472, 5.646/9.336, 5.258/8.474, 6.394/8.006, 5.767/8.400, 6.993/9.602, 5.134/9.092, 6.823/9.013, 6.198/9.429, 6.015/8.300, 5.336/9.109, 5.967/9.330, 6.444/8.098, 6.104/7.595, 6.722/9.240, 6.505/7.716, 5.435/9.450, 5.817/7.873, 5.112/8.782, 7.192/7.915, 5.332/9.564, 6.553/9.323, 5.755/7.725, 6.078/7.853, 6.911/8.940, 6.535/9.635, 6.777/9.108, 6.995/9.477, 5.522/7.586, 5.552/7.665, 5.335/9.552, 5.739/9.373, 5.858/8.713, 6.556/9.615, 6.779/8.868, 7.162/9.111, 5.310/7.652, 7.161/9.073, 5.946/8.016, 5.869/7.748, 6.692/9.569, 5.700/8.029, 6.901/8.220, 6.918/8.400, 5.818/9.181, 6.064/8.768, 5.344/8.939, 6.960/9.647, 7.079/7.634, 5.888/8.169, 6.771/8.542, 5.610/7.553, 5.065/9.628, 5.624/7.961, 5.223/9.238, 5.969/7.574, 6.981/7.723, 6.992/7.694, 5.323/9.430, 7.035/9.339, 6.971/7.672, 6.690/8.021, 6.782/8.791, 7.046/9.289, 5.507/7.954, 5.537/9.051, 5.810/8.509, 6.482/7.770, 6.278/9.613, 7.013/8.902, 5.993/9.183, 6.088/7.754, 5.593/9.369, 7.049/7.850, 6.037/9.442, 6.774/9.356, 5.437/9.360, 6.406/8.388, 6.497/9.658, 5.706/9.604, 6.148/7.905, 7.193/7.605, 6.042/8.689, 5.672/7.937, 5.610/9.422, 6.486/8.071, 6.852/8.885, 6.778/9.529, 6.075/8.190, 5.280/7.767, 5.875/9.338, 5.370/9.438, 7.049/9.565, 6.188/8.008, 5.207/9.493, 6.399/8.394, 6.839/9.646, 5.720/9.583, 7.155/9.413, 5.660/9.308, 6.716/9.525, 7.025/8.086, 5.073/9.036, 6.409/8.335, 5.975/7.862, 5.395/9.289, 6.576/7.965, 6.378/9.213, 5.124/7.715, 7.012/8.548, 7.007/9.640, 6.964/9.128, 5.388/8.128, 6.412/7.920, 7.049/8.502, 5.897/7.741, 5.103/9.430, 6.952/8.493, 7.188/7.719, 5.371/9.462, 5.241/8.320, 7.076/9.700, 7.161/7.703
}
\def\uniformThinInside{5.554/8.784, 6.478/8.909, 5.504/8.634, 6.565/8.426, 5.755/8.782, 6.563/8.209, 6.460/8.937, 6.493/8.193, 5.616/8.093, 6.726/8.630, 5.992/8.912, 6.510/8.582, 6.735/8.637, 6.595/8.080, 5.683/8.906, 5.634/8.812, 6.641/8.950, 5.784/8.900, 5.734/8.759, 5.926/8.941, 6.021/8.962, 5.783/8.840, 6.646/8.539, 6.164/9.015, 6.643/8.150, 6.322/9.085, 5.601/8.828, 5.827/9.054, 6.600/8.928, 6.597/8.786, 5.900/9.032, 5.676/8.846, 5.562/8.315, 5.481/8.103, 6.667/8.535, 6.487/8.749, 6.721/8.478, 6.377/8.973, 5.893/9.092, 6.037/8.918, 5.674/8.616, 6.524/9.014, 6.104/8.997, 5.624/8.353, 5.629/8.143, 6.138/9.040, 5.572/8.357, 6.708/8.614, 6.066/8.950, 5.579/8.051, 5.552/8.159, 5.539/8.872, 6.408/9.031, 6.705/8.660, 5.848/8.962, 6.672/8.686, 5.781/9.052, 6.586/8.864, 6.587/8.239, 5.824/8.963, 5.985/9.086, 5.627/8.824, 6.506/8.790
}

%% file: appendix.tex
\section*{Appendix A}
\begin{minipage}{\textwidth}%
\makebox[\textwidth]{%
    \renewcommand{\arraystretch}{0.8}
    \begin{tabular}{p{1.8cm}cccccc}
        \toprule
        \textbf{Object} & \textbf{Point Cloud} & \textbf{Reconstruction} & \textbf{Reconst. Error} & \textbf{Segmentation} & \textbf{Seg. Error} & \textbf{Reference} \\ 
        \midrule
        Skull
            & \includegraphics[width=0.09\textwidth, margin=0pt 0pt 0pt 0pt,valign=m]{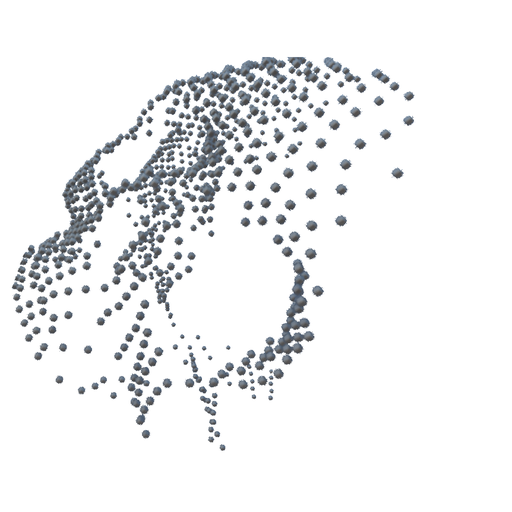}
            & \includegraphics[width=0.09\textwidth, margin=0pt 0pt 0pt 0pt,valign=m]{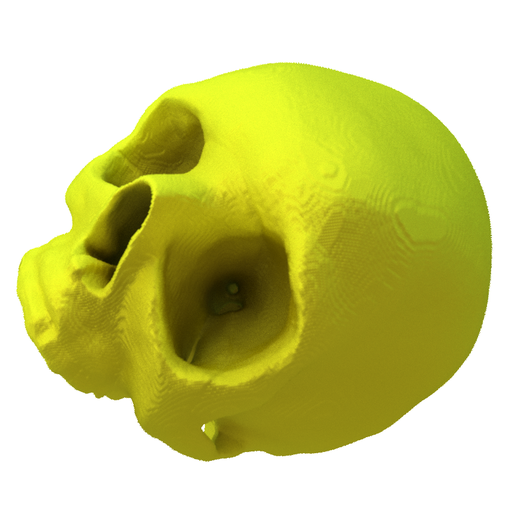}
            & \includegraphics[width=0.09\textwidth, margin=0pt 0pt 0pt 0pt,valign=m]{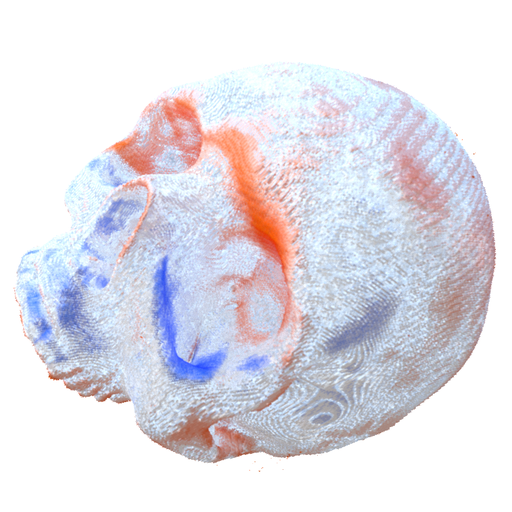}
            & \includegraphics[width=0.09\textwidth, margin=0pt 0pt 0pt 0pt,valign=m]{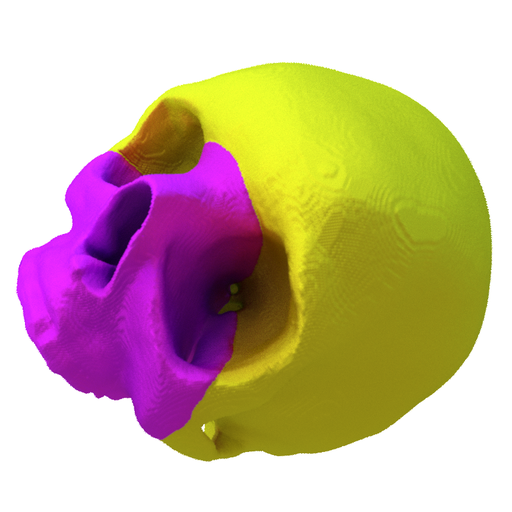}
            & \includegraphics[width=0.09\textwidth, margin=0pt 0pt 0pt 0pt,valign=m]{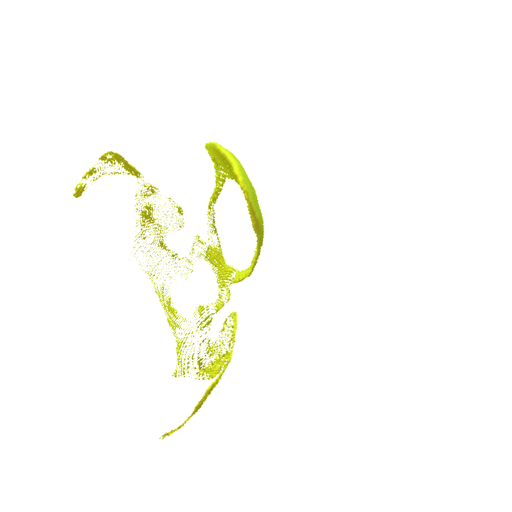}
            & \includegraphics[width=0.09\textwidth, margin=0pt 0pt 0pt 0pt,valign=m]{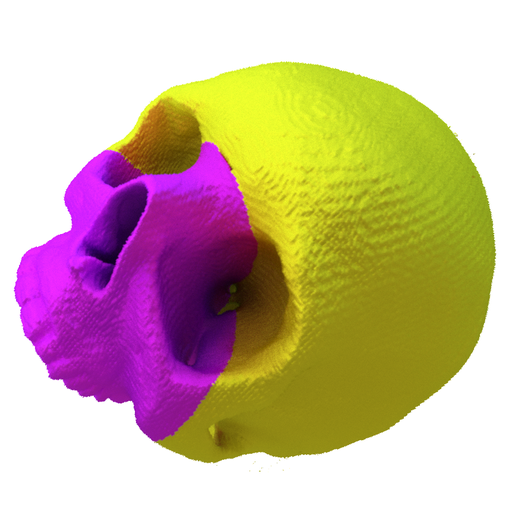} \\
            & & \multicolumn{2}{c}{\small IoU: 0.917} & \multicolumn{2}{c}{\small \small mIoU: 0.910} & \\
        Engine
            & \includegraphics[width=0.09\textwidth, margin=0pt 0pt 0pt 0pt,valign=m]{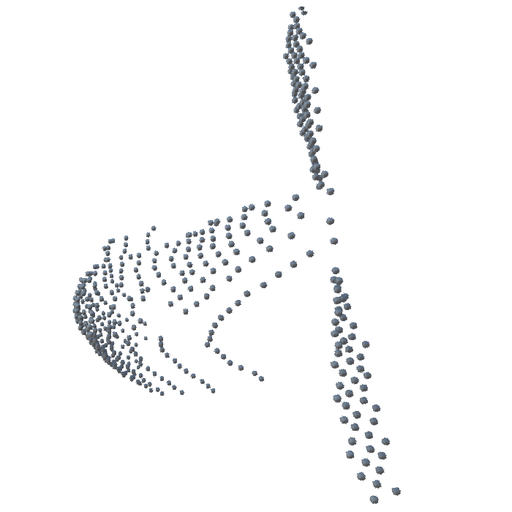}
            & \includegraphics[width=0.09\textwidth, margin=0pt 0pt 0pt 0pt,valign=m]{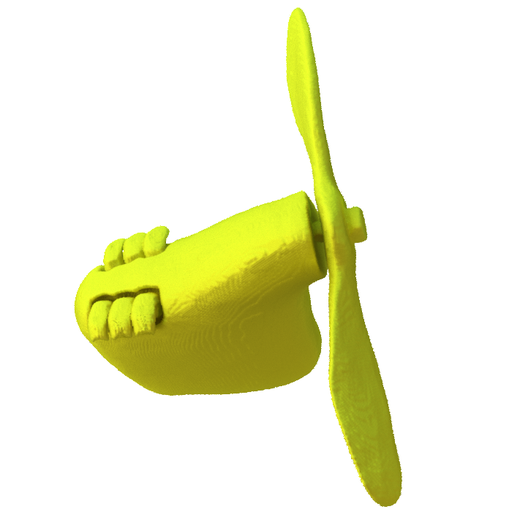}
            & \includegraphics[width=0.09\textwidth, margin=0pt 0pt 0pt 0pt,valign=m]{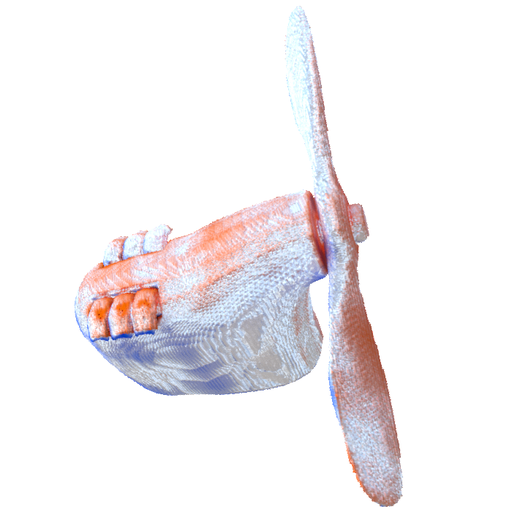}
            & \includegraphics[width=0.09\textwidth, margin=0pt 0pt 0pt 0pt,valign=m]{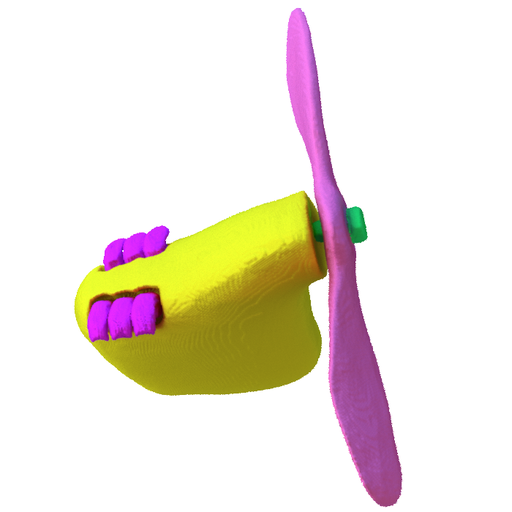}
            & \includegraphics[width=0.09\textwidth, margin=0pt 0pt 0pt 0pt,valign=m]{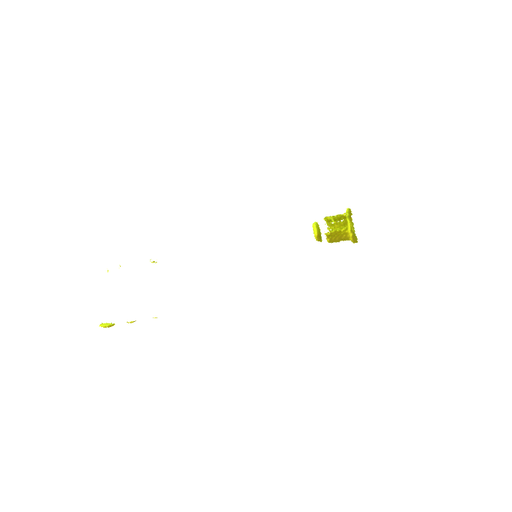}
            & \includegraphics[width=0.09\textwidth, margin=0pt 0pt 0pt 0pt,valign=m]{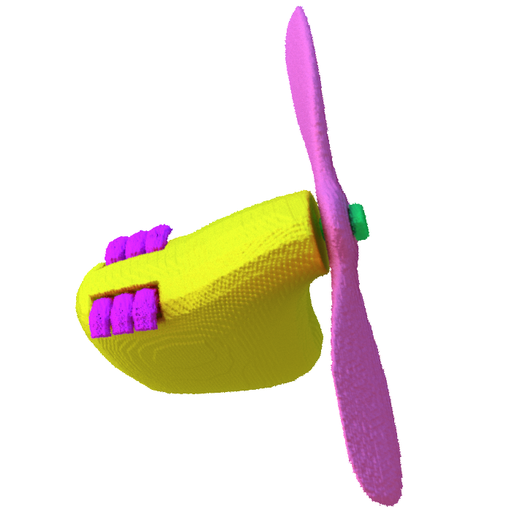} \\
            & & \multicolumn{2}{c}{\small IoU: 0.931} & \multicolumn{2}{c}{\small mIoU: 0.729} & \\
        Liver \& Gallbladder
            & \includegraphics[width=0.09\textwidth, margin=0pt 0pt 0pt 0pt,valign=m]{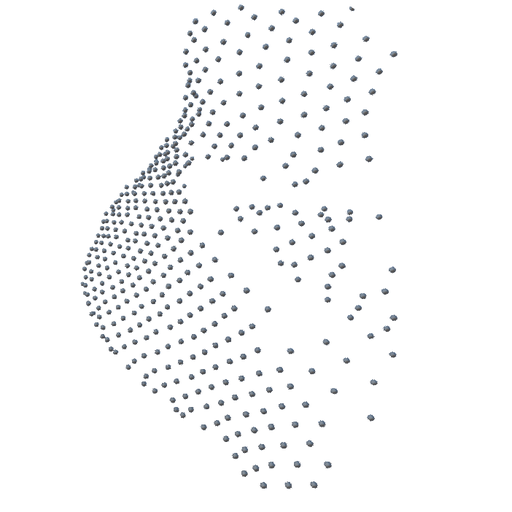}
            & \includegraphics[width=0.09\textwidth, margin=0pt 0pt 0pt 0pt,valign=m]{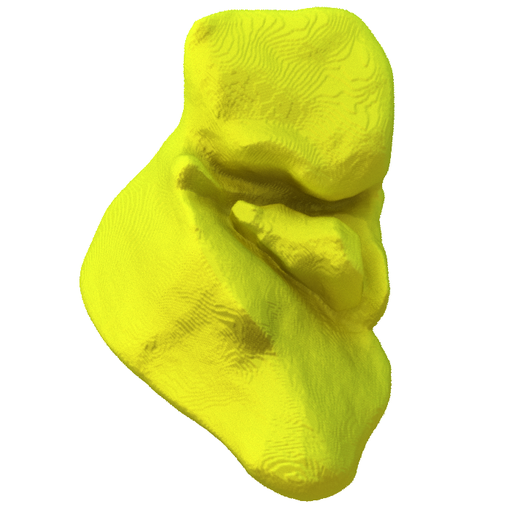}
            & \includegraphics[width=0.09\textwidth, margin=0pt 0pt 0pt 0pt,valign=m]{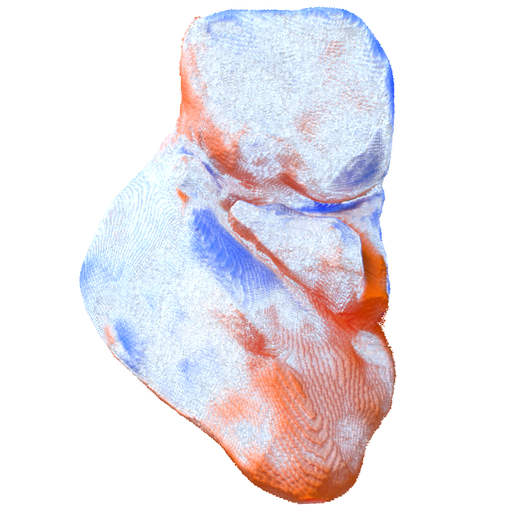}
            & \includegraphics[width=0.09\textwidth, margin=0pt 0pt 0pt 0pt,valign=m]{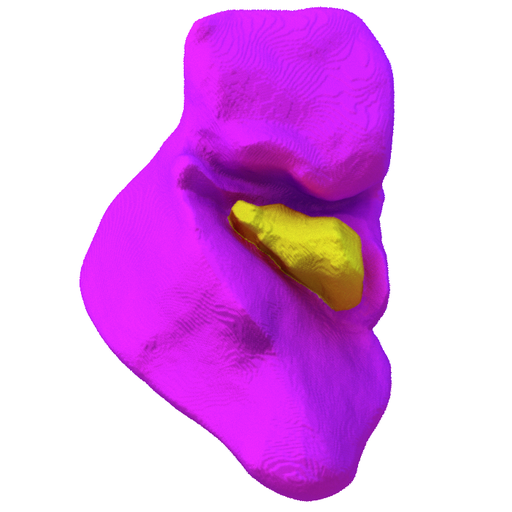}
            & \includegraphics[width=0.09\textwidth, margin=0pt 0pt 0pt 0pt,valign=m]{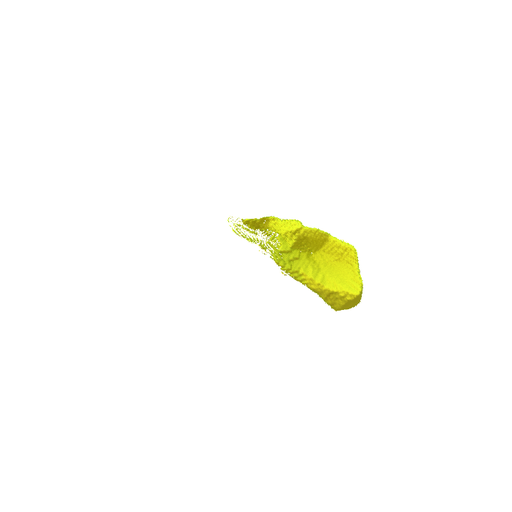}
            & \includegraphics[width=0.09\textwidth, margin=0pt 0pt 0pt 0pt,valign=m]{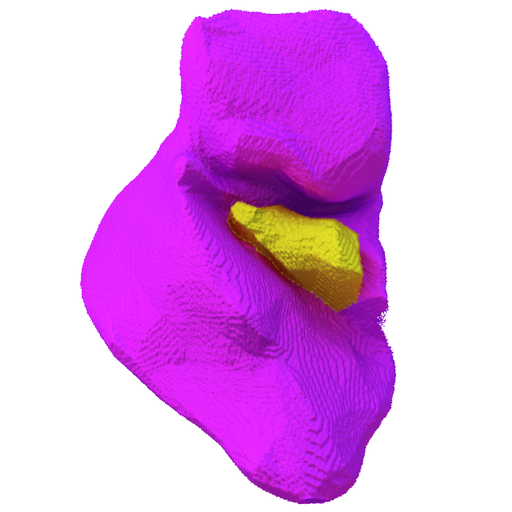} \\
            & & \multicolumn{2}{c}{\small IoU: 0.968} & \multicolumn{2}{c}{\small mIoU: 0.926} & \\
        Shapes
            & \includegraphics[width=0.09\textwidth, margin=0pt 0pt 0pt 0pt,valign=m]{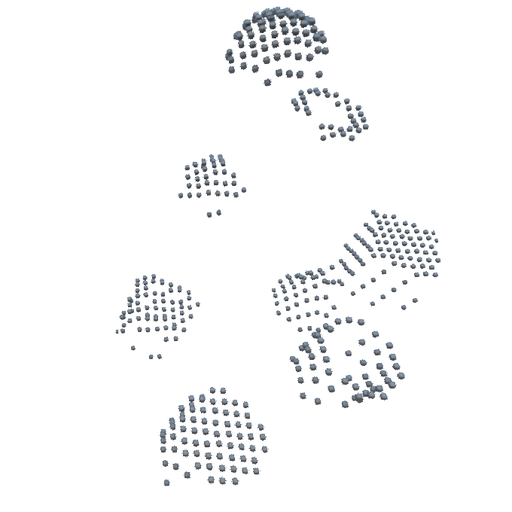}
            & \includegraphics[width=0.09\textwidth, margin=0pt 0pt 0pt 0pt,valign=m]{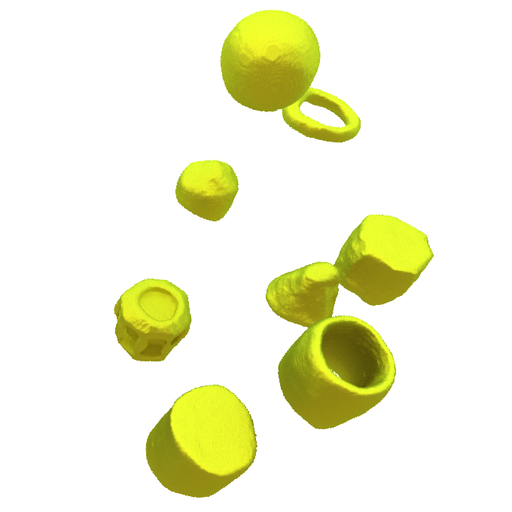}
            & \includegraphics[width=0.09\textwidth, margin=0pt 0pt 0pt 0pt,valign=m]{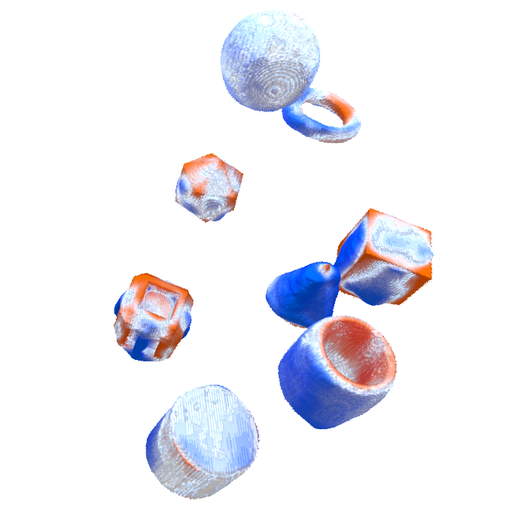}
            & \includegraphics[width=0.09\textwidth, margin=0pt 0pt 0pt 0pt,valign=m]{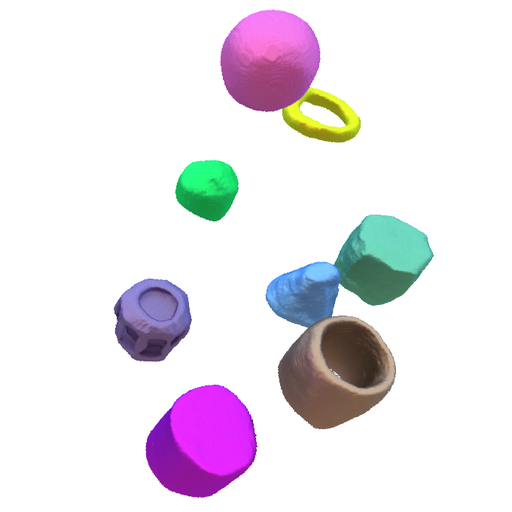}
            & \includegraphics[width=0.09\textwidth, margin=0pt 0pt 0pt 0pt,valign=m]{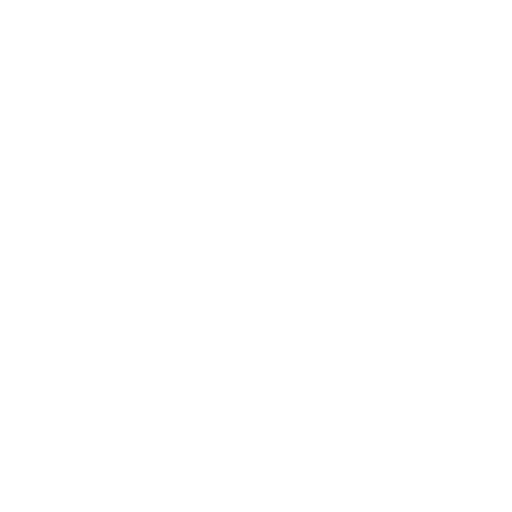}
            & \includegraphics[width=0.09\textwidth, margin=0pt 0pt 0pt 0pt,valign=m]{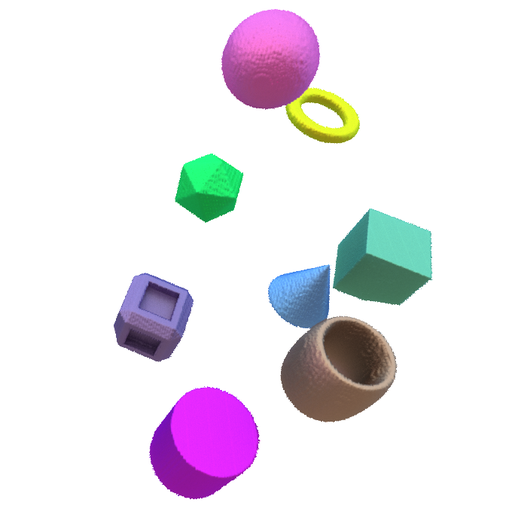} \\
            & & \multicolumn{2}{c}{\small IoU: 0.819} & \multicolumn{2}{c}{\small mIoU: 0.791} & \\
        Sheet
            & \includegraphics[width=0.09\textwidth, margin=0pt 0pt 0pt 0pt,valign=m]{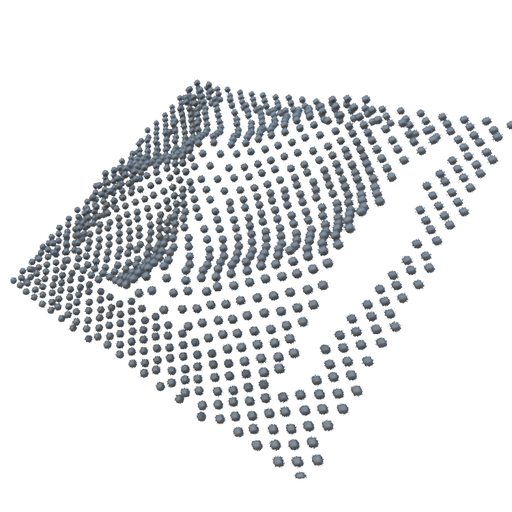}
            & \includegraphics[width=0.09\textwidth, margin=0pt 0pt 0pt 0pt,valign=m]{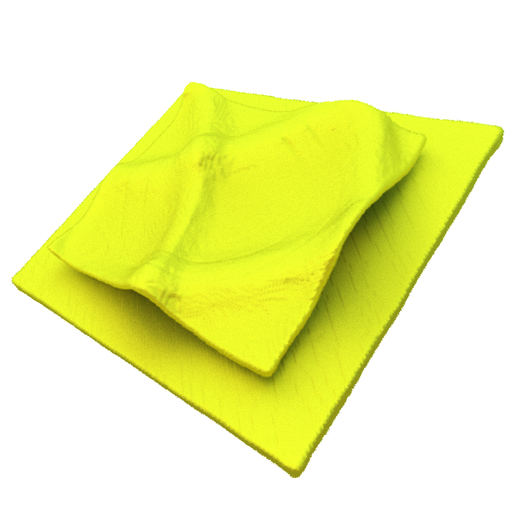}
            & \includegraphics[width=0.09\textwidth, margin=0pt 0pt 0pt 0pt,valign=m]{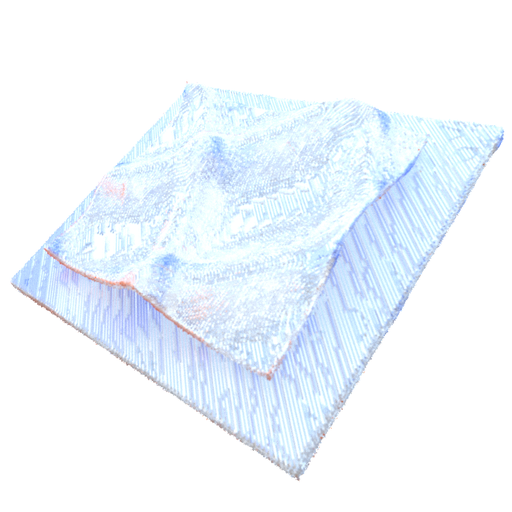}
            & \includegraphics[width=0.09\textwidth, margin=0pt 0pt 0pt 0pt,valign=m]{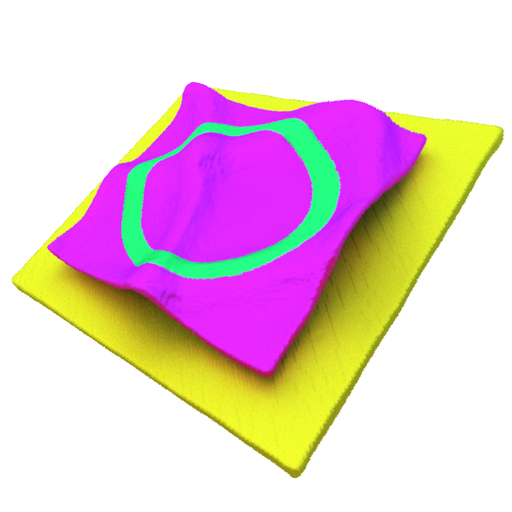}
            & \includegraphics[width=0.09\textwidth, margin=0pt 0pt 0pt 0pt,valign=m]{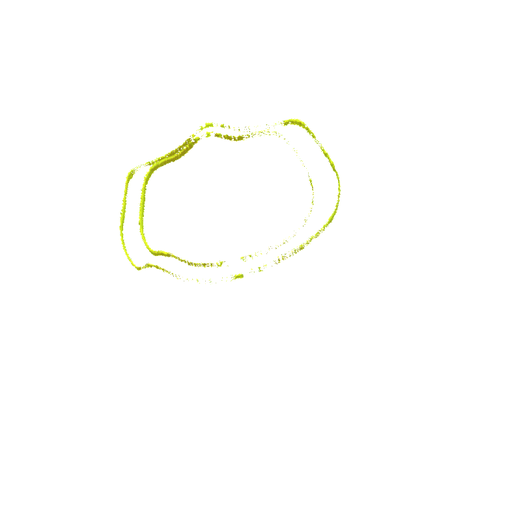}
            & \includegraphics[width=0.09\textwidth, margin=0pt 0pt 0pt 0pt,valign=m]{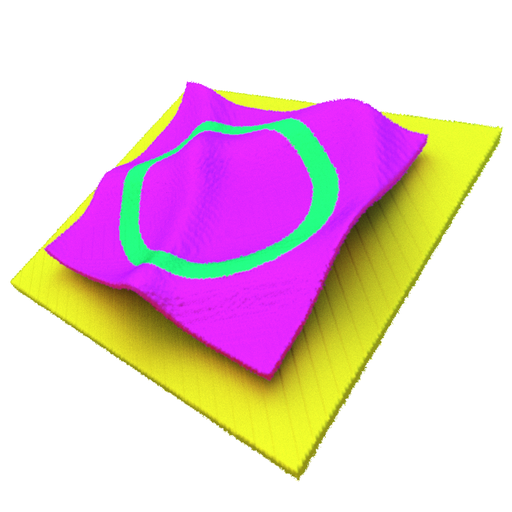} \\
            & & \multicolumn{2}{c}{\small IoU: 0.948} & \multicolumn{2}{c}{\small mIoU: 0.876} & \\
        \bottomrule
    \end{tabular}
}
\vspace{\abovecaptionskip}
\captionof{figure}{
        Continuation of ~\cref{fig:reconstructions}.
        Examples of reconstructions generated by conditioning on the input \textbf{Point Cloud}.
        The \textbf{Reconstruction} takes all non-empty classes to be the same.
        \textbf{Reconstruction Error} identifies \textcolor{blue}{over-reconstruction} and \textcolor{orange}{under-reconstruction} when compared with the reference.
        \textbf{Segmentation} colors each class uniquely, resulting in a \textbf{Segmentation Error} anywhere it differs from the \textbf{Reference}.
        IoU and mIoU values are averaged over all test data, not just the rendered examples.
}
\label{fig:reconstructionsCont}
\vspace{\textfloatsep}
\end{minipage}

\clearpage %

\section*{Appendix B}

\paragraph{Objects}
        \label{sec:scenes}
        \begin{figure}[tbh]
            \centering
            \includegraphics{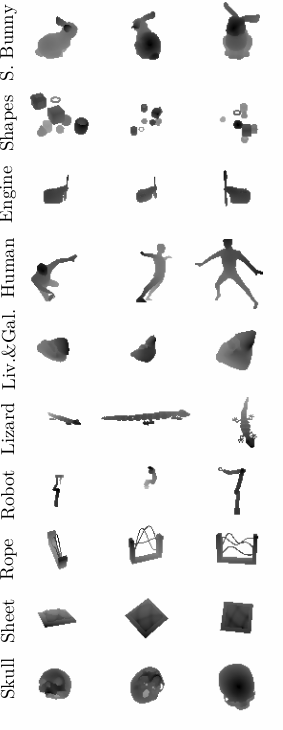}
            \caption{Examples of depth images of deformed objects and random camera perspectives that are used as input to the system.}
            \label{fig:app:depth_examples}
        \end{figure}      
        Although our method is generally applicable to 3D reconstruction tasks, we focus our experiments on objects where detailed prior information exists, but the deformation must be inferred at test time.
        
        Each scene includes a Python script that applies a set of random deformations to the scene within hand-crafted bounds.
        Examples of these random deformations are shown in \Cref{fig:scenes}.

        The Lizard is copyright of Javi Rodríguez under the Creative Commons License and can be found at \href{https://www.thingiverse.com/thing:3505006}{thingiverse.com/thing:3505006}.
        The Stanford Bunny is copyright of \href{http://graphics.stanford.edu/data/3Dscanrep/}{the Stanford Computer Graphics Laboratory}.

        Some objects (\ie the Robot and Lizard) feature articulated deformations, while others feature continuous deformations (\eg the Human, Rope, \etc) or a combination of both.
        Several objects, such as the human, are nearly symmetrical along some axis, yet incorporate a handedness which may be difficult to detect.
        
        Each scene contains a virtual camera that is oriented towards the objects from a random position within a cone of aperture $140\deg$ at a distance from the object between $100$ and $200$ units.
        A $96$x$96$ depth image is captured from the camera's perspective, see \cref{fig:app:depth_examples}.
        Gaussian noise with a standard deviation of $0.1$ units is added to the depth values (the shortest dimension of each scene is $\sim 20$ units).

    \paragraph{Encoder Architectures}
        \label{sec:encoders}
        In this context, an encoder is a neural network that receives a point cloud as input and distils a latent encoding of the information in it.
        We test our framework with $3$ different encoder architectures.
        PointNet++~\cite{qiPointNet2017, qiPointNet2017a} is a well-studied architecture for classifying and segmenting point clouds.
        The PointNet++ encoder contains $2$ PointConv layers and a global max pool layer.
        The PointTransformer~\cite{zhao2021point} contains $5$ transformer blocks.
        DeepSDFs~\cite{parkDeepSDF2019} autodecoder computes a latent encoding as the result of a test-time optimization process.
        The input point cloud is reinterpreted as the locus of points where the signed distance should be zero.
        This allows an optimization to find the latent code that best explains why the all points in the point cloud have a distance of zero.
        While it is possible to define the CE loss during training, at inference time only the L1 loss can be optimized.
        We found this to be enough to infer the segmentation.
        In all cases, the latent encoding has length $1024$.

    \paragraph{Metrics}
        \label{sec:metrics}
        The quality of the reconstruction depends on two factors: 1) the quality of reconstruction itself, and 2) the quality of it's segmentation.
        In line with previous literature~\cite{roldao3D2022}, we evaluate these factors with \gls{iou} and \gls{miou}~\cite{everinghamPascal2015}, respectively.
        \gls{miou} is defined for multi-class segmentation as the average of \gls{iou} of each individual class, ignoring free space.
        Multi-class \gls{iou} is computed by assigning all non-empty classes to an \textit{occupied} class in both the reference and the prediction.
        
        \begin{equation}
            \text{mIoU} = \frac{1}{|C|-1} \sum_{c}^{\{C\} \setminus c_{\text{empty}}} \frac{\text{TP}_c}{\text{TP}_c + \text{FP}_c + \text{FN}_c}
        \end{equation}
        
        Both metrics are calculated over all points in the evaluation dataset.

\paragraph{Evaluation Data}
        The evaluation dataset is created by discretizing the joint bounding box of all objects into a $100^3$ voxel-grid.
        As a performance optimization, points not within the enlarged (by 50\%) bounding boxes of any segments are discarded.

\paragraph{Hyperparameters: PontNet++}
        \label{sec:hyperparameters}
        $28800$ training examples, $128$ test examples.
        Batch size of $40$.
        Learning rate of $0.0005$.
        $300$ epochs.
        $256$ occupancy query points per segment, $128$ inside and $128$ outside.
        $n=k$ for SortSample.
        $n_{uniform}$ is chosen as $15\%$ of total points.
        Variance of Gaussian noise is $0.1$.
        Latent vector has a size of $1024$.
        Positional encoding exponents in $\{-4,\cdots5\}$.
        For more hyperparameters, see Appendix B.
\paragraph{Hyperparameters: Transformer}
Same as hyperparameters of PointNet++ method, see \cref{sec:hyperparameters}.
\paragraph{Hyperparameters: Autodecoder}
Same as hyperparameters of PointNet++ method, see \cref{sec:hyperparameters}. With exception of: Batch size of $64$. Network learning rate of $0.0005$. Latent vector learning rate of $0.001$. $1000$ epochs. 75 latent vector optimization steps at inference time. Dropout in MLP of $0.2$.

\section*{Appendix C}
\paragraph{Noise in Depth Camera}
    We further investigate how the quality of the reconstruction is affected by increasing noise in the depth measurement.
    Data is generated for Robot with $3$ different levels of Gaussian noise applied to the depth values: $0.1$, $0.5$, and $2.0$ units.
    PointNet++ is trained on each dataset using CE+L1 loss, and all $3$ models are cross-evaluated on all other noise levels.
    
    \begin{table}[tbh]
        \caption{IoU and mIoU values from PointNet++ trained using CE+L1 loss on Robot with 3 different levels of Gaussian noise in the depth values. Models trained with one noise level (given by the row) are evaluated on another noise level (given by the column).}
        \label{tab:ablate:noise}
        \centering
        \setlength{\tabcolsep}{4pt} %
        \newcommand{\headercell}[1]{\multicolumn{1}{c}{\multirow{2}{*}{#1}}}
        \begin{tabular}{c  R R R || R R R}
            \toprule
            test \tiny \textrightarrow & \multicolumn{1}{c}{$0.1$} & \multicolumn{1}{c}{$0.5$} & \multicolumn{1}{c}{$2.0$} & \multicolumn{1}{c}{$0.1$} & \multicolumn{1}{c}{$0.5$} & \multicolumn{1}{c}{$2.0$} \\
            \midrule
            $0.1$ & 0.882 & 0.581 & 0.346 & 0.864 & 0.541 & 0.289 \\
            $0.5$ & 0.852 & 0.852 & 0.763 & 0.829 & 0.830 & 0.734 \\
            $2.0$ & 0.808 & 0.816 & 0.815 & 0.781 & 0.789 & 0.789 \\
            \midrule
            train \tiny \textuparrow & \multicolumn{3}{c}{IoU} & \multicolumn{3}{c}{mIoU} \\
            \bottomrule
        \end{tabular}      
    \end{table}

    \gls{iou} and \gls{miou} values are shown in \cref{tab:ablate:noise}.
    For comparison with a maximum noise level of 2 units, the shortest dimension for Robot is 5 units.
    Despite extremely noisy point clouds, the network is able to reconstruct Robot accurately, so long as an equivalent or higher level of noise is seen during training.
    This may be beneficial to real world applications, where an accurate depth measurement is not always economical or feasible.
    Only if the noise significantly exceeds the noise distribution in the training data is reconstruction likely to fail.

\clearpage %

\section*{Appendix D}
\begin{minipage}{\textwidth}%
\vspace{\abovecaptionskip}

\label{fig:reconstructions_real_world_all}
\makebox[\textwidth]{%
    \renewcommand{\arraystretch}{0.8}
    \begin{tabular}{cccccc}
        \toprule
        \textbf{Point Cloud} & \textbf{Reconstruction} & \textbf{Reconst. Error} & \textbf{Segmentation} & \textbf{Seg. Error} & \textbf{Reference} \\ 
        \midrule
              \rotatebox{-90}{\includegraphics[width=0.06\textwidth, margin=0pt 0pt 0pt 0pt,valign=m]{images/real_world_data_trimmed/0/point_cloud_sphere.png}}
            & \rotatebox{-90}{\includegraphics[width=0.06\textwidth, margin=0pt 0pt 0pt 0pt,valign=m]{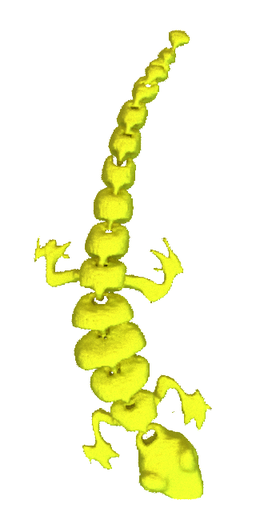}}
            & \rotatebox{-90}{\includegraphics[width=0.06\textwidth, margin=0pt 0pt 0pt 0pt,valign=m]{images/real_world_data_trimmed/0/reconstruction_error.png}}
            & \rotatebox{-90}{\includegraphics[width=0.06\textwidth, margin=0pt 0pt 0pt 0pt,valign=m]{images/real_world_data_trimmed/0/segmentation.png}}
            & \rotatebox{-90}{\includegraphics[width=0.06\textwidth, margin=0pt 0pt 0pt 0pt,valign=m]{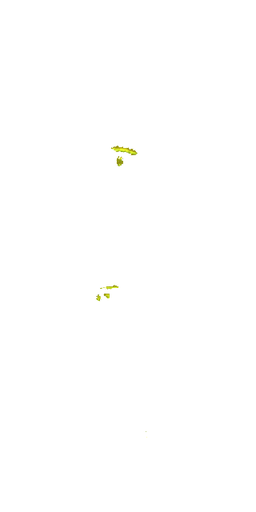}}
            & \rotatebox{-90}{\includegraphics[width=0.06\textwidth, margin=0pt 0pt 0pt 0pt,valign=m]{images/real_world_data_trimmed/0/reference.png}}\\

              \rotatebox{-90}{\includegraphics[width=0.06\textwidth, margin=0pt 0pt 0pt 0pt,valign=m]{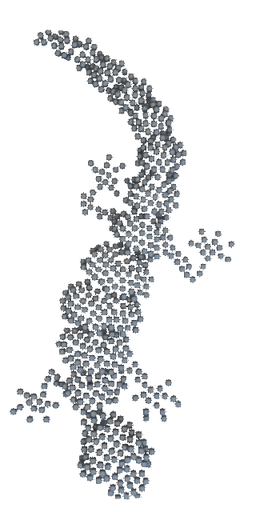}}
            & \rotatebox{-90}{\includegraphics[width=0.06\textwidth, margin=0pt 0pt 0pt 0pt,valign=m]{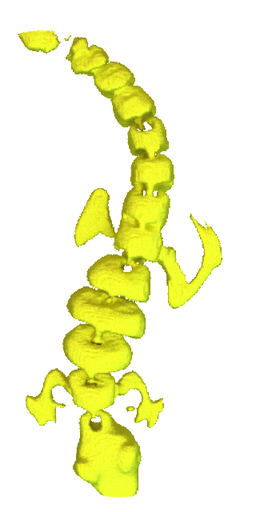}}
            & \rotatebox{-90}{\includegraphics[width=0.06\textwidth, margin=0pt 0pt 0pt 0pt,valign=m]{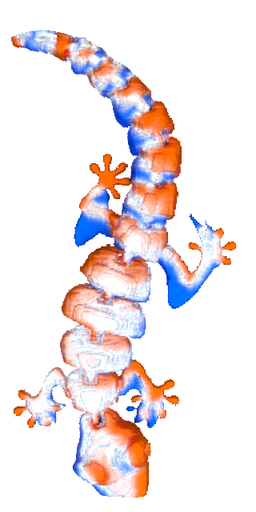}}
            & \rotatebox{-90}{\includegraphics[width=0.06\textwidth, margin=0pt 0pt 0pt 0pt,valign=m]{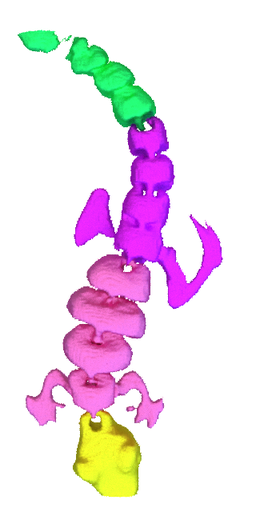}}
            & \rotatebox{-90}{\includegraphics[width=0.06\textwidth, margin=0pt 0pt 0pt 0pt,valign=m]{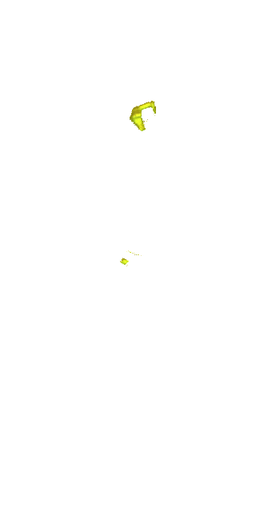}}
            & \rotatebox{-90}{\includegraphics[width=0.06\textwidth, margin=0pt 0pt 0pt 0pt,valign=m]{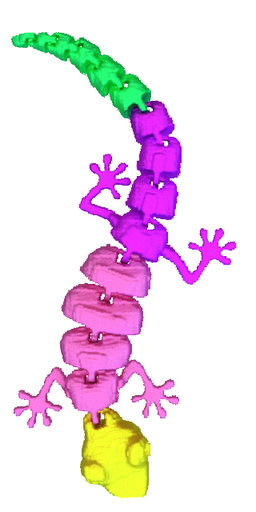}}\\

              \rotatebox{-90}{\includegraphics[width=0.06\textwidth, margin=0pt 0pt 0pt 0pt,valign=m]{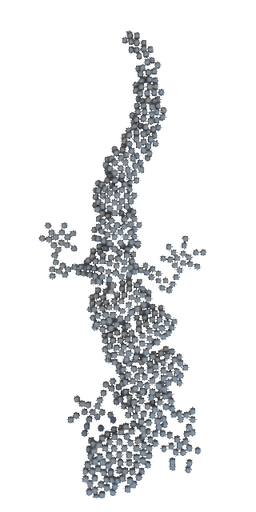}}
            & \rotatebox{-90}{\includegraphics[width=0.06\textwidth, margin=0pt 0pt 0pt 0pt,valign=m]{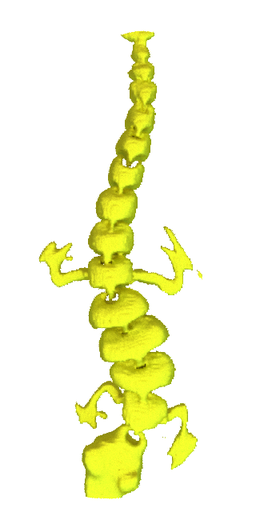}}
            & \rotatebox{-90}{\includegraphics[width=0.06\textwidth, margin=0pt 0pt 0pt 0pt,valign=m]{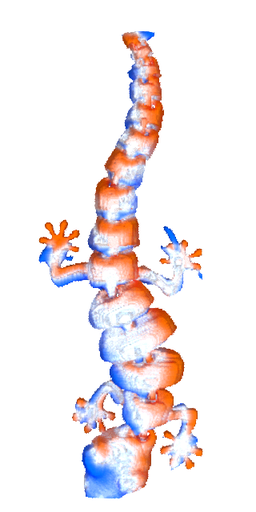}}
            & \rotatebox{-90}{\includegraphics[width=0.06\textwidth, margin=0pt 0pt 0pt 0pt,valign=m]{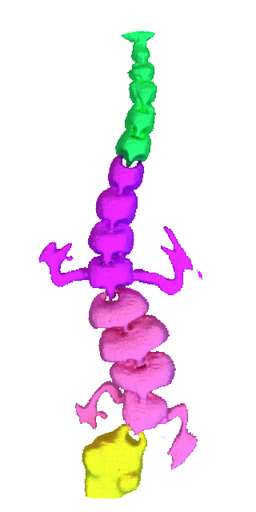}}
            & \rotatebox{-90}{\includegraphics[width=0.06\textwidth, margin=0pt 0pt 0pt 0pt,valign=m]{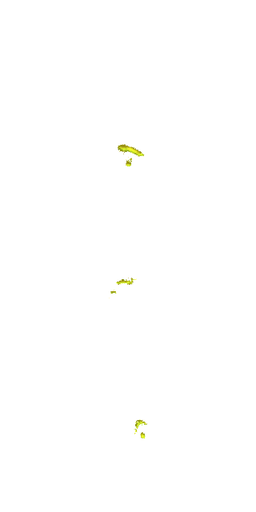}}
            & \rotatebox{-90}{\includegraphics[width=0.06\textwidth, margin=0pt 0pt 0pt 0pt,valign=m]{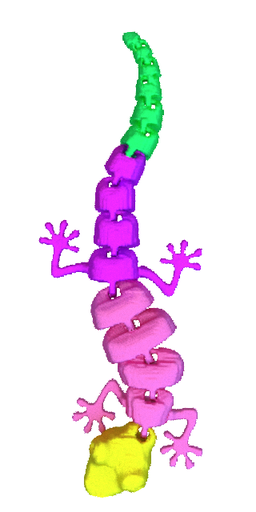}}\\

              \rotatebox{-90}{\includegraphics[width=0.06\textwidth, margin=0pt 0pt 0pt 0pt,valign=m]{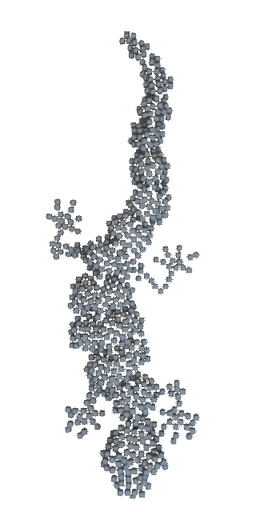}}
            & \rotatebox{-90}{\includegraphics[width=0.06\textwidth, margin=0pt 0pt 0pt 0pt,valign=m]{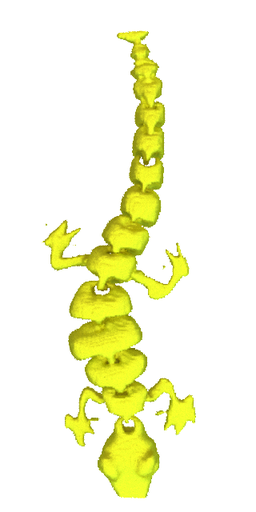}}
            & \rotatebox{-90}{\includegraphics[width=0.06\textwidth, margin=0pt 0pt 0pt 0pt,valign=m]{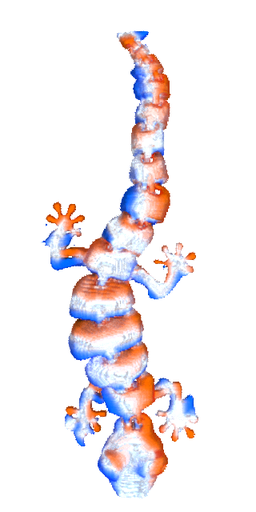}}
            & \rotatebox{-90}{\includegraphics[width=0.06\textwidth, margin=0pt 0pt 0pt 0pt,valign=m]{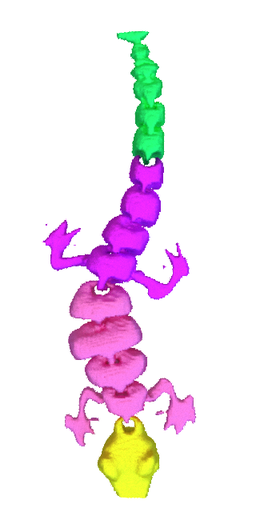}}
            & \rotatebox{-90}{\includegraphics[width=0.06\textwidth, margin=0pt 0pt 0pt 0pt,valign=m]{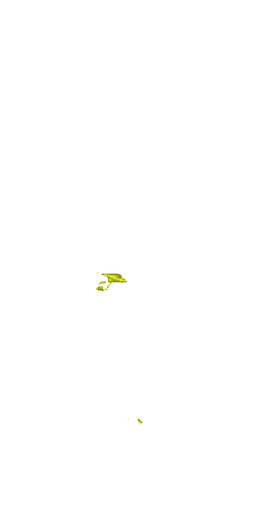}}
            & \rotatebox{-90}{\includegraphics[width=0.06\textwidth, margin=0pt 0pt 0pt 0pt,valign=m]{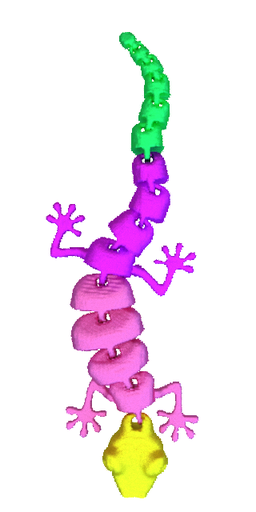}}\\

              \rotatebox{90}{\includegraphics[width=0.06\textwidth, margin=0pt 0pt 0pt 0pt,valign=m]{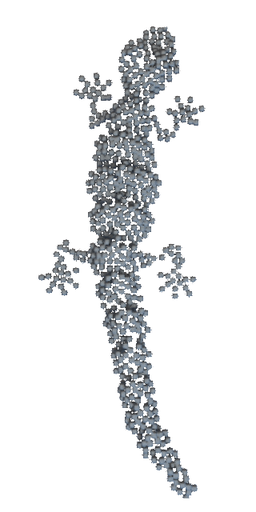}}
            & \rotatebox{90}{\includegraphics[width=0.06\textwidth, margin=0pt 0pt 0pt 0pt,valign=m]{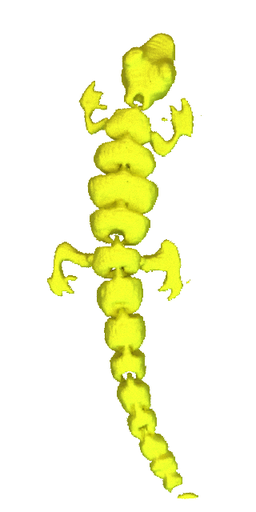}}
            & \rotatebox{90}{\includegraphics[width=0.06\textwidth, margin=0pt 0pt 0pt 0pt,valign=m]{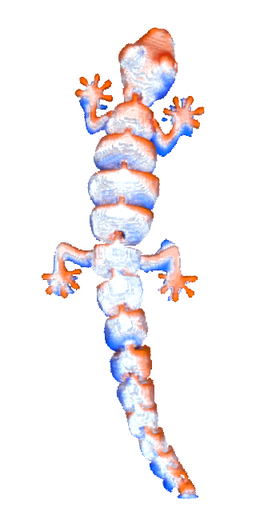}}
            & \rotatebox{90}{\includegraphics[width=0.06\textwidth, margin=0pt 0pt 0pt 0pt,valign=m]{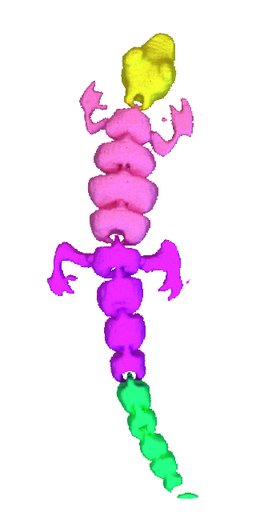}}
            & \rotatebox{90}{\includegraphics[width=0.06\textwidth, margin=0pt 0pt 0pt 0pt,valign=m]{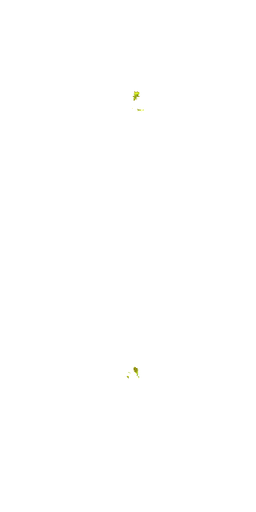}}
            & \rotatebox{90}{\includegraphics[width=0.06\textwidth, margin=0pt 0pt 0pt 0pt,valign=m]{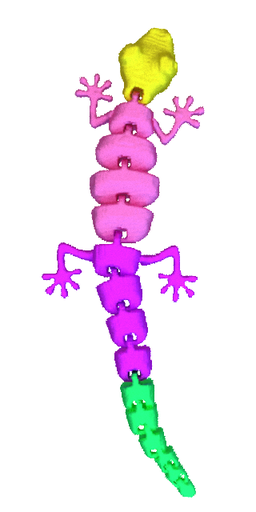}}\\

              \rotatebox{90}{\includegraphics[width=0.06\textwidth, margin=0pt 0pt 0pt 0pt,valign=m]{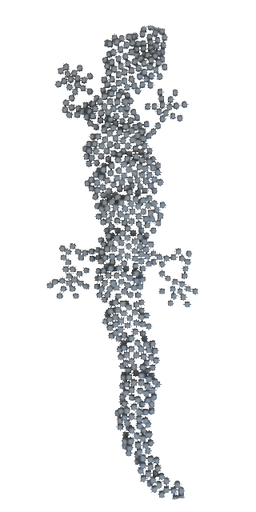}}
            & \rotatebox{90}{\includegraphics[width=0.06\textwidth, margin=0pt 0pt 0pt 0pt,valign=m]{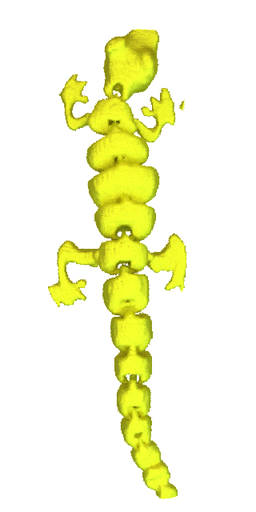}}
            & \rotatebox{90}{\includegraphics[width=0.06\textwidth, margin=0pt 0pt 0pt 0pt,valign=m]{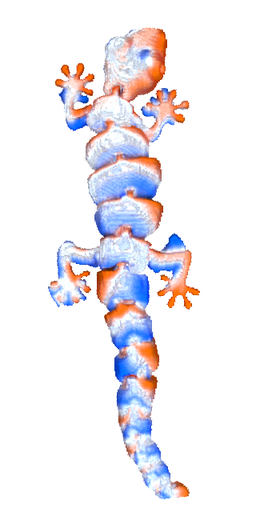}}
            & \rotatebox{90}{\includegraphics[width=0.06\textwidth, margin=0pt 0pt 0pt 0pt,valign=m]{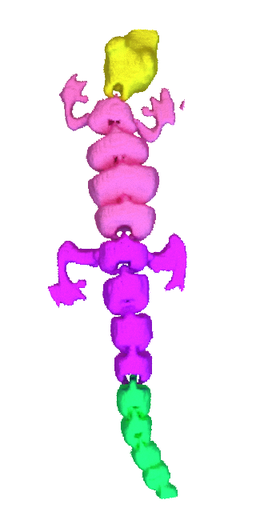}}
            & \rotatebox{90}{\includegraphics[width=0.06\textwidth, margin=0pt 0pt 0pt 0pt,valign=m]{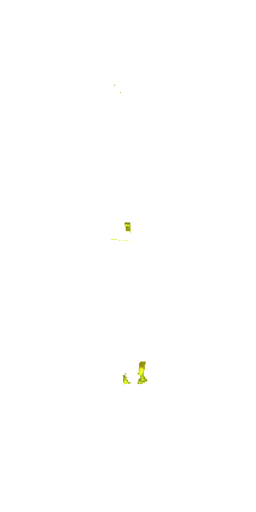}}
            & \rotatebox{90}{\includegraphics[width=0.06\textwidth, margin=0pt 0pt 0pt 0pt,valign=m]{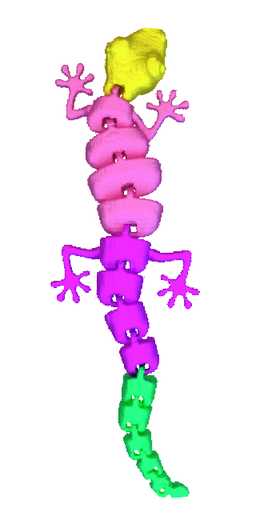}}\\

              \rotatebox{90}{\includegraphics[width=0.06\textwidth, margin=0pt 0pt 0pt 0pt,valign=m]{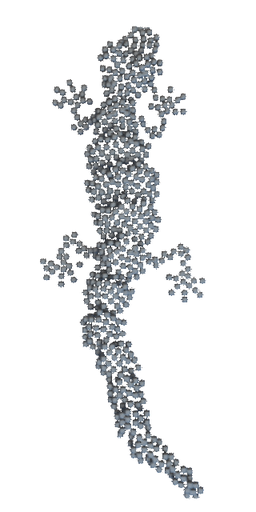}}
            & \rotatebox{90}{\includegraphics[width=0.06\textwidth, margin=0pt 0pt 0pt 0pt,valign=m]{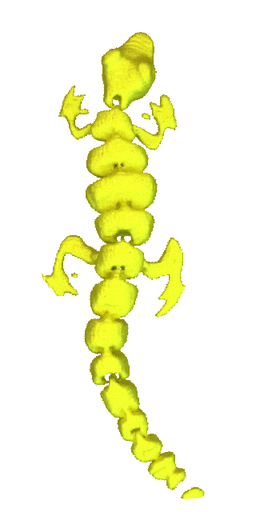}}
            & \rotatebox{90}{\includegraphics[width=0.06\textwidth, margin=0pt 0pt 0pt 0pt,valign=m]{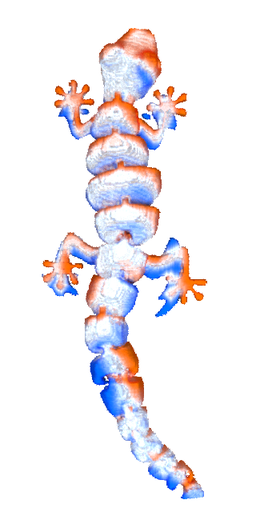}}
            & \rotatebox{90}{\includegraphics[width=0.06\textwidth, margin=0pt 0pt 0pt 0pt,valign=m]{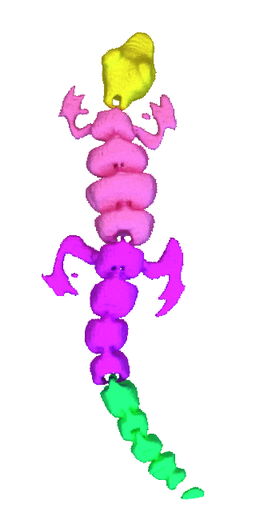}}
            & \rotatebox{90}{\includegraphics[width=0.06\textwidth, margin=0pt 0pt 0pt 0pt,valign=m]{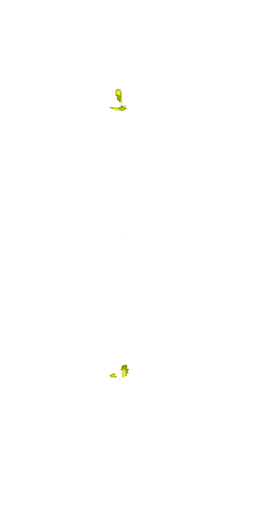}}
            & \rotatebox{90}{\includegraphics[width=0.06\textwidth, margin=0pt 0pt 0pt 0pt,valign=m]{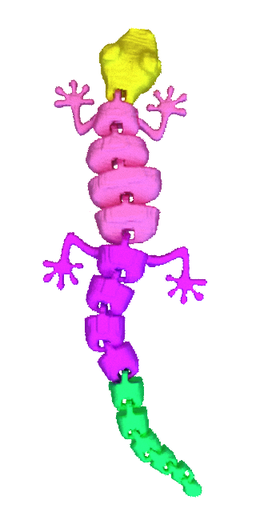}}\\

              \rotatebox{90}{\includegraphics[width=0.06\textwidth, margin=0pt 0pt 0pt 0pt,valign=m]{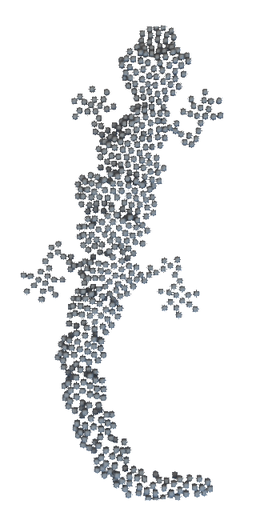}}
            & \rotatebox{90}{\includegraphics[width=0.06\textwidth, margin=0pt 0pt 0pt 0pt,valign=m]{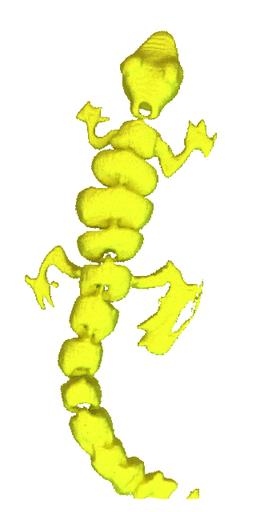}}
            & \rotatebox{90}{\includegraphics[width=0.06\textwidth, margin=0pt 0pt 0pt 0pt,valign=m]{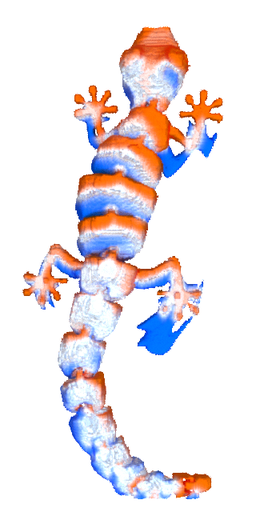}}
            & \rotatebox{90}{\includegraphics[width=0.06\textwidth, margin=0pt 0pt 0pt 0pt,valign=m]{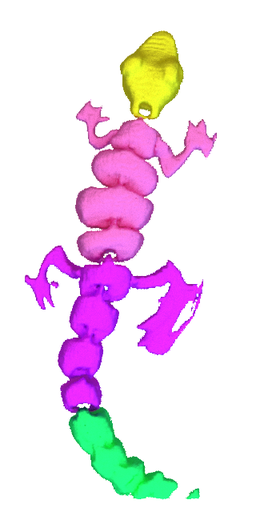}}
            & \rotatebox{90}{\includegraphics[width=0.06\textwidth, margin=0pt 0pt 0pt 0pt,valign=m]{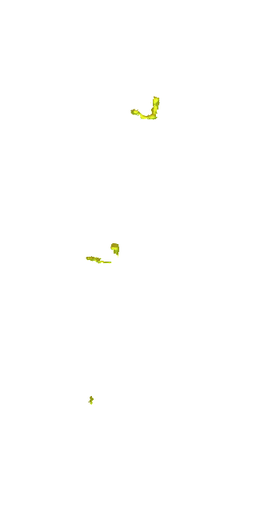}}
            & \rotatebox{90}{\includegraphics[width=0.06\textwidth, margin=0pt 0pt 0pt 0pt,valign=m]{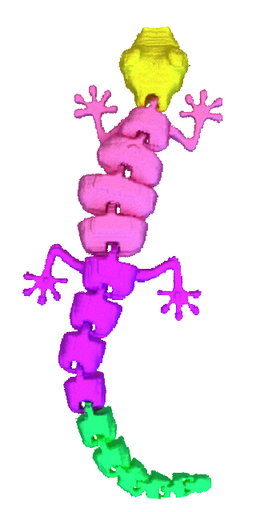}}\\

              \rotatebox{90}{\includegraphics[width=0.06\textwidth, margin=0pt 0pt 0pt 0pt,valign=m]{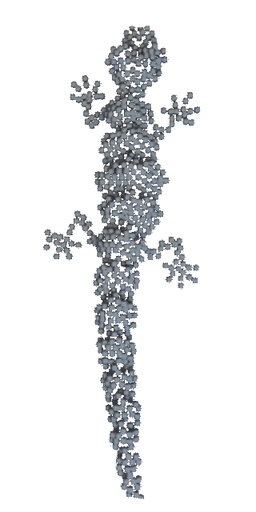}}
            & \rotatebox{90}{\includegraphics[width=0.06\textwidth, margin=0pt 0pt 0pt 0pt,valign=m]{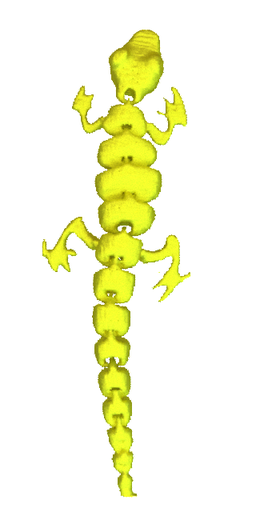}}
            & \rotatebox{90}{\includegraphics[width=0.06\textwidth, margin=0pt 0pt 0pt 0pt,valign=m]{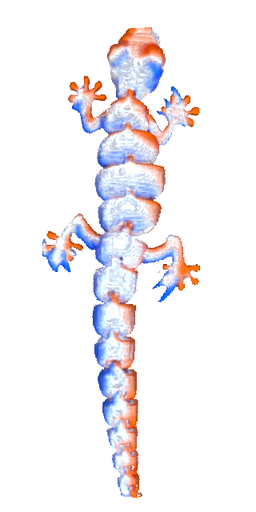}}
            & \rotatebox{90}{\includegraphics[width=0.06\textwidth, margin=0pt 0pt 0pt 0pt,valign=m]{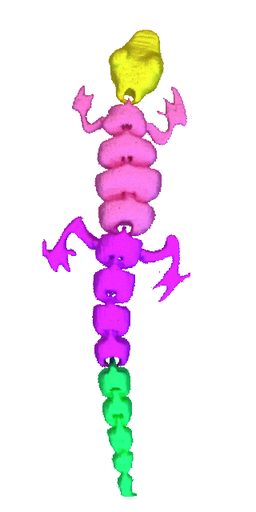}}
            & \rotatebox{90}{\includegraphics[width=0.06\textwidth, margin=0pt 0pt 0pt 0pt,valign=m]{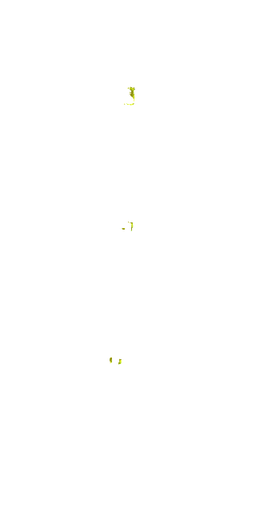}}
            & \rotatebox{90}{\includegraphics[width=0.06\textwidth, margin=0pt 0pt 0pt 0pt,valign=m]{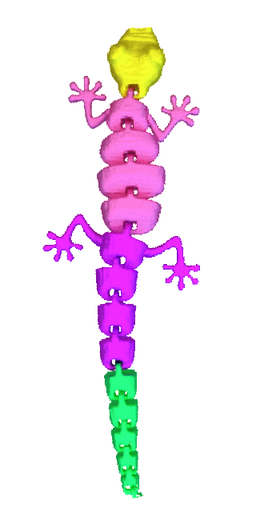}}\\

              \rotatebox{90}{\includegraphics[width=0.06\textwidth, margin=0pt 0pt 0pt 0pt,valign=m]{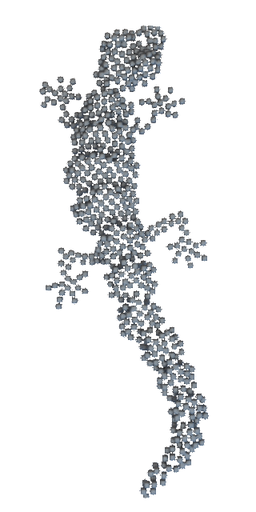}}
            & \rotatebox{90}{\includegraphics[width=0.06\textwidth, margin=0pt 0pt 0pt 0pt,valign=m]{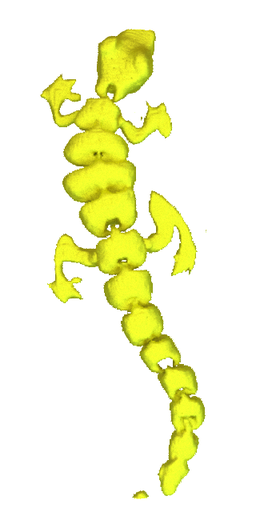}}
            & \rotatebox{90}{\includegraphics[width=0.06\textwidth, margin=0pt 0pt 0pt 0pt,valign=m]{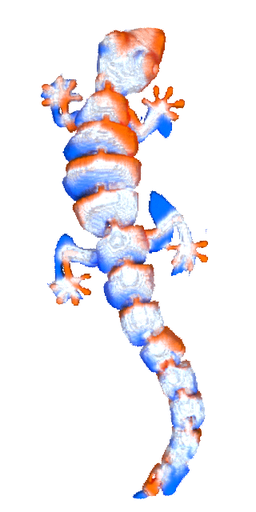}}
            & \rotatebox{90}{\includegraphics[width=0.06\textwidth, margin=0pt 0pt 0pt 0pt,valign=m]{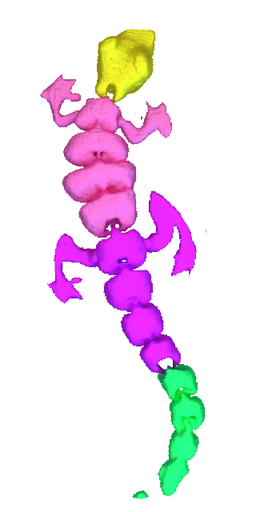}}
            & \rotatebox{90}{\includegraphics[width=0.06\textwidth, margin=0pt 0pt 0pt 0pt,valign=m]{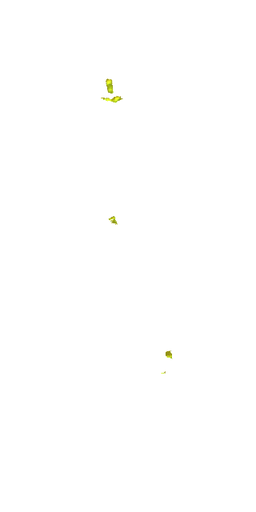}}
            & \rotatebox{90}{\includegraphics[width=0.06\textwidth, margin=0pt 0pt 0pt 0pt,valign=m]{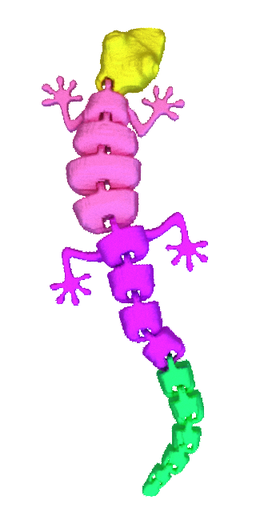}}\\

              \rotatebox{90}{\includegraphics[width=0.06\textwidth, margin=0pt 0pt 0pt 0pt,valign=m]{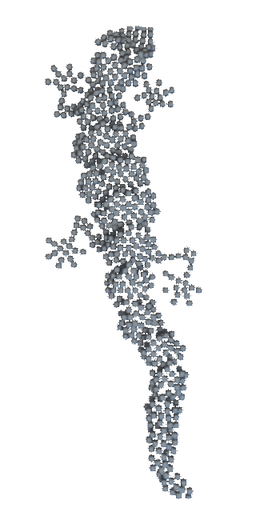}}
            & \rotatebox{90}{\includegraphics[width=0.06\textwidth, margin=0pt 0pt 0pt 0pt,valign=m]{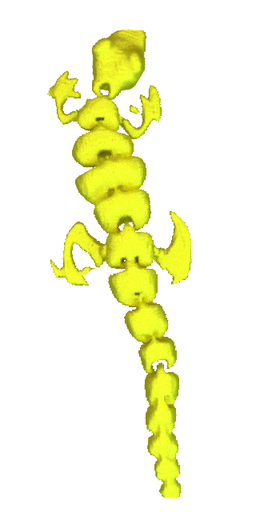}}
            & \rotatebox{90}{\includegraphics[width=0.06\textwidth, margin=0pt 0pt 0pt 0pt,valign=m]{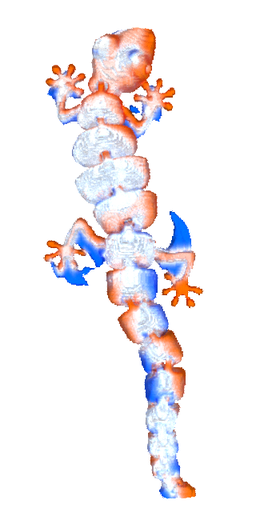}}
            & \rotatebox{90}{\includegraphics[width=0.06\textwidth, margin=0pt 0pt 0pt 0pt,valign=m]{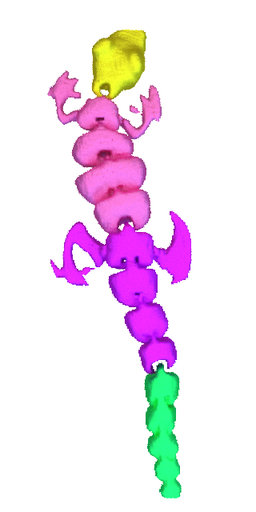}}
            & \rotatebox{90}{\includegraphics[width=0.06\textwidth, margin=0pt 0pt 0pt 0pt,valign=m]{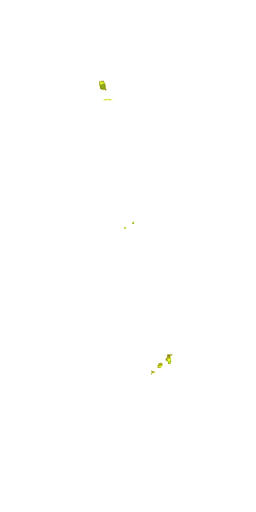}}
            & \rotatebox{90}{\includegraphics[width=0.06\textwidth, margin=0pt 0pt 0pt 0pt,valign=m]{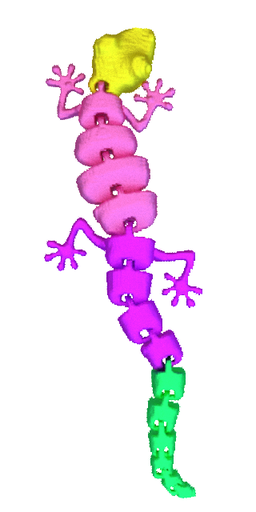}}\\

              \rotatebox{90}{\includegraphics[width=0.06\textwidth, margin=0pt 0pt 0pt 0pt,valign=m]{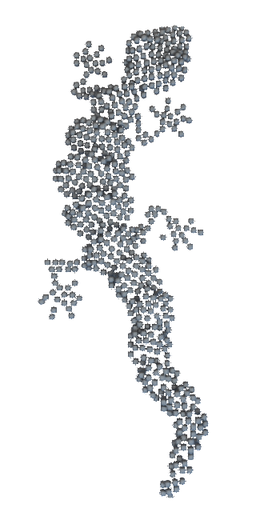}}
            & \rotatebox{90}{\includegraphics[width=0.06\textwidth, margin=0pt 0pt 0pt 0pt,valign=m]{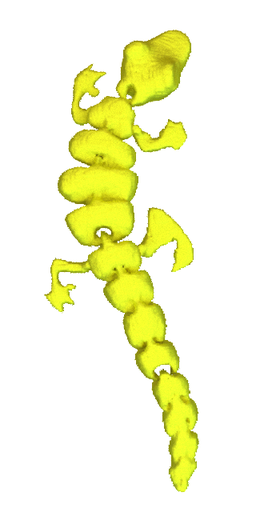}}
            & \rotatebox{90}{\includegraphics[width=0.06\textwidth, margin=0pt 0pt 0pt 0pt,valign=m]{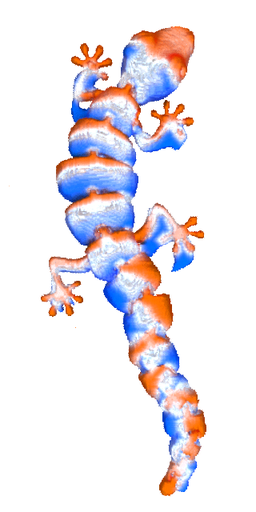}}
            & \rotatebox{90}{\includegraphics[width=0.06\textwidth, margin=0pt 0pt 0pt 0pt,valign=m]{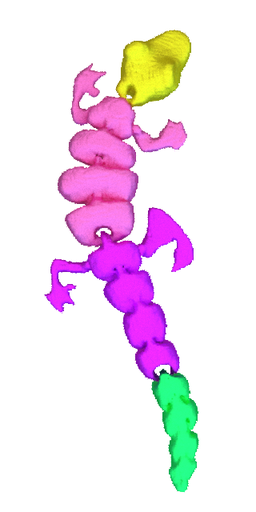}}
            & \rotatebox{90}{\includegraphics[width=0.06\textwidth, margin=0pt 0pt 0pt 0pt,valign=m]{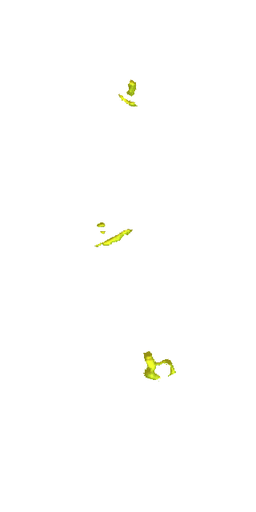}}
            & \rotatebox{90}{\includegraphics[width=0.06\textwidth, margin=0pt 0pt 0pt 0pt,valign=m]{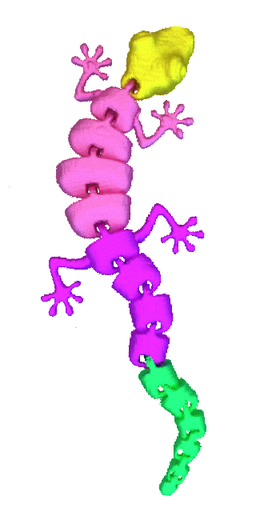}}\\

              \rotatebox{90}{\includegraphics[width=0.06\textwidth, margin=0pt 0pt 0pt 0pt,valign=m]{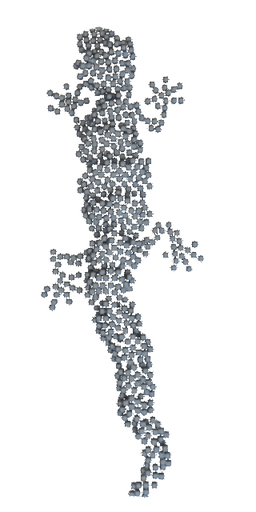}}
            & \rotatebox{90}{\includegraphics[width=0.06\textwidth, margin=0pt 0pt 0pt 0pt,valign=m]{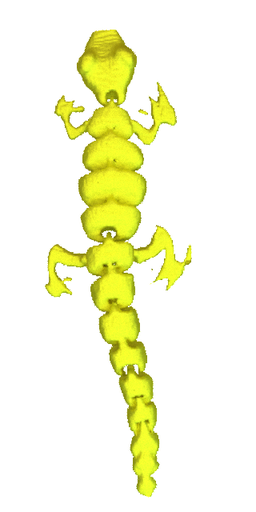}}
            & \rotatebox{90}{\includegraphics[width=0.06\textwidth, margin=0pt 0pt 0pt 0pt,valign=m]{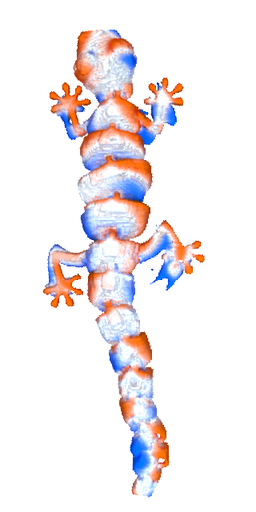}}
            & \rotatebox{90}{\includegraphics[width=0.06\textwidth, margin=0pt 0pt 0pt 0pt,valign=m]{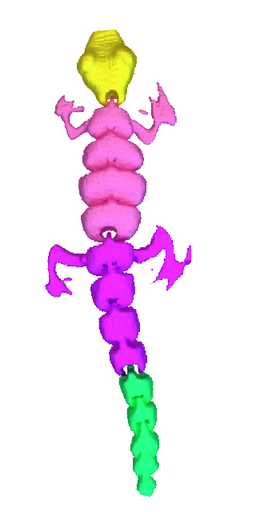}}
            & \rotatebox{90}{\includegraphics[width=0.06\textwidth, margin=0pt 0pt 0pt 0pt,valign=m]{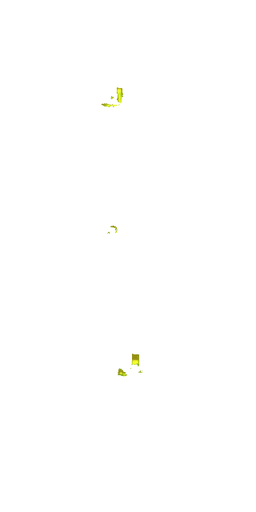}}
            & \rotatebox{90}{\includegraphics[width=0.06\textwidth, margin=0pt 0pt 0pt 0pt,valign=m]{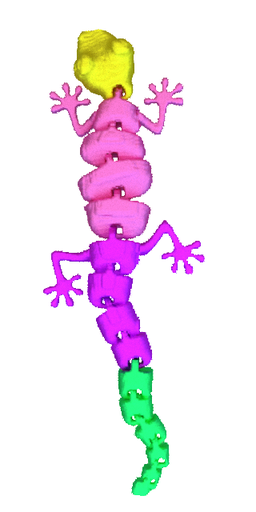}}\\

              \rotatebox{90}{\includegraphics[width=0.06\textwidth, margin=0pt 0pt 0pt 0pt,valign=m]{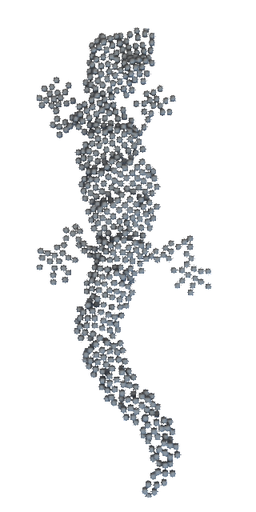}}
            & \rotatebox{90}{\includegraphics[width=0.06\textwidth, margin=0pt 0pt 0pt 0pt,valign=m]{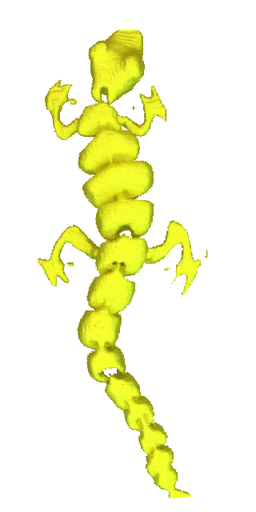}}
            & \rotatebox{90}{\includegraphics[width=0.06\textwidth, margin=0pt 0pt 0pt 0pt,valign=m]{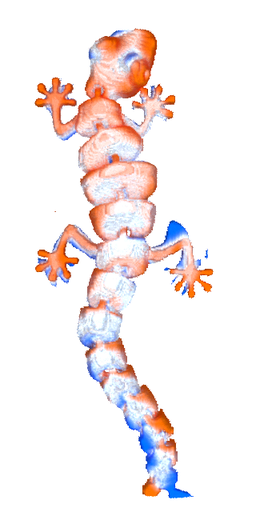}}
            & \rotatebox{90}{\includegraphics[width=0.06\textwidth, margin=0pt 0pt 0pt 0pt,valign=m]{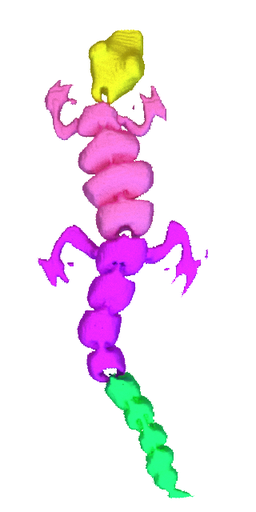}}
            & \rotatebox{90}{\includegraphics[width=0.06\textwidth, margin=0pt 0pt 0pt 0pt,valign=m]{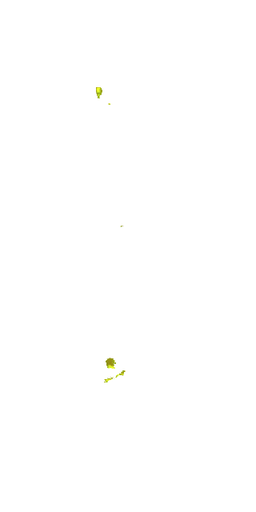}}
            & \rotatebox{90}{\includegraphics[width=0.06\textwidth, margin=0pt 0pt 0pt 0pt,valign=m]{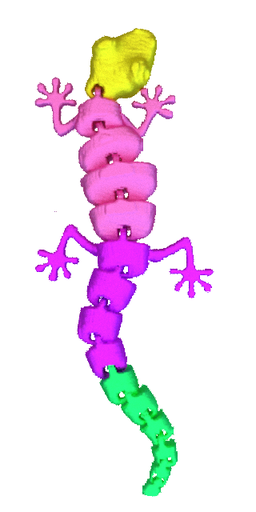}}\\

              \rotatebox{-90}{\includegraphics[width=0.06\textwidth, margin=0pt 0pt 0pt 0pt,valign=m]{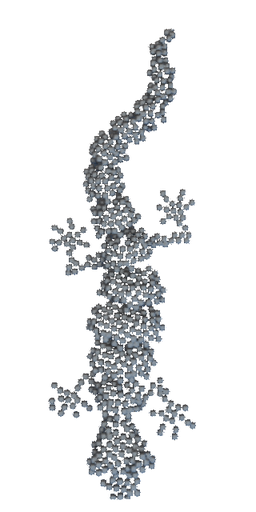}}
            & \rotatebox{-90}{\includegraphics[width=0.06\textwidth, margin=0pt 0pt 0pt 0pt,valign=m]{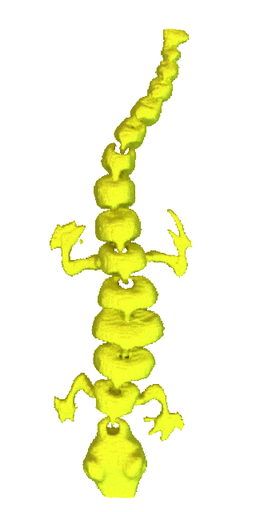}}
            & \rotatebox{-90}{\includegraphics[width=0.06\textwidth, margin=0pt 0pt 0pt 0pt,valign=m]{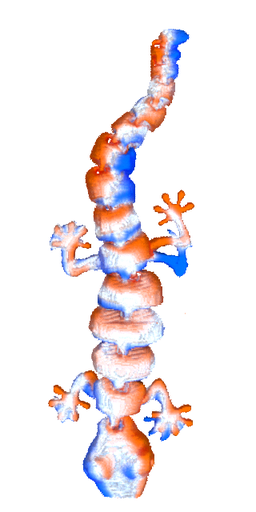}}
            & \rotatebox{-90}{\includegraphics[width=0.06\textwidth, margin=0pt 0pt 0pt 0pt,valign=m]{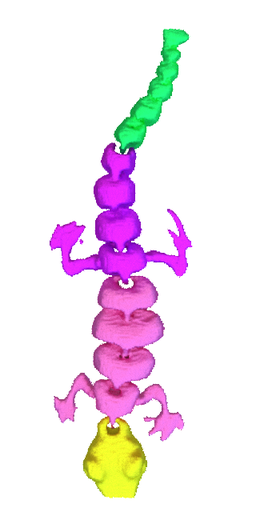}}
            & \rotatebox{-90}{\includegraphics[width=0.06\textwidth, margin=0pt 0pt 0pt 0pt,valign=m]{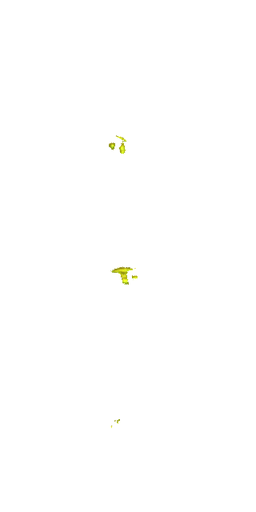}}
            & \rotatebox{-90}{\includegraphics[width=0.06\textwidth, margin=0pt 0pt 0pt 0pt,valign=m]{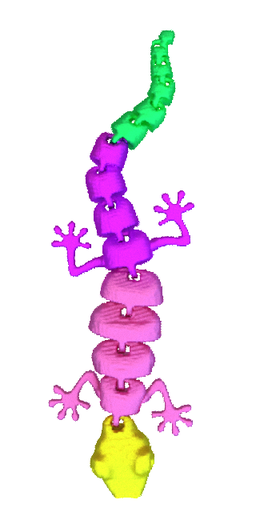}}\\

        \bottomrule
    \end{tabular}
}
\captionof{figure}{Extension of Table~\ref{fig:reconstructions_real_world_small}.
        All reconstructions generated by conditioning the system trained on synthetic data using a real-world input \textbf{Point Cloud}.
        The \textbf{Reconstruction} takes all non-empty classes to be the same.
        \textbf{Reconstruction Error} identifies \textcolor{blue}{over-reconstruction} and \textcolor{orange}{under-reconstruction} when compared with the reference.
        \textbf{Segmentation} colors each class uniquely, resulting in a \textbf{Segmentation Error} anywhere it differs from the \textbf{Reference}.}
\vspace{\textfloatsep}
\end{minipage}